%% file: main.tex
\PassOptionsToPackage{table}{xcolor}  
\documentclass[sigconf]{acmart}       

\AtBeginDocument{%
  \providecommand\BibTeX{{\textrm B\kern-.05em{\sc i\kern-.025em b}\kern-.08em\TeX}}}

\copyrightyear{2025}
\acmYear{2025}
\setcopyright{acmlicensed}
\acmConference[KDD '25]{Proceedings of the 31st ACM SIGKDD Conference on Knowledge Discovery and Data Mining}{August 3--7, 2025}{Toronto, ON, Canada}
\acmDOI{10.1145/3711896.3737161}
\acmISBN{979-8-4007-1454-2/2025/08}
\makeatletter
\@ifundefined{acmBooktitle}{}{%
  \acmBooktitle{Proceedings of the 31st ACM SIGKDD Conference on Knowledge Discovery and Data Mining (KDD~'25), August 3--7, 2025, Toronto, ON, Canada}}
\makeatother

\usepackage[utf8]{inputenc}
\usepackage{amsmath,amsfonts}
\usepackage{graphicx}
\usepackage{subcaption}
\usepackage[table]{xcolor}   
\usepackage{booktabs}
\usepackage{enumitem}
\usepackage{multirow}
\usepackage{algorithm}
\usepackage{algpseudocode}
\usepackage{makecell}
\usepackage{pifont}
\usepackage{placeins}
\usepackage{xspace}
\usepackage{soul}            

\newcommand{\matDeltaSigma}{{\delta}_{\Sigma}}
\newcommand{\vecDeltaSigma}{\mathrm{vec}({\delta}_{\Sigma})}
\newcommand{\ours}{\texttt{FedAKD}\xspace}

\definecolor{OliveGreenRGB}{rgb}{0,0.6,0}
\definecolor{PineGreen}{RGB}{0,139,114}
\definecolor{BrickRed}{RGB}{140,55,62}
\definecolor{greyL}{RGB}{230,248,255}

\newtheorem{assumption}{Assumption}

\begin{document}

\title{Towards Collaborative Fairness in Federated Learning Under Imbalanced Covariate Shift}

\author{Tianrun Yu}
\affiliation{%
  \institution{The Pennsylvania State University}
  \city{University Park}\state{PA}\country{USA}}
\email{tvy5242@psu.edu}

\author{Jiaqi Wang}
\affiliation{%
  \institution{The Pennsylvania State University}
  \city{University Park}\state{PA}\country{USA}}
\email{jqwang@psu.edu}

\author{Haoyu Wang}
\affiliation{%
  \institution{State University of New York at Albany}
  \city{Latham}\state{NY}\country{USA}}
\email{hwang28@albany.edu}

\author{Mingquan Lin}
\affiliation{%
  \institution{University of Minnesota Twin Cities}
  \city{Minneapolis}\state{MN}\country{USA}}
\email{lin01231@umn.edu}

\author{Han Liu}
\affiliation{%
  \institution{Dalian University of Technology}
  \city{Dalian}\state{Liaoning}\country{China}}
\email{liu.han.dut@gmail.com}

\author{Nelson S. Yee}
\affiliation{%
  \institution{The Pennsylvania State University}
  \city{Hershey}\state{PA}\country{USA}}
\email{nyee@pennstatehealth.psu.edu}

\author{Fenglong Ma}
\affiliation{%
  \institution{The Pennsylvania State University}
  \city{University Park}\state{PA}\country{USA}}
\email{fenglong@psu.edu}
\authornote{Corresponding author.}

\renewcommand{\shortauthors}{Tianrun Yu et al.}

\begin{abstract}
Collaborative fairness is a crucial challenge in federated learning. However, existing approaches often overlook a practical yet complex form of heterogeneity: \textit{imbalanced covariate shift}. We provide a theoretical analysis of this setting, which motivates the design of \ours{} (Federated Asynchronous Knowledge Distillation)---a simple yet effective approach that balances accurate prediction with collaborative fairness. \ours{} consists of client and server updates. In the client update, we introduce a novel asynchronous knowledge distillation strategy based on our preliminary analysis, which reveals that while correctly predicted samples exhibit similar feature distributions across clients, incorrectly predicted samples show significant variability. This suggests that imbalanced covariate shift primarily arises from misclassified samples. Leveraging this insight, our approach first applies traditional knowledge distillation to update client models while keeping the global model fixed. Next, we select correctly predicted high-confidence samples and update the global model using these samples while keeping client models fixed. The server update simply aggregates all client models. We further provide a theoretical proof of \ours{}'s convergence. Experimental results on public datasets (FashionMNIST and CIFAR10) and a real-world Electronic Health Records (EHR) dataset demonstrate that \ours{} significantly improves collaborative fairness, enhances predictive accuracy, and fosters client participation even under highly heterogeneous data distributions.
\footnote{Source code is available at \url{https://github.com/Tianrun-Yu/FedAKD}.}
\end{abstract}


\maketitle

\input{Introduction}

\input{shift}

\input{Approach}

\input{ConvergenceAnalysis}
\input{Experiments}
\input{RelatedWork}

\input{Conclusion}
\bibliographystyle{ACM-Reference-Format}
\bibliography{references}
\input{Appendix}

\end{document}

%% file: Introduction.tex
\section{Introduction}\label{sec:intro}
Federated Learning (FL) has emerged as a promising distributed paradigm that enables multiple participants (or clients) to collaboratively train a global model without sharing their raw local data~\cite{konecny2016federated,li2020federated,zhao2018federated,mcmahan2017communication,yan2023criticalfl}. However, disparities in data quality, quantity, and distribution among clients make uniform treatment in global model aggregation unfair, particularly for those with higher-quality or larger datasets. To address this, \textbf{collaborative fairness} (CF) has been introduced to ensure that each client’s final reward or benefit is commensurate with its contribution to the global model~\cite{Lyu2020CollaborativeFairness}. In other words, clients with a greater impact on model performance should receive proportionally higher gains.

\begin{figure*}[t]
    \centering
    \includegraphics[width=1\linewidth]{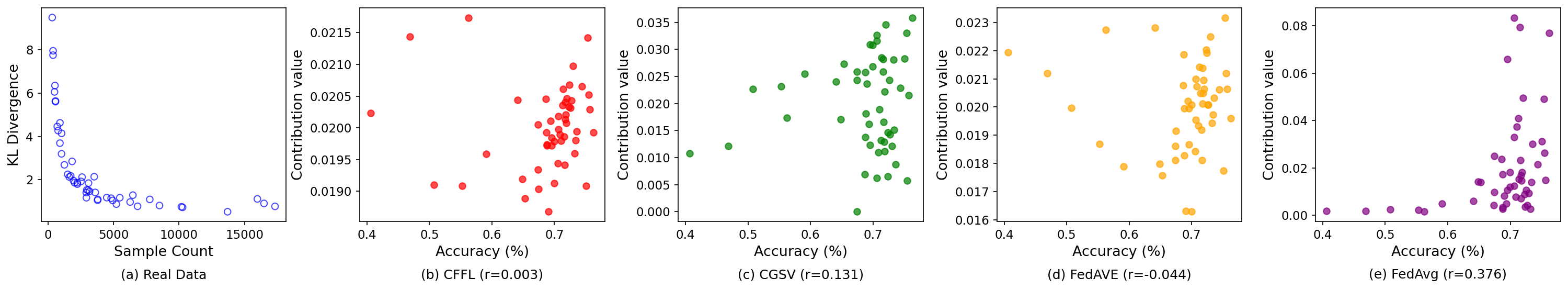}
    \vspace{-0.3in}
    \caption{(a) KL divergence vs.\ sample size for each client’s local data, revealing both data imbalance and feature covariate shifts. (b)-(e) compare different fairness methods’ contribution definitions to each client’s standalone training accuracy. Their poor correlation highlights the limitations of explicit contribution metrics under extra covariate shifts.}
    \label{fig:5contri}
    \vspace{-0.15in}
\end{figure*}

Several fairness-aware FL approaches, such as CGSV~\cite{xu2021gradient}, CFFL~\cite{Lyu2020CollaborativeFairness}, FedAve~\cite{wang2024fedave}, and FedSAC~\cite{wang2024fedsac}, have been developed to assign different contribution-based rewards (i.e., weights) during model aggregation. While these methods help mitigate fairness disparities, they still face the following challenges:

$\bullet$ \textbf{Unrealistic assumptions about data heterogeneity.}
Existing CF approaches primarily assume that data heterogeneity arises from imbalanced data sizes~\cite{xu2021gradient,Lyu2020CollaborativeFairness}, imbalanced class distributions~\cite{xu2021gradient,Lyu2020CollaborativeFairness}, or both~\cite{g,yurochkin2019bayesian}. However, real-world datasets exhibit greater complexity. Beyond these imbalances, client data often differ significantly in feature distributions, leading to the \textit{covariate shift} problem~\cite{karimireddy2020scaffold, g, li2020federated, goksu2024robust}.
Figure~\ref{fig:5contri}(a) presents a preliminary analysis on a real-world electronic health record (EHR) dataset for pancreatic cancer prediction\footnote{Details on the EHR dataset and the preliminary analysis are provided in Sections~\ref{sec:exp3-analysis} and~\ref{sec:theory}, respectively.}. The $x$-axis represents client dataset sizes, while the $y$-axis shows the Kullback–Leibler (KL) divergence between each client’s fitted latent feature distribution and that of the entire dataset. Each circle represents a client, i.e., a U.S. state in the EHR dataset. The preliminary data analysis reveals that not only do clients have varying dataset sizes, but their latent feature distributions also significantly diverge from the global reference distribution. Thus, a more realistic FL heterogeneity setting should account for \textbf{imbalanced covariate shift} rather than just data quantity or class imbalance.

$\bullet$ \textbf{Weak correlation between client accuracy and assigned contributions.}
Beyond unrealistic data distribution assumptions, existing approaches to collaborative fairness typically follow a two-step pipeline: \textit{(1) explicitly defining a contribution metric and (2) allocating rewards based on this metric}. For example, CGSV~\cite{xu2021gradient} estimates client contributions via gradient similarity, CFFL~\cite{Lyu2020CollaborativeFairness} and FedAVE~\cite{wang2024fedave} rely on performance improvements measured by a global validation set, and FedSAC~\cite{wang2024fedsac} bases contributions on standalone local training results. However, in real-world datasets characterized by imbalanced covariate shift, these approaches fail to establish a strong correlation between client accuracy and assigned contributions, contradicting their underlying design assumptions.

Figures~\ref{fig:5contri}(b)--(e) illustrate the relationship between client standalone accuracy ($x$-axis) on the testing set and the learned contribution value ($y$-axis) either on the training set or validation set under different methods -- CFFL, CGSV, FedAVE, and the baseline FedAvg~\cite{mcmahan2017communication} -- using the real-world EHR dataset, which is the same as we analyzed in Figure~\ref{fig:5contri}(a). Each dot represents an individual client, and in the case of FedAvg, the contribution score is simply the proportion of data owned by the client. The results show that existing CF-based methods exhibit significantly lower Pearson correlation scores compared to the simple FedAvg baseline. These findings highlight the limitations of existing approaches in handling realistic imbalanced covariate shift settings.

\textbf{Theoretical analysis on imbalanced covariate shift.}
This paper aims to develop a novel model to ensure collaborative fairness in federated learning under the realistic yet challenging imbalanced covariate shift setting. To achieve this, in Section~\ref{sec:theory}, we first provide a theoretical analysis in Theorem~\ref{theorem1} demonstrating that imbalanced covariate shift -- quantified as the KL divergence between each client's empirical data distribution and the ideal global distribution -- is primarily influenced by the perturbation $\delta$. Specifically, if the underlying data distributions of individual clients and the entire dataset follow a multivariate normal distribution $\mathcal{N}(\mu,\Sigma)$, where $\mu$ is the mean vector, we show that $\delta$ and the covariance $\Sigma$ play key roles in determining the extent of distributional divergence in Theorem~\ref{theorem2}.
The two theorems motivate us to mitigate the imbalanced covariate shift to achieve collaborative fairness by correctly qualifying $\delta$ and $\Sigma$. However, directly calculating these values is infeasible, as the true data distributions are inherently \textit{unknown}.

\begin{figure*}[t]
    \centering
    \includegraphics[width=0.9\linewidth]{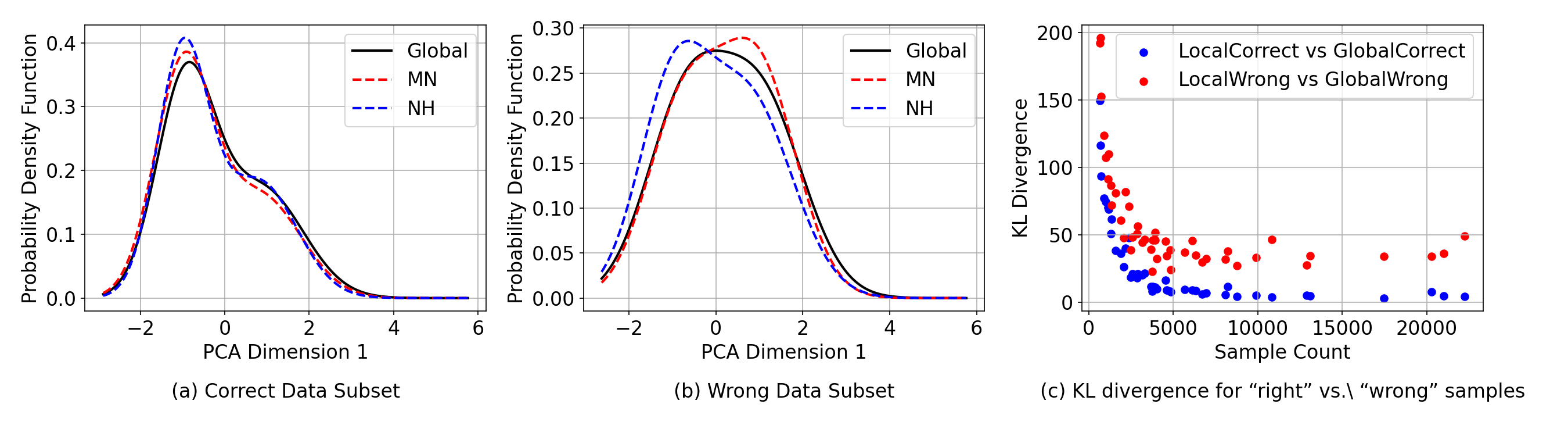}
    \vspace{-0.25in}
\caption{(a) Distribution of locally versus globally correct samples.
         (b) Distribution of locally versus globally incorrect samples.
         (c) KL divergence for ``right'' vs.\ ``wrong'' samples. We observe that the feature distributions of correctly classified samples closely resemble the global distribution, whereas those of misclassified samples deviate significantly. This suggests that imbalanced covariate shift primarily arises from incorrectly classified samples.}
    \label{fig:right_wrong_dis}
\end{figure*}

\textbf{Motivations of model design.}
To address this challenge, we conducted a preliminary analysis on the results of FedAvg applied to the entire EHR dataset. For each client $k$, we categorized the correctly and incorrectly predicted samples as $\mathcal{I}_k$ and $\mathcal{D}_k-\mathcal{I}_k$, respectively, and aggregated these categories across all clients to form the global sets $\{\mathcal{I}_k\}_{k=1}^K$ and $\{\mathcal{D}_k-\mathcal{I}_k\}_{k=1}^K$, where $\mathcal{D}_k$ denotes the client dataset, and $K$ is the number of clients. Next, we applied principal component analysis (PCA) to project each sample's latent representation (i.e., the encoder output from each client model) onto a 1-D space and used kernel density estimation (KDE) to estimate the probability density function (PDF) for each set.
Figures~\ref{fig:right_wrong_dis}(a) and (b) show a comparison of the global density function with the density functions of two clients (\textit{Minnesota} and \textit{New Hampshire}). The $x$-axis represents the projected PCA 1-D values, and the $y$-axis represents the estimated PDF values. Similar to Figure~\ref{fig:5contri}(a), we also analyze the distribution differences between local and global data in terms of correct and incorrect classifications, as illustrated in Figure~\ref{fig:right_wrong_dis}(c).
These results indicate that \textit{the primary source of imbalanced covariate shift lies in the ``incorrect'' samples, as clients generally show agreement on the distribution of ``correct'' predictions}.

\textbf{Our approach.}
Building on our theoretical and empirical analysis, we propose \ours, a novel framework designed to address the imbalanced covariate shift challenge while ensuring collaborative fairness. \ours leverages a new \underline{\textbf{fed}}erated \underline{\textbf{a}}synchronous \underline{\textbf{k}}nowledge \underline{\textbf{d}}istillation approach, comprising client updates and server updates in each communication round $t$. Specifically, the \textit{client update} includes three key steps: (1) \textbf{Global $\to$ Local Distillation:} Using traditional knowledge distillation~\cite{hinton2015distilling}, we employ the global model $\mathbf{w}_g^t$ as a teacher to guide the client model $\mathbf{w}_k^t$ in learning from its full training set $\mathcal{D}_k$. (2) \textbf{High-confidence Sample Selection:} Inspired by our observations in Figure~\ref{fig:right_wrong_dis}, correctly predicted samples positively contribute to the global model update. Thus, we first identify the correctly classified samples from the updated local model, denoted as $\mathcal{I}_k^t$. (3) \textbf{Local $\to$ Global Distillation}: The global model $\mathbf{w}_{g}^t$ is then refined by distilling ``high-confidence'' client knowledge from $\mathbf{w}_{k}^t$ using each client's correctly predicted set $\mathcal{I}_k^t$. This design helps mitigate distortions caused by misclassified data under covariate shift conditions. The updated global model for each client (i.e., $\mathbf{w}_{g,k}^t$) is uploaded to the server, where it is aggregated following FedAvg~\cite{mcmahan2017communication} in the \textit{server update} step. These two updates iterate until \ours converges. The theoretical convergence of \ours is established in Theorem~\ref{thm:fedakd_convergence_full}.

Through this two-stage asynchronous distillation process, \ours effectively encourages fair collaboration even when clients have highly divergent feature distributions. High-quality participants benefit by sharing more correct samples, whereas lower-quality participants gain from the improved global knowledge, thus collectively promoting \emph{collaborative fairness} without imposing rigid or impractical contribution measurements.

\textbf{Contributions.}
The main contributions of this work include:
\begin{itemize}[leftmargin=*]
    \item \textbf{A new heterogeneity setting.} 
    We introduce collaborative fairness under the imbalanced covariate shift, a practical challenge in real-world medical datasets where both sample-size imbalance and feature-distribution mismatch significantly hinder the effectiveness of existing collaborative fairness metrics.

    \item \textbf{A simple yet effective solution.} 
    We propose \ours, a novel asynchronous distillation framework that first distills knowledge from correctly predicted local samples to improve the global model quality, followed by an inverse distillation step to enhance client learning across the full dataset.

    \item \textbf{Theoretical guarantees.} 
    We provide a rigorous theoretical analysis of imbalanced covariate shift, expanding the KL divergence parametrically to model real-world heterogeneity. Additionally, we prove the convergence of \ours under broad heterogeneity conditions, ensuring both theoretical soundness and improved collaborative fairness.

    \item \textbf{Promising results.}
    We conduct extensive experiments on three datasets, evaluating \ours against ten baselines across four heterogeneity settings using three evaluation metrics. The results demonstrate that \ours effectively addresses imbalanced covariate shift and outperforms all baselines across diverse heterogeneity scenarios.
\end{itemize}

%% file: shift.tex
\section{Imbalanced Covariate Shift Analysis}
\label{sec:theory}

Imbalanced covariate shift presents a significant and open challenge in federated learning, requiring a deeper mathematical understanding to effectively mitigate disparities in collaborative fairness. To formalize this, we assume the existence of a global data feature distribution \(p_{\omega}\). Each client's data distribution is modeled as a small parametric perturbation, denoted as \(p_{\omega + \delta}\), where $\delta \in \mathbb{R}^N$ represents the perturbation vector and $N$ is the model size. When a client draws an i.i.d. sample set of size $A$ from \(p_{\omega + \delta}\), it results in the empirical distribution \(\widehat{p}_{\omega + \delta}\). Consequently, the imbalanced covariate shift can be quantitatively assessed by measuring the KL divergence between the empirical distribution \(\widehat{p}_{\omega + \delta}\) and the original global distribution \(p_{\omega}\).

\begin{theorem}[Ideal Imbalance Covariate Shift Quantification]
\label{theorem1}
Let $\{p_\theta\}_{\theta \in \Theta}$ be a smooth parametric family of probability distributions, and let $\omega \in \Theta$ denote a baseline parameter.
Suppose that a perturbed distribution $p_{\omega+\delta}$ is defined by a small perturbation $\delta\in\mathbb{R}^N$.
Given that $p_{\omega+\delta}$ is estimated via an empirical distribution $\widehat{p}_{\omega+\delta}$ using $A$ i.i.d.\ samples and $R$ (free) parameters, we have the following approximation under standard regularity conditions and a large-sample limit:
\begin{align*}
    D_{\mathrm{KL}}\!\Bigl(\widehat{p}_{\,\omega+\delta} \,\Big\|\; p_{\omega}\Bigr)
~\approx~
\frac{1}{2}\delta^\top I(\omega)\,\delta+\frac{1}{2}\delta^\top (\nabla_\omega I(\omega)\,\delta) \delta 
\;+\;
\frac{R}{2\,A},
\end{align*}
where $I(\omega)$ is the Fisher information matrix at $\omega$, and $\nabla_\omega I(\omega)$ denotes the gradients with respect to $\omega$.
\end{theorem}

We assume that the probabilistic distribution follows an $M$-dimensional Gaussian distribution. We then extend Theorem~\ref{theorem1} as follows:

\begin{theorem}[Imbalance Covariate Shift Quantification Under Gaussian Distribution]
\label{theorem2}
Let \(p_{\mu,\Sigma}(x) = \mathcal{N}(x;\,\mu,\Sigma)\) be a \(M\)-dimensional Gaussian distribution,
where \(\mu \in \mathbb{R}^M\) is the mean vector and \(\Sigma \in \mathbb{R}^{M \times M}\) is a symmetric positive-definite covariance matrix.
Suppose \(\omega = (\mu_0,\Sigma_0)\) is a baseline parameter,
and consider a small perturbation
$
  \delta = (\delta_\mu,\,\delta_\Sigma), 
  \theta' = \omega + \delta \approx (\mu_0 + \delta_\mu,\,\Sigma_0 + \delta_\Sigma).
$
Under a large-sample limit, the Kullback-Leibler divergence
between the empirical distribution \(\widehat{p}_{\,\theta'}\) (fitted from \(A\) i.i.d.\ samples
drawn from \(p_{\theta'}\)) and the baseline model \(p_{\omega}\) can be approximated by:
\begin{equation}
\begin{split}
    D_{\mathrm{KL}}(\widehat{p}_{\,\theta'} \,\| p_{\omega})
    ~\approx~&
  \frac{1}{4}\|\Sigma^{-1}\,\matDeltaSigma\,\Sigma^{-1}\|_F^2-\,\frac{1}{2}
\mathrm{trace}\!\Bigl((\matDeltaSigma\,\Sigma^{-1})^3\Bigr)\\ &+\frac{1}{2}(\delta_\mu)^\top (\Sigma^{-1}(I-\matDeltaSigma\Sigma^{-1}))\delta_\mu+\frac{M(M+3)}{4 A}.
\end{split}
\label{eq:gaussian_KL}   
\end{equation}
\end{theorem}
Together, these two theorems provide a unified mathematical framework to model the combined effects of \emph{imbalanced sample sizes} and \emph{covariate shift}. This framework provides a principled approach to analyzing how local client distributions diverge from the global baseline, offering deeper insights into collaborative fairness in federated learning. Detailed proofs are provided in \textbf{Appendix~\ref{sec:appedixa}}. We also provide a theoretical approximation validation experiment in \textbf{Appendix~\ref{sec:the-2}} to validate the correctness of our theorems.


%% file: Approach.tex
\section{The Proposed \ours}
\label{sec:3}

While our theoretical analysis in Section~\ref{sec:theory} provides valuable insights into imbalanced covariate shift, it cannot be directly applied to model design, as the global distribution remains \textbf{\textit{unknown}} in federated learning. However, these theorems reveal that the imbalanced covariate shift arises due to small perturbations in client distributions. Our preliminary analysis (Figure~\ref{fig:right_wrong_dis} in Section~\ref{sec:intro}) further suggests that these perturbations predominantly stem from incorrectly classified samples. This observation motivates us to develop an effective collaborative fairness approach named \ours that mitigates the imbalanced covariate shift by addressing the impact of misclassified samples via a simple asynchronous knowledge distribution strategy. The algorithm is shown in Algorithm~\ref{alg:fedxx}.

Similar to existing collaborative fairness approaches in federated learning, \ours comprises both client and server updates. However, unlike prior methods that require carefully designing reward weights for each client~\cite{xu2021gradient,Lyu2020CollaborativeFairness,wang2024fedave,wang2024fedsac}, \ours simplifies aggregation by directly following the standard FedAvg~\cite{mcmahan2017communication} in the server update. The novelty of \ours lies in the client update, where we introduce a new \textbf{asynchronous knowledge distillation} strategy, inspired by our preliminary analysis. Specifically, the client update consists of three key steps: (1) global $\to$ local distillation, (2) high-confidence sample selection, and (3) local $\to$ global distillation. Next, we provide the details of these three steps.

\begin{algorithm}[ht]
\caption{\ours}
\label{alg:fedxx}
{\small
\begin{algorithmic}[1]

\Require 
  $K$ clients; local datasets $\{\mathcal{D}_k\}$;  
  total rounds $T$; learning rate $\eta$; distillation coefficients $\alpha$ and $\beta$.

\vspace{0.3em}
\State \textbf{Initialization:}
\State \quad Generate an initial model $\mathbf{w}^{0}$ (e.g., randomly);
\State \quad \textbf{Client side (for each $k$)}: set each local model $\mathbf{w}_k^{0} = \mathbf{w}^{0}$;
\State \quad \textbf{Server side}: set the global model $\mathbf{w}_g^{1} = \mathbf{w}^{0}$ and distribute 
\State \quad $\mathbf{w}_g^{1}$ to clients;

\vspace{0.5em}
\For{$t=1,\cdots,T$}
    \State  {// \textcolor{red}{\textbf{Client Update}}}
    \For{$k=1,2,\cdots,K$}
        \State  {// \textcolor{blue}{Step 1: Global $\to$ Local Distillation}}
        \State $\mathbf{w}_{g,k}^t \leftarrow \mathbf{w}_{g}^t$;
        \State  // Loss computation by fixing $\mathbf{w}_{g,k}^t$ using $\mathcal{D}_k$
        \State $\overrightarrow{\mathcal{L}_t} = \text{CE}(\mathbf{w}_{k}^{t-1}; \mathcal{D}_k) + \alpha \text{KD}(\mathbf{w}_{k}^{t-1}; \mathbf{w}_{g,k}^{t}, \mathcal{D}_k)$;
        \State  //update model parameters
        \State  $
           \mathbf{w}_{k}^{t}\leftarrow
           \mathbf{w}^{t-1}_{k} -
           \eta \nabla
           \overrightarrow{\mathcal{L}_t}$;
        \State // \textcolor{blue}{Step 2: High-confidence Sample Selection}
        \State $\mathcal{I}_k^{t}
\;=\;
\Bigl\{
  (\mathbf{x}, y) \in \mathcal{D}_k 
  \;\Big|\;
  \text{Pred}(\mathbf{w}_k^{t}, \mathbf{x}) \;=\; y
\Bigr\};$
    \State  {// \textcolor{blue}{Step 3: Local $\to$ Global Distillation}}
    \State  // Loss computation by fixing $\mathbf{w}_{k}^t$ using $\mathcal{I}_k^t$
        \State $\overleftarrow{\mathcal{L}_t} = \text{CE}(\mathbf{w}_{g,k}^{t}; \mathcal{I}_k^t) + \beta \text{KD}(\mathbf{w}_{g,k}^t; \mathbf{w}_k^{t}, \mathcal{I}_k^t)$;
        \State  // Update model parameters
        \State  $
           \mathbf{w}_{g,k}^{t+1}\leftarrow
           \mathbf{w}_{g,k}^{t} -
           \eta \nabla
           \overleftarrow{\mathcal{L}_t}$;
        \State  Upload $\mathbf{w}_{g,k}^{t+1}$  to the server;
    \EndFor

    \State {// \textcolor{red}{\textbf{Server Update}}}
    \State $\mathbf{w}^{t+1}_g
      \;=\;
      \frac{1}{
         \sum_{k=1}^{K}\!\bigl\lvert \mathcal{D}_k\bigr\rvert
      }
      \;\sum_{k=1}^{K}
         \Bigl(
            \bigl\lvert \mathcal{D}_k\bigr\rvert 
            \,\mathbf{w}_{g,k}^{t+1}
         \Bigr);
    $
    \State  Distribute $\mathbf{w}_{g}^{t+1}$ to each client;

\EndFor

\State \textbf{Output:} The global model $\mathbf{w}^{T}_g$ and local models $\{\mathbf{w}^{T}_k\}$.

\end{algorithmic}
}
\end{algorithm}

\subsection{Step 1: Global $\to$ Local Distillation}
\label{sec:global2local}

The global model $\mathbf{w}_g^t$ contains aggregated knowledge, but forcing all clients to adopt \(\mathbf{w}_{g,k}^{t} = \mathbf{w}_g^t\) directly may harm local performance due to the imbalanced covariate shift. To avoid this issue,  we propose global $\to$ local distillation, enabling each client to \emph{selectively} adopt global insights while retaining local specialization by optimizing the following loss:
\begin{equation}
\label{eq:glob2locl_loss}
\overrightarrow{\mathcal{L}_t} = \text{CE}(\mathbf{w}_{k}^{t-1}; \mathcal{D}_k)
+
\alpha \,\text{KD}(\mathbf{w}_{k}^{t-1}; \mathbf{w}_{g,k}^{t}, \mathcal{D}_k),
\end{equation}
where $\text{CE}$ denotes the cross-entropy loss, $\text{KD}$ is the knowledge distillation loss, and $\alpha$ is the hyperparameter. The gradient update yields:
\begin{equation}
\mathbf{w}_{k}^{t}
\,=\,
\mathbf{w}_{k}^{t-1} 
\;-\;
\eta\,\nabla \overrightarrow{\mathcal{L}_t},
\end{equation}
where $\eta$ is the learning rate. In this step, we fix the global model parameters $\mathbf{w}_{g,k}^t$ and only update the client model parameters $\mathbf{w}_{k}^{t-1}$ using the full training set $\mathcal{D}_k$. 
This procedure allows each client to merge the updated global knowledge with its local parameters, safeguarding performance for distribution-mismatched (yet high-quality) clients. Consequently, no client is penalized for joining the federation, reinforcing the incentives for collaborative fairness under the imbalanced covariate shift.

\subsection{Step 2: High-confidence Sample Selection}
\label{sec:selection}
Our preliminary analysis (Figures~\ref{fig:right_wrong_dis} in Section~\ref{sec:intro}) suggests that imbalanced covariate shift primarily arises from misclassified samples on each client, whereas correctly classified samples positively contribute to global model learning. 
To address this, we select high-confidence samples (i.e., correctly classified samples) to update the global model $\mathbf{w}_{g,k}^t$, which is denoted as: 
\begin{equation}
  \mathcal{I}_k^{t}
\;=\;
\Bigl\{
  (\mathbf{x}, y) \in \mathcal{D}_k 
  \;\Big|\;
  \text{Pred}(\mathbf{w}_k^{t}, \mathbf{x}) \;=\; y
\Bigr\}.  
\end{equation}

\subsection{Step 3: Local $\to$ Global Distillation}
\label{sec:local2global}
Unlike existing bidirectional knowledge distillation~\cite{shang2023fedbikd,kweon2021bidirectional} that updates two models simultaneously using the same dataset, we propose an asynchronous knowledge distillation approach for this step. Additionally, we leverage only the selected high-confidence samples $\mathcal{I}_k^{t}$, enabling the local model $\mathbf{w}_k^t$ to guide the learning of the global model $\mathbf{w}_{g,k}^t$ by optimizing the following loss: 
\begin{equation}
\label{eq:locl2glob_loss}
\overleftarrow{\mathcal{L}_t} = \text{CE}(\mathbf{w}_{g,k}^{t}; \mathcal{I}_k^t)
+
\beta \,\text{KD}(\mathbf{w}_{g,k}^{t}; \mathbf{w}_{k}^{t}, \mathcal{I}_k^t),
\end{equation}
where $\beta$ is the hyperparameter. The gradient update yields:
\begin{equation}
\mathbf{w}_{g,k}^{t+1}
\,=\,
\mathbf{w}_{g,k}^{t} 
\;-\;
\eta\,\nabla \overleftarrow{\mathcal{L}_t},
\end{equation}

The proposed asynchronous knowledge distillation offers three key benefits: (1) \textit{Robustness to Noisy Updates}. It protects the global model from noisy or erroneous updates by discarding locally misclassified samples.
(2) \textit{Fair Collaboration}. It promotes fairness by allowing high-quality clients—those that classify more samples correctly—to have a stronger influence without explicitly revealing their accuracy or contributions.
(3) \textit{Adaptability to Imbalanced Covariate Shift}. It ensures that even if a client’s data distribution differs significantly from the global average, it can still contribute reliable knowledge.




%% file: ConvergenceAnalysis.tex
\subsection{Convergence Analysis}
\subsubsection{Notions and Assumptions}
\label{subsubsec:notion_assumptions}

In this section, we consider a binary classification problem following~\cite{Ni2022} with input space 
\(
  X \in \mathbb{R}^d
\) 
and label space 
\(
  Y = \{0,1\}.
\)
We employ a linear classification setting: for each local sample \(\mathbf{x}\in X\), 
the logits are 
\(
  z = \mathbf{x}^\top \mathbf{w},
\)
and the predicted probability is 
\(
  \hat{y}(\mathbf{x}) = \sigma\bigl(z\bigr) 
           = \frac{1}{1 + e^{-z}}.
\)
In our codistillation setup, the distillation temperature is set to \(\tau = 1\), 
keeping the standard sigmoid form.
We denote the cross-entropy loss by 
\(\mathcal{L}(\mathbf{w}; \mathcal{D}) = \text{CE}(\mathbf{w}; \mathcal{D}) 
= \frac{1}{|\mathcal{D}|} \sum_{\mathbf{x}_i \in \mathcal{D}}
\Bigl[
  -\,y_i \,\log\bigl(\hat{y}(\mathbf{x}_i)\bigr)
  -\bigl(1 - y_i\bigr)\,\log\bigl(1 - \hat{y}(\mathbf{x}_i)\bigr)
\Bigr]\).
Here, \(\mathbf{w}\) is the model parameter vector, and \(\mathcal{D}\) is the training dataset consisting of samples \(\mathbf{x}_i\) with labels \(y_i\).
We further introduce the KD loss, denoted by 
\(\mathrm{KD}(\mathbf{w}, \mathbf{w}_0; \mathcal{D}) 
= \frac{1}{|\mathcal{D}|} \sum_{\mathbf{x}_i \in \mathcal{D}}
\Bigl[
  -\,\hat{y}_0(\mathbf{x}_i)\,\log\!\bigl(\hat{y}(\mathbf{x}_i)\bigr)
  -\bigl(1 - \hat{y}_0(\mathbf{x}_i)\bigr)\,\log\!\bigl(1 - \hat{y}(\mathbf{x}_i)\bigr)
\Bigr]\).
This can also be written in expectation form as 
\(\mathbb{E}_{\mathbf{x}_{i} \in \mathcal{D}}
\Bigl[
  -\,\hat{y}_0(\mathbf{x}_i)\,\log\!\bigl(\hat{y}(\mathbf{x}_i)\bigr)
  -\bigl(1 - \hat{y}_0(\mathbf{x}_i)\bigr)\,\log\!\bigl(1 - \hat{y}(\mathbf{x}_i)\bigr)
\Bigr]\).
Here, \(\hat{y}_0(\mathbf{x}_i)\) is the (fixed) teacher model’s output, 
\(\hat{y}_0(\mathbf{x}_i) = \sigma\!\bigl(\mathbf{x}_i^\top \mathbf{w}_0\bigr)\),
and \(\mathbf{w}_0\) denotes the teacher’s parameter vector. This KD objective is equivalent (up to a constant) to minimizing the KL divergence from the teacher distribution~\cite{Li2020FedOpt,Ni2022}.

We define the notion of a \(\gamma\)-inexact solution following~\cite{Li2020FedOpt,Ni2022}, which quantifies the improvement made by local updates:

\begin{definition}[\(\gamma_1\)-inexact solution \cite{Li2020FedOpt,Ni2022}]
\label{def:gamma-inexact}
For a function \(\overrightarrow{\mathcal{L}}(\mathbf{w}, \mathbf{w}_{0};\, \mathcal{D})=\mathcal{L}(\mathbf{w}; \mathcal{D})+\alpha\mathrm{KD}(\mathbf{w}, \mathbf{w}_0; \mathcal{D})\), 
and let \(\gamma_1 \in [0,1]\). Suppose \(\mathbf{w}_0\) is an initial point for 
\(\min_{\mathbf{w}} \; \overrightarrow{\mathcal{L}}\bigl(\mathbf{w};\, \mathbf{w}_{0};\, \mathcal{D}\bigr)\). 
We say \(\mathbf{w}^*\) is a \(\gamma_1\)-inexact solution if
\(
\bigl\|\nabla \overrightarrow{\mathcal{L}}\bigl(\mathbf{w}^*,\, \mathbf{w}_{0};\, \mathcal{D}\bigr)\bigr\|
\;\le\;
\gamma_1 \,\bigl\|\nabla \overrightarrow{\mathcal{L}}\bigl(\mathbf{w}_0,\, \mathbf{w}_{0};\, \mathcal{D}\bigr)\bigr\|.
\)
\end{definition}

A smaller \(\gamma_1\) indicates a greater reduction in the gradient norm relative to the initial point, implying more significant local improvement. Conversely, a larger \(\gamma_1\) indicates a less complete local optimization. Similarly, for the function \(\overleftarrow{\mathcal{L}}\) with coefficient \(\beta\).  
We define 
\(\displaystyle \overleftarrow{\mathcal{L}}(\mathbf{w}, \mathbf{w}_{0};\, \mathcal{D})
= \mathcal{L}(\mathbf{w}; \mathcal{D}) 
  + \beta\,\mathrm{KD}(\mathbf{w}, \mathbf{w}_0; \mathcal{D})\),
and let \(\gamma_2 \in [0,1]\).
we say \(\mathbf{w}^*\) is a \(\gamma_2\)-inexact solution with
\(
  \bigl\|\nabla \overleftarrow{\mathcal{L}}\bigl(\mathbf{w}^*,\, \mathbf{w}_{0};\, \mathcal{D}\bigr)\bigr\|
  \;\le\;
  \gamma_2 \,\bigl\|\nabla \overleftarrow{\mathcal{L}}\bigl(\mathbf{w}_0,\, \mathbf{w}_{0};\, \mathcal{D}\bigr)\bigr\|.
\)

\begin{assumption}[$L$-smoothness \cite{li2019convergence,Li2020FedOpt}]
\label{assump:L-smooth}
There exists \(L>0\) such that for all \(\mathbf{w}, \mathbf{w}'\),
\(
\|\nabla \mathcal{L}(\mathbf{w};\mathcal{D}) - \nabla \mathcal{L}(\mathbf{w}';\mathcal{D})\|
\;\le\;
L \,\|\mathbf{w} - \mathbf{w}'\|.
\)
\end{assumption}
\begin{assumption}[$\mu$-strong convexity \cite{li2019convergence,Li2020FedOpt}]
\label{assump:mu-strong-convex}
There exists \(\mu>0\) such that for all \(\mathbf{w}, \mathbf{w}'\),
\(
\mathcal{L}(\mathbf{w};\mathcal{D})
\;\ge\;
\mathcal{L}(\mathbf{w}';\mathcal{D})
\;+\;
\nabla \mathcal{L}(\mathbf{w}';\mathcal{D})^\top (\mathbf{w}-\mathbf{w}')
\;+\;
\frac{\mu}{2}\|\mathbf{w}-\mathbf{w}'\|^2
\),
\(
\overrightarrow{\mathcal{L}}(\mathbf{w},\mathbf{w}_0;\mathcal{D})
\;\ge\;
\overrightarrow{\mathcal{L}}(\mathbf{w}',\mathbf{w}_0;\mathcal{D})
\;+\;
\nabla \overrightarrow{\mathcal{L}}(\mathbf{w}',\mathbf{w}_0;\mathcal{D})^\top (\mathbf{w}-\mathbf{w}')
\;+\;
\frac{\mu}{2}\|\mathbf{w}-\mathbf{w}'\|^2
\), and \(
\overleftarrow{\mathcal{L}}(\mathbf{w},\mathbf{w}_0;\mathcal{D})
\;\ge\;
\overleftarrow{\mathcal{L}}(\mathbf{w}',\mathbf{w}_0;\mathcal{D})
\;+\;
\nabla \overleftarrow{\mathcal{L}}(\mathbf{w}',\mathbf{w}_0;\mathcal{D})^\top (\mathbf{w}-\mathbf{w}')
\;+\;
\frac{\mu}{2}\|\mathbf{w}-\mathbf{w}'\|^2.
\)
\end{assumption}

\begin{assumption}[Bounded Gradient Dissimilarity \cite{Li2020FedOpt,Ni2022}]
\label{assump:bounded-grad-dissimilarity}
Let \(\mathcal{D}_g := \bigcup_k \mathcal{D}_k\) be the global dataset 
consisting of all local datasets \(\mathcal{D}_k\).
For some \(\epsilon > 0\), define
\(\mathcal{S}_{c}^{\,\epsilon} 
= \{\mathbf{w} \mid \|\nabla \mathcal{L}(\mathbf{w};\mathcal{D}_g)\|^{2} > \epsilon\}\).
There exists \(B_{\epsilon}\) such that for all \(\mathbf{w} \in \mathcal{S}_{c}^{\,\epsilon}\),
\(
  B(\mathbf{w})
  = \sqrt{
      \frac{
        \mathbb{E}_k\!\bigl[
          \|\nabla \mathcal{L}(\mathbf{w};\mathcal{D}_k)\|^{2}
        \bigr]
      }{
        \|\nabla \mathcal{L}(\mathbf{w};\mathcal{D}_g)\|^{2}
      }
    }
  \;\le\;
  B_{\epsilon}.
\)
\end{assumption}

\noindent
Here, \(B(\mathbf{w})\) measures data heterogeneity across devices. If data are IID and \(n_k\to\infty\), 
then \(B(\mathbf{w})\to 1\). Generally, \(B_{\epsilon}\ge 1\), and larger values capture more dissimilar data distributions.

\begin{assumption}[Bounded Gradient Dissimilarity on Subset]
\label{assump:bounded-subset-dissimilarity}
Let \(\mathcal{I} \subseteq \mathcal{D}\) be a subset. There exists \(\theta \ge 0\) such that for all \(\mathbf{w}\),
\(
\bigl\|\nabla \mathcal{L}\bigl(\mathbf{w}; \,\mathcal{I}\bigr)
-
\nabla \mathcal{L}\bigl(\mathbf{w}; \,\mathcal{D}\bigr)
\bigr\|
\;\le\;
\theta
\bigl\|\nabla \mathcal{L}\bigl(\mathbf{w}; \,\mathcal{D}\bigr)\bigr\|.
\)
\end{assumption}

\noindent
This condition ensures that the gradient on a chosen subset does not deviate excessively 
from the gradient on the entire local dataset, thus quantifying the heterogeneity 
between these two distributions.

\subsubsection{Main Results}
\begin{theorem}[\ours Convergence]
\label{thm:fedakd_convergence_full}
Under Assumptions 1--4, assume that $\mathbf{w}^t$ is not a stationary solution and 
the loss function $\mathcal{L}$ is $B$-dissimilar, i.e., $B(\mathbf{w}^t) \le B$. 
If $\alpha$, $\beta$ and $\gamma := \max\{\gamma_1,\gamma_2\}$ are chosen such that
\(
    r=(\frac{4}{\beta||\Omega_2|| }+\frac{4}{\alpha||\Omega_1|| })B-\frac{LB^2}{2}((L(1+\gamma)+\mu)\frac{\gamma(1+\theta)}{\mu}+(1+\theta)+\frac{(1+\gamma)\beta||\Omega_2||}{4\mu})^2
    - (\frac{4}{\beta||\Omega_2|| }+\frac{4}{\alpha||\Omega_1|| })(r_1+r_2)B\)>0,
where $r_1 =((L(1+\gamma)+\mu)\frac{\gamma(1+\theta)}{\mu}+(1+\theta)+\frac{(1+\gamma)\beta||\Omega_2||}{4\mu})(1+\theta)L+\theta+(L(1+\gamma)+\mu)\frac{\gamma(1+\theta)}{\mu}) $, $r_2 = \frac{L(1+\gamma)}{\mu}+\gamma$, $\Omega_1 =\mathbb{E}_k[\mathbb{E}_{\mathbf{x}_{k,i} \in \mathcal{D}_k}[\mathbf{x}_{k,i}\,\mathbf{x}_{k,i}]]$ and $\Omega_2 =\mathbb{E}_k[\mathbb{E}_{\mathbf{x}_{k,i} \in \mathcal{I}_k^t}[\mathbf{x}_{k,i}\,\mathbf{x}_{k,i}]]$,
then FedAKD satisfies
\begin{align*}
\mathcal{L}(\mathbf{w}_g^{t+1};\mathcal{D}_g) - \mathcal{L}(\mathbf{w}^*;\mathcal{D}_g)
\leq
(1-2\mu r)
[\mathcal{L}(\mathbf{w}^t_g;\mathcal{D}_g) - \mathcal{L}(\mathbf{w}^*;\mathcal{D}_g)].
\end{align*}
\end{theorem}
\noindent
Theorem~\ref{thm:fedakd_convergence_full} shows that the global model in FedAKD converges effectively. The detailed proof is provided in Appendix~\ref{appendix:c}.

%% file: Experiments.tex
\begin{table*}[t]
  \centering
  \small
  \setlength{\tabcolsep}{3pt}
  \caption{Performance evaluation of simulation experiments on two datasets (mean $\pm$ std across three runs).}
  \label{tab:exp2_res}

  \subcaptionbox{Max Client Accuracy (\%)}{\label{tab:max-acc}
  \begin{tabular}{|l|c|c|c|c|c|c|c|c|c|c|}
    \hline
    \multirow{2}{*}{\textbf{Method}} &
    \multicolumn{5}{c|}{\textbf{FashionMNIST}} &
    \multicolumn{5}{c|}{\textbf{CIFAR10}} \\
    \cline{2-11}
      & \textbf{POW} & \textbf{BCS} & \textbf{ICS(2.0)} & \textbf{ICS(5.0)} & \textbf{ICS(10.0)}
      & \textbf{POW} & \textbf{BCS} & \textbf{ICS(2.0)} & \textbf{ICS(5.0)} & \textbf{ICS(10.0)} \\
    \hline
    \textbf{Standalone} & 88.09$\pm$0.06 & 96.06$\pm$0.16 & 96.93$\pm$0.16 & 96.80$\pm$0.05 & 96.94$\pm$0.02
                       & 54.86$\pm$1.93 & 54.53$\pm$0.66 & 58.82$\pm$1.19 & 59.95$\pm$1.29 & 56.20$\pm$1.71 \\
    \textbf{FedAvg}    & 94.47$\pm$0.46 & 97.50$\pm$0.14 & 94.57$\pm$0.32 & 96.23$\pm$0.16 & 94.32$\pm$0.13
                       & 67.08$\pm$0.17 & 65.80$\pm$0.65 & 66.08$\pm$0.15 & 64.68$\pm$0.60 & 66.67$\pm$0.95 \\
    \hline
    \textbf{CFFL}      & 94.66$\pm$0.42 & 99.61$\pm$0.16 & 97.41$\pm$2.44 & 98.18$\pm$1.59 & 98.63$\pm$1.19
                       & 67.76$\pm$1.87 & 70.71$\pm$3.34 & 71.37$\pm$0.07 & 64.34$\pm$4.98 & 69.27$\pm$2.03 \\
    \textbf{CGSV}      & 96.33$\pm$0.31 & 98.64$\pm$0.11 & 98.29$\pm$0.52 & 98.51$\pm$0.25 & 97.97$\pm$0.09
                       & 76.16$\pm$2.13 & 76.31$\pm$0.26 & 76.04$\pm$2.28 & 72.29$\pm$2.05 & 78.40$\pm$1.37 \\
    \textbf{FedAVE}    & 92.42$\pm$0.93 & 98.72$\pm$0.75 & 96.44$\pm$0.76 & 95.46$\pm$1.83 & 96.78$\pm$0.14
                       & 62.39$\pm$1.56 & 58.27$\pm$0.09 & 60.53$\pm$3.10 & 60.79$\pm$0.86 & 60.96$\pm$1.26 \\
    \textbf{FedSAC}    & 96.42$\pm$0.56 & 96.73$\pm$0.45 & 95.33$\pm$0.64 & 97.31$\pm$1.34 & 96.34$\pm$0.33
                       & 65.27$\pm$0.22 & 63.04$\pm$2.57 & 65.42$\pm$0.06 & 64.13$\pm$0.84 & 65.59$\pm$0.08 \\
    \hline
    \textbf{pFedCK}    & 94.15$\pm$0.18 & 97.52$\pm$0.37 & 97.11$\pm$0.15 & 98.62$\pm$0.54 & 98.25$\pm$0.87
                       & 80.13$\pm$0.87 & 79.54$\pm$0.31 & 80.88$\pm$1.24 & 79.99$\pm$1.46 & 80.71$\pm$0.98 \\
    \textbf{FedDC}     & 91.71$\pm$0.08 & 96.22$\pm$0.40 & 94.13$\pm$0.42 & 94.45$\pm$0.39 & 94.37$\pm$0.13
                       & 67.10$\pm$1.06 & 64.67$\pm$0.25 & 66.11$\pm$0.64 & 65.63$\pm$1.34 & 65.81$\pm$0.65 \\
    \textbf{FedAS}     & 88.20$\pm$0.38 & 98.89$\pm$0.21 & 96.08$\pm$0.23 & 96.76$\pm$0.20 & 96.37$\pm$0.33
                       & 75.56$\pm$1.23 & 73.13$\pm$0.66 & 74.59$\pm$0.79 & 74.25$\pm$1.27 & 75.58$\pm$1.61 \\
    \textbf{FedMPR}    & 94.80$\pm$0.23 & 97.72$\pm$0.21 & 95.20$\pm$0.67 & 95.93$\pm$0.33 & 95.88$\pm$0.43
                       & 69.06$\pm$0.36 & 66.40$\pm$0.57 & 66.03$\pm$0.54 & 67.84$\pm$1.43 & 69.03$\pm$0.80 \\
    \hline
    \rowcolor{greyL}
    \textbf{\ours} & \textbf{97.45}$\pm$0.09 & \textbf{99.67}$\pm$0.14 & \textbf{99.56}$\pm$0.04 & \textbf{99.66}$\pm$0.04 & \textbf{99.56}$\pm$0.11
                   & \textbf{83.80}$\pm$1.12 & \textbf{81.27}$\pm$0.19 & \textbf{81.93}$\pm$1.03 & \textbf{82.73}$\pm$0.38 & \textbf{81.30}$\pm$1.43 \\
    \hline
  \end{tabular}}
  \bigskip

  \subcaptionbox{Average Client Accuracy (\%)}{\label{tab:avg-acc}
  \begin{tabular}{|l|c|c|c|c|c|c|c|c|c|c|}
    \hline
    \multirow{2}{*}{\textbf{Method}} &
    \multicolumn{5}{c|}{\textbf{FashionMNIST}} &
    \multicolumn{5}{c|}{\textbf{CIFAR10}} \\
    \cline{2-11}
      & \textbf{POW} & \textbf{BCS} & \textbf{ICS(2.0)} & \textbf{ICS(5.0)} & \textbf{ICS(10.0)}
      & \textbf{POW} & \textbf{BCS} & \textbf{ICS(2.0)} & \textbf{ICS(5.0)} & \textbf{ICS(10.0)} \\
    \hline
    \textbf{Standalone} & 86.26$\pm$0.06 & 92.04$\pm$0.03 & 89.15$\pm$0.19 & 89.50$\pm$0.05 & 90.01$\pm$0.13
                       & 45.44$\pm$0.13 & 50.03$\pm$0.45 & 47.25$\pm$0.38 & 47.20$\pm$0.96 & 46.33$\pm$0.75 \\
    \textbf{FedAvg}    & 91.16$\pm$0.24 & 92.98$\pm$0.06 & 91.60$\pm$0.13 & 91.98$\pm$0.08 & 91.28$\pm$0.03
                       & 63.95$\pm$1.00 & 62.37$\pm$0.66 & 63.10$\pm$0.16 & 62.03$\pm$0.55 & 62.14$\pm$0.69 \\
    \hline
    \textbf{CFFL}      & 87.79$\pm$0.91 & 93.92$\pm$0.04 & 89.43$\pm$0.56 & 88.39$\pm$0.83 & 87.36$\pm$0.36
                       & 53.91$\pm$3.31 & 62.88$\pm$4.27 & 46.14$\pm$2.92 & 48.27$\pm$3.29 & 54.18$\pm$2.71 \\
    \textbf{CGSV}      & 91.23$\pm$0.15 & 92.21$\pm$0.35 & 90.52$\pm$1.35 & 89.32$\pm$0.36 & 91.14$\pm$0.15
                       & 66.68$\pm$1.50 & 67.00$\pm$2.93 & 64.32$\pm$2.32 & 63.65$\pm$2.11 & 65.56$\pm$1.35 \\
    \textbf{FedAVE}    & 89.44$\pm$0.45 & 92.36$\pm$0.19 & 89.83$\pm$0.47 & 88.99$\pm$0.64 & 89.66$\pm$0.04
                       & 55.85$\pm$1.44 & 52.19$\pm$1.19 & 55.32$\pm$2.09 & 51.85$\pm$2.41 & 53.42$\pm$1.84 \\
    \textbf{FedSAC}    & 89.54$\pm$0.63 & 88.92$\pm$0.83 & 89.75$\pm$0.92 & 88.65$\pm$0.54 & 89.76$\pm$0.63
                       & 54.63$\pm$0.32 & 60.53$\pm$0.81 & 60.52$\pm$0.29 & 57.52$\pm$0.72 & 60.84$\pm$0.64 \\
    \hline
    \textbf{pFedCK}    & 90.83$\pm$0.18 & 93.63$\pm$0.28 & 91.82$\pm$0.28 & 91.04$\pm$0.77 & 91.98$\pm$0.25
                       & 67.51$\pm$0.52 & 64.18$\pm$0.84 & 65.19$\pm$0.27 & 65.65$\pm$0.71 & 65.88$\pm$0.39 \\
    \textbf{FedDC}     & 87.83$\pm$0.46 & 91.08$\pm$1.46 & 90.06$\pm$0.56 & 90.20$\pm$0.24 & 90.00$\pm$0.42
                       & 55.34$\pm$0.65 & 53.13$\pm$1.31 & 52.98$\pm$0.05 & 52.30$\pm$1.73 & 55.83$\pm$0.73 \\
    \textbf{FedAS}     & 84.36$\pm$0.22 & 91.47$\pm$0.23 & 88.27$\pm$0.40 & 87.84$\pm$0.27 & 88.22$\pm$0.24
                       & 64.64$\pm$0.75 & 64.68$\pm$0.30 & 63.75$\pm$0.60 & 64.39$\pm$0.32 & 63.81$\pm$0.54 \\
    \textbf{FedMPR}    & 91.13$\pm$0.04 & 92.90$\pm$0.07 & 91.74$\pm$0.05 & 91.73$\pm$0.13 & 91.71$\pm$0.19
                       & 64.95$\pm$0.56 & 62.79$\pm$0.48 & 63.77$\pm$0.06 & 63.20$\pm$0.58 & 64.67$\pm$0.07 \\
    \hline
    \rowcolor{greyL}
    \textbf{\ours} & \textbf{93.50}$\pm$0.19 & \textbf{95.68}$\pm$0.15 & \textbf{92.62}$\pm$0.20 & \textbf{92.47}$\pm$0.14 & \textbf{92.92}$\pm$0.12
                   & \textbf{70.96}$\pm$0.58 & \textbf{67.59}$\pm$0.49 & \textbf{68.96}$\pm$0.31 & \textbf{68.50}$\pm$0.42 & \textbf{68.55}$\pm$1.35 \\
    \hline
  \end{tabular}}
  \bigskip

  \subcaptionbox{Collaborative Fairness (CF) Coefficient}{\label{tab:cf-coef}
  \resizebox{0.95\textwidth}{!}{%
  \begin{tabular}{|l|c|c|c|c|c|c|c|c|c|c|}
    \hline
    \multirow{2}{*}{\textbf{Method}} &
    \multicolumn{5}{c|}{\textbf{FashionMNIST}} &
    \multicolumn{5}{c|}{\textbf{CIFAR10}} \\
    \cline{2-11}
      & \textbf{POW} & \textbf{BCS} & \textbf{ICS(2.0)} & \textbf{ICS(5.0)} & \textbf{ICS(10.0)}
      & \textbf{POW} & \textbf{BCS} & \textbf{ICS(2.0)} & \textbf{ICS(5.0)} & \textbf{ICS(10.0)} \\
    \hline
    \textbf{FedAvg}  & 45.45$\pm$3.60 & 62.43$\pm$1.11 & 71.76$\pm$2.14 & 87.60$\pm$4.47 & 74.43$\pm$2.37
                     & 3.62$\pm$25.88 & 26.00$\pm$9.58 & 67.60$\pm$6.97 & 33.97$\pm$27.58 & 74.79$\pm$3.52 \\
    \hline
    \textbf{CFFL}    & 33.85$\pm$15.44 & 79.80$\pm$10.74 & 60.80$\pm$20.34 & 74.22$\pm$6.33 & 67.01$\pm$8.69
                     & 45.90$\pm$31.33 & 16.94$\pm$32.09 & 55.03$\pm$15.49 & 13.17$\pm$7.51 & 26.82$\pm$11.03 \\
    \textbf{CGSV}    & 38.11$\pm$14.34 & 69.03$\pm$6.63 & 52.58$\pm$7.38 & 57.32$\pm$11.69 & 68.46$\pm$16.67
                     & 36.25$\pm$26.04 & 45.87$\pm$19.76 & 18.50$\pm$21.40 & -4.15$\pm$18.53 & 5.54$\pm$22.85 \\
    \textbf{FedAVE}  & 21.43$\pm$23.31 & 65.10$\pm$1.47 & 78.15$\pm$8.32 & 80.02$\pm$7.00 & 65.79$\pm$4.98
                     & 16.60$\pm$18.27 & 49.08$\pm$15.03 & 17.07$\pm$22.03 & 24.56$\pm$34.43 & 25.51$\pm$27.50 \\
    \textbf{FedSAC}  & 28.10$\pm$7.93 & 79.54$\pm$5.13 & 48.78$\pm$14.97 & 83.30$\pm$5.13 & 71.26$\pm$4.76
                     & 62.04$\pm$12.03 & 43.61$\pm$10.38 & 56.89$\pm$7.30 & 23.90$\pm$12.70 & 20.71$\pm$19.95 \\
    \hline
    \textbf{pFedCK}  & 23.15$\pm$5.26 & 41.09$\pm$13.56 & 34.15$\pm$9.36 & 36.65$\pm$8.61 & 15.27$\pm$15.31
                     & 51.14$\pm$9.15 & 70.78$\pm$11.15 & 24.64$\pm$21.54 & 8.76$\pm$19.93 & 36.83$\pm$10.66 \\
    \textbf{FedDC}   & -40.85$\pm$18.38 & -11.12$\pm$22.49 & 2.39$\pm$15.37 & 18.37$\pm$9.41 & -15.32$\pm$38.39
                     & -4.22$\pm$32.08 & -9.54$\pm$31.25 & -21.57$\pm$5.97 & -9.38$\pm$27.68 & -0.52$\pm$42.33 \\
    \textbf{FedAS}   & 45.09$\pm$2.61 & 75.81$\pm$4.81 & 27.74$\pm$4.12 & 79.61$\pm$3.11 & 70.23$\pm$3.12
                     & 61.34$\pm$6.55 & 74.28$\pm$9.43 & 70.01$\pm$6.07 & 60.54$\pm$4.05 & 76.54$\pm$4.10 \\
    \textbf{FedMPR}  & 31.33$\pm$14.55 & 83.20$\pm$2.59 & 70.43$\pm$2.21 & 76.66$\pm$5.21 & 61.59$\pm$3.03
                     & 50.93$\pm$14.09 & 40.63$\pm$4.25 & 29.78$\pm$18.15 & 44.96$\pm$6.39 & 53.06$\pm$9.69 \\
    \hline
    \rowcolor{greyL}
    \textbf{\ours} & \textbf{70.61}$\pm$5.82 & \textbf{86.88}$\pm$0.81 & \textbf{78.25}$\pm$2.02 & \textbf{89.51}$\pm$2.40 & \textbf{79.72}$\pm$1.47
                   & \textbf{88.02}$\pm$2.09 & \textbf{82.15}$\pm$2.23 & \textbf{81.25}$\pm$1.55 & \textbf{82.53}$\pm$2.42 & \textbf{84.17}$\pm$2.77 \\
    \hline
  \end{tabular}}}
\end{table*}

\section{Simulation Experiments}
\label{sec:5}
\subsection{Non-IID Settings}
\label{sec:no-iid}
Since most real-world federated applications involve non-IID data distributions, particularly in the imbalanced covariate shift setting, we consider the following three non-IID client partitions in our simulation experiments:
\begin{itemize}[leftmargin=1em]
    \item \textbf{Imbalanced Dataset Sizes (POW) \cite{xu2021gradient,Lyu2020CollaborativeFairness}:}
    Each client’s dataset size \(|\mathcal{D}_k|\) follows a power-law distribution, leading to significant disparities in data quantity across clients. However, the feature distributions remain similar across clients.
    
    \item \textbf{Balanced Covariate Shift (BCS):}
    Clients exhibit covariate shifts in their data feature distributions while maintaining a similar number of samples.      
    \item \textbf{Imbalanced Covariate Shift (ICS):} ICS combines the characteristics of POW and BCS, where client dataset sizes follow a power-law distribution (POW), and different institutions experience significant variations in feature distributions.
\end{itemize}
Additionally, we evaluate the proposed \ours framework under two ideal yet commonly used \textbf{label shift} settings: \emph{Imbalanced Class Distributions (CLA)}~\cite{xu2021gradient,Lyu2020CollaborativeFairness} and \emph{Imbalanced Sizes \& Class Distributions (DIR)}~\cite{g,yurochkin2019bayesian}. The experimental setting and results for these two non-IID partitions are presented in \textbf{Appendix~\ref{sec:appendix-exp1}}.

\subsection{Federated Data Simulation}
In the simulation experiments, we use two image classification datasets: \textbf{Fashion MNIST}~\cite{xiao2017fashion} and 
\textbf{CIFAR10}~\cite{krizhevsky2009learning}. 
\textbf{Fashion MNIST} contains 70,000 grayscale images 
(\(28\times28\)) evenly split into 10 classes (e.g., T-shirt/top, trousers). \textbf{CIFAR10} consists of 60,000 color images (\(32\times32\)) across 10 classes (e.g., airplane, bird). 
We partition the datasets into training, validation, and testing in a ratio of 7:1:2. In addition, we set the number of clients as $K=10$. 
To simulate the \textbf{POW} partition, we follow a power law with an exponent of $1$ to divide the global data into 10 clients. For the $k$-th client, its data size is $|\mathcal{D}_k|=\frac{1}{kZ}|\mathcal{D}_g|$, where $Z = \sum_{k=1}^{10} \frac{1}{k}$. 

Simulating the \textbf{covariate shift} setting is nontrivial, and as far as we know, no existing work provides an automated way to generate such federated datasets. To fill this research gap, we design a novel algorithm based on Theorem~\ref{theorem2} for covariate shift data generation, as shown in \textbf{Appendix~\ref{sec:appendix-single-gauss}} Algorithm~\ref{alg:gaussian-covariate-shift}.
To stimulate the \textbf{BCS} partition, we set $C=5$ and add a perturbation $\delta$ (satisfying $\delta^\top\,\Sigma^{-1}\,\delta = 5$) to the global mean. In such a way, each client receives an equal number of samples (i.e., balanced), thus focusing on the covariate shift while keeping dataset sizes uniform.
To stimulate the \textbf{ICS} partition, we run Algorithm~\ref{alg:gaussian-covariate-shift} again to produce three variants with $C = 2$, $C = 5$, and $C = 10$, representing increasing levels of covariate shift. Meanwhile, the total number of samples is partitioned among the 10 clients according to the power law distribution with the exponent as $1$, thus coupling imbalanced dataset sizes with covariate shifts.

\subsection{Baselines}
We compare our method against two categories of baselines: 
\textit{Collaborative Fairness} algorithms designed for non-IID data, 
and \textit{Covariate Shift} algorithms focusing on feature-level discrepancies. 
Specifically, we include \textbf{CGSV}~\cite{xu2021gradient}, 
\textbf{CFFL}~\cite{Lyu2020CollaborativeFairness}, 
\textbf{FedAVE}~\cite{wang2024fedave}, 
and \textbf{FedSAC}~\cite{wang2024fedsac}, 
which explicitly address fairness by measuring client contributions or customizing reward allocations. 
Meanwhile, \textbf{FedAS}~\cite{yang2024fedas}, 
\textbf{FedDC}~\cite{g}, 
\textbf{pFedCK}~\cite{zhang2024pFedCK}, 
and \textbf{FedMPR}~\cite{goksu2024robust} focus on mitigating feature-level drift across clients. 
We also include two traditional baselines: 
\textbf{Standalone} (each client trains independently without aggregation) 
and the classic \textbf{FedAvg}~\cite{mcmahan2017communication} for federated averaging.
The details of the baselines can be found in \textbf{Appendix~\ref{sec:appendix-baselines}}.


\subsection{Implementation}
\label{sec:exp-implementation}
We implement a network consisting of two convolutional 
layers (each followed by batch normalization and ReLU activation), interleaved with max-pooling, 
and ending with a fully connected output layer for the simulation evaluation. 
We implement all baselines and our model in PyTorch and train them on an NVIDIA RTX A6000 GPU. 
All details of parameter setting can be found in \textbf{Appendix~\ref{sec:appendix-exp2}}.

\subsection{Evaluation Metrics}
Due to the imbalanced covariate shift setting, each client data has a unique distribution. Thus, we conduct local evaluations and then report the average values of all the clients for \textbf{three runs}. 
Let $\textbf{Acc}_p$ denote the prediction accuracy of clients after federation. Following~\cite{wang2024fedsac}, we use three metrics: 
\begin{itemize}[leftmargin=*]
    \item \textit{Average Client Accuracy}, i.e., $\frac{\sum_{k}\textbf{Acc}_p[k]}{K}$;
    \item \textit{Maximum Client Accuracy}, i.e, $\max(\textbf{Acc}_p)$;
    \item  \emph{Collaborative Fairness (CF) Coefficient}, which reflects how uniformly performance is distributed across clients. We use the CF defined in~\cite{Lyu2020CollaborativeFairness}, i.e, $CF= 100 \times \rho\bigl(\textbf{Acc}_s,\textbf{Acc}_p\bigr) \,\in\, [-100,\, 100]$, where \(\rho(\cdot,\cdot)\) is Pearson’s correlation coefficient, and $\textbf{Acc}_s$ represents the standalone accuracy of clients.
\end{itemize}
{The greater these three metric values, the better the performance.}

\subsection{Results of Simulation Experiments}
Table~\ref{tab:exp2_res} presents the results on FashionMNIST and CIFAR10 under three types of non-IID partitions. As shown in Table~\ref{tab:exp2_res}(a), our method consistently achieves superior or highly competitive \emph{Max Client Accuracy} across all partitions for both datasets. Table~\ref{tab:exp2_res}(b) further highlights our approach's advantage in \emph{Avg Client Accuracy}, particularly on the more challenging CIFAR10 dataset. The most critical metric, \textit{Collaborative Fairness} (CF), is reported in Table~\ref{tab:exp2_res}(c), where \ours demonstrates significantly higher fairness under varying degrees of non-IID settings. These results collectively validate the effectiveness of \ours in improving both accuracy and fairness.


\begin{table}[t]
\scriptsize   
\centering
\caption{Experimental results on the EHR dataset.}
\label{tab:exp3_res}
\vspace{-0.15in}
\begin{tabular}{|l|c|c|c|}
\hline
\textbf{Method} &\textbf{CF}& \textbf{ Max Acc} & \textbf{ Avg. Acc}  \\
\hline
\textbf{Standalone}  & -- &  76.27$\pm$0.13& 70.41$\pm$0.04   \\ 
\textbf{FedAvg}  & -12.63$\pm$5.22 &  74.75$\pm$1.24& 70.13$\pm$0.21 \\ 
\hline
\textbf{CFFL}       & 52.63$\pm$12.43 & 73.26$\pm$2.49&69.57$\pm$0.92 \\ 
\textbf{CGSV}        & 46.39$\pm$4.24 & 72.55$\pm$1.71 &69.31$\pm$0.42 \\
\textbf{FedAVE}      & 38.24$\pm$17.24 & 75.66$\pm$1.59 &69.24$\pm$0.81 \\ 
\textbf{FedSAC}      & 70.54$\pm$1.24 & 74.21$\pm$0.32 &68.33$\pm$ 0.56 \\ 
\hline
\textbf{pFedCK}     & 35.28$\pm$13.74 & 76.87$\pm$0.18 & 70.01$\pm$0.10  \\ 
\textbf{FedDC}        & 17.10$\pm$3.21 & 75.08$\pm$0.32& 
 69.61$\pm$0.29 \\
\textbf{FedAS}     & 8.50$\pm$3.28 &74.89 $\pm$0.29 & 69.05$\pm$0.25  \\ 
\textbf{FedMPR}     & -5.59$\pm$2.31 & 76.39$\pm$0.25 & 70.10$\pm$0.13  \\ 
\hline
\rowcolor{greyL}
\textbf{\ours}  & \textbf{78.42}$\pm$1.09 & \textbf{78.98}$\pm$1.01& \textbf{71.23}$\pm$0.27 \\ \hline
\end{tabular}

\end{table}
\begin{table}[t]
\scriptsize   
\centering
\caption{Performance of all methods when constrained to the same wall-clock time as \textbf{FedAKD} (40 rounds).}
\label{tab:equal_runtime}
\vspace{-0.15in}
\begin{tabular}{|l|c|c|c|c|}
\hline
\textbf{Method} & \textbf{Round} & \textbf{CF} & \textbf{Max Acc} & \textbf{Avg. Acc} \\
\hline
\textbf{FedAvg}   & 80 &  -12.63$\pm$5.22 & 74.75$\pm$1.24 & 70.13$\pm$0.21 \\
\textbf{CFFL}     & 43 &   48.63$\pm$11.43 & 75.25$\pm$1.42 & 67.31$\pm$0.95 \\
\textbf{CGSV}     & 61 &   33.11$\pm$6.31 & 73.62$\pm$0.55 & 68.84$\pm$0.84 \\
\textbf{FedAVE}   & 39 &   58.62$\pm$9.35 & 70.35$\pm$1.24 & 66.52$\pm$0.99 \\
\textbf{FedSAC}   & 65 &   60.22$\pm$3.21 & 74.02$\pm$0.44 & 68.89$\pm$0.21 \\
\textbf{pFedCK}  &58 & 33.51$\pm$8.51 & 75.01$\pm$0.43 & 69.11$\pm$0.30  \\ 
\textbf{FedDC}    & 51 &   30.10$\pm$5.32 & 73.54$\pm$1.35 & 68.53$\pm$0.88 \\
\textbf{FedAS} & 59 & 15.89$\pm$4.09 & 72.88$\pm$1.09 & 68.66$\pm$0.11  \\ 
\textbf{FedMPR}   & 47 &    8.22$\pm$4.24 & 74.39$\pm$0.21 & 69.28$\pm$0.26 \\
\hline
\rowcolor{greyL}
\textbf{FedAKD}   & 40 &   \textbf{71.89}$\pm$1.82 & \textbf{77.10}$\pm$0.58 & \textbf{71.02}$\pm$0.53 \\
\hline
\end{tabular}
\end{table}

\begin{table}[t]
\scriptsize  
\centering
\caption{Ablation study on three variants of \ours versus the full method. 
\textbf{CF} denotes a fairness metric among clients (higher is better),
whereas \textbf{Max Acc} and \textbf{Avg. Acc} refer to the maximum and average accuracies (in \%), respectively, across global rounds.}
\label{tab:ablation}
\vspace{-0.15in}
\begin{tabular}{|l|c|c|c|}
\hline
\textbf{Method} & \textbf{CF} & \textbf{Max Acc} & \textbf{Avg. Acc} \\
\hline
\ours(All-Data)    &  66.02$\pm$2.93   &  75.20$\pm$0.82    & 70.66$\pm$0.42 \\
\ours(Single-Dist) &  58.49$\pm$4.32   &  73.88$\pm$1.61    & 64.02$\pm$2.82 \\
\ours(Correct-Agg.) &  77.31$\pm$1.98   &  78.22$\pm$0.91    & 71.17$\pm$0.66 \\
\rowcolor{greyL}
\ours(Full)        & \textbf{78.42}$\pm$1.09    & \textbf{78.98}$\pm$1.01     & \textbf{71.23}$\pm$0.27 \\
\hline
\end{tabular}

\end{table}

\section{Real-World Experiments}
\subsection{Experimental Settings}\label{sec:exp3-analysis}
The EHR Dataset is extracted from the TriNetX database\footnote{\url{https://trinetx.com/}}, which contains patients' claims data from all 50 states in the USA. This dataset is curated for the early prediction of pancreatic cancer and includes 259,480 \textit{de-identified} patient records (161,345 negative vs. 98,135 positive). It consists of both static features (\textit{sex}, \textit{zip code}) and time-series events (\textit{medications}, \textit{lab tests}, \textit{vital signs}, etc.). Further details on the EHR dataset and experimental settings are provided in \textbf{Appendix~\ref{sec:appendix-ehr}}.
In this experiment, we adopt a two-layer bidirectional GRU with attention mechanisms~\cite{ma2017dipole,wickramaratne2020sepsis,zhang2018patient2vec,yang2019disease,ye2020predicting,tan2020data,guo2019crossover} to predict whether a patient eventually develops pancreatic cancer. We use the same baselines and evaluation metrics as the simulation experiments.

\subsection{Performance Analysis}
Table~\ref{tab:exp3_res} presents the results on the real-world EHR dataset, which naturally follows an \textbf{ICS} (Imbalanced Covariate Shift) pattern due to varying sample sizes and feature distributions across different medical institutions. As shown in the table, \ours significantly outperforms the baselines, particularly in the CF metric, demonstrating a more equitable distribution of benefits among clients. While FedMPR achieve competitive Max Acc, their fairness metrics remain comparatively limited.

\begin{figure}[t]
  \centering
  \begin{subfigure}[t]{0.49\linewidth}
    \centering
    \includegraphics[width=\linewidth]{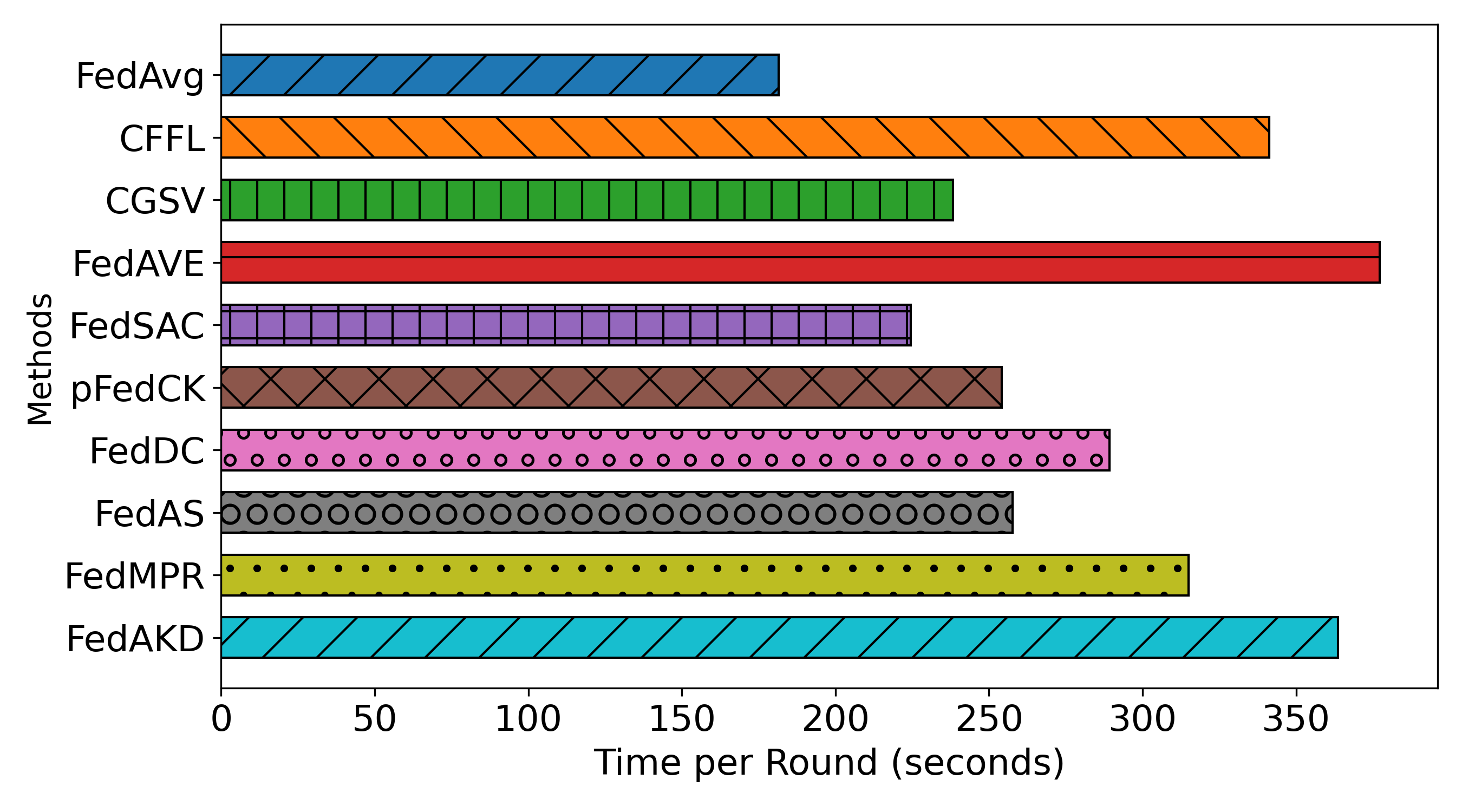}
    \caption{Average per-round runtime.}
    \label{fig:cost}
  \end{subfigure}
  \hfill
  \begin{subfigure}[t]{0.49\linewidth}
    \centering
    \includegraphics[width=\linewidth]{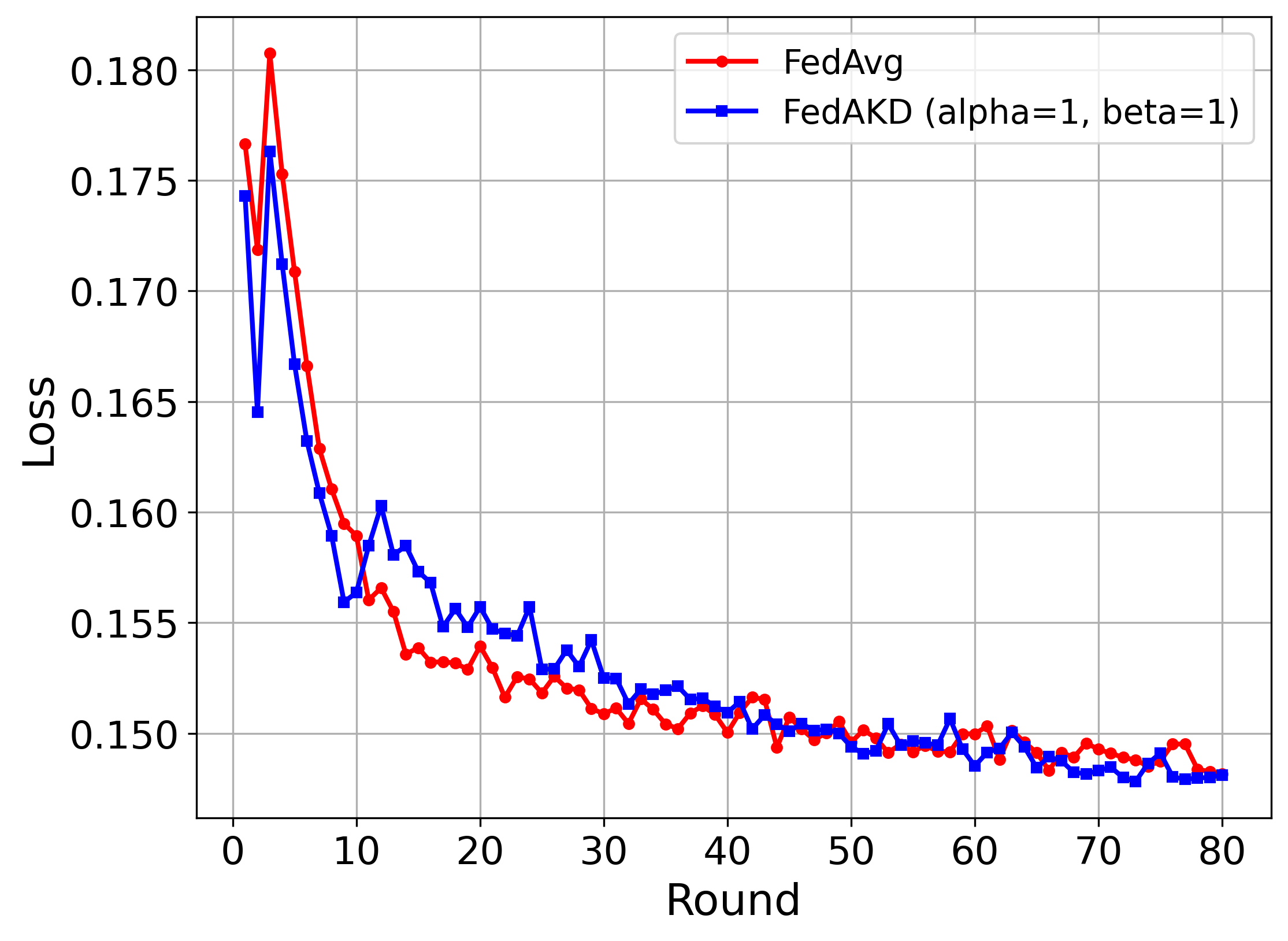}
    \caption{Training loss on global dataset $\mathcal{D}_g$.}
    \label{fig:loss}
  \end{subfigure}

  \vspace{-0.15in}  
  \caption{Efficiency comparison of \ours{} and baselines.}
  \label{fig:cost-loss}
  \vspace{-0.15in}  
\end{figure}

\subsection{Ablation Studies}

To investigate how each component of the proposed \ours (Algorithm~\ref{alg:fedxx}) contributes to its final performance, we conduct three ablation experiments. In each variant, we selectively remove or modify part of the procedure to gauge its impact:
\textbf{(1) Local $\to$ Global Distillation Using All Data $\mathcal{D}_k$.}
The biggest key finding of this work is to use the correctly predicted subset $\mathcal{I}_k^{t}$ to guide the update of $\mathbf{w}_{g,k}^{t}$. This baseline uses \emph{all} local samples $\mathcal{D}_k$ for local $\to$ global distillation to test whether restricting to accurately labeled data is necessary for effective knowledge distillation.
\textbf{(2) Only Local $\to$ Global Distillation.}
After receiving the aggregated global model, each client simply overwrites its local model with $\mathbf{w}_g^{t}$, instead of performing an additional global $\to$ local distillation. This variant helps isolate the effect of double-direction distillation.
\textbf{(3) Correct Sample Count for Aggregation.}
In the standard procedure, the server weights each client’s update by the total local dataset size. In this ablation, we replace that term with the number of \emph{correctly predicted} local samples ($|\mathcal{I}_k^{t}|$), thus investigating whether ``correct sample counts'' lead to better aggregation fairness or performance.

The results of these ablation studies are summarized in Table~\ref{tab:ablation}. Our findings indicate that the proposed asynchronous knowledge distillation strategy has the most significant impact on both performance and fairness. Additionally, conducting knowledge distillation on the entire dataset negatively affects model performance. However, using the size of either the full dataset or only correctly classified data does not lead to notable performance differences. These results validate the rationale behind our model design.

\vspace{-0.05in}
\subsection{Computational Cost Analysis}
\label{subsec:cost_convergence}
In our model design, the client training contains three steps, while baselines also use other techniques to calculate rewards. To validate the efficiency of our model, we show the average per-round running times (in seconds) as depicted in Figure~\ref{fig:cost}. Although \ours shows a slightly higher time consumption per round, primarily due to asynchronous knowledge distillation, its computational overhead is still comparable to other baselines (e.g., \textbf{CFFL} and \textbf{FedAVE}, which involve additional validation or sparsification steps).To further evaluate compute efficiency, we fixed \textbf{FedAKD} to 40 rounds, recorded its total wall-clock time, and then allowed every baseline to train for the \emph{same} duration.  
Table~\ref{tab:equal_runtime} shows that—even under this strict budget—\ours{} achieves the largest fairness improvement and the highest predictive performance.


\vspace{-0.05in}
\subsection{Convergence Analysis}
\label{subsec:cost_convergence}
We next analyze the convergence behavior of these methods by evaluating the global parameter \(\mathbf{w}_g^t\) on the global EHR dataset, i.e., computing \(\mathcal{L}(\mathbf{w}_g^t; \mathcal{D}_g)\) at each round. Figure~\ref{fig:loss} shows two representative convergence curves, comparing \textbf{FedAvg} and \textbf{\ours}. We observe that \ours converges slightly more slowly in the early stages; however, it remains stable and ultimately achieves a low global loss. This result empirically verifies Theorem~\ref{thm:fedakd_convergence_full}, demonstrating that the global model in \ours converges effectively to a desirable minimum on \(\mathcal{D}_g\).

\bigskip

%% file: RelatedWork.tex
\vspace{-0.05in}
\section{Related Work}
\label{sec:related_work}
This work mainly focuses on \textbf{collaborative fairness (CF)} in federated learning, which regards the global model as the core reward and seeks to ensure that the final performance of each client reflects its actual contribution. For instance, 
{CGSV}~\cite{xu2021gradient} computes a cosine gradient Shapley value to measure how closely each client’s local gradient aligns with the global gradient, and allocates model updates based on this similarity.
{CFFL}~\cite{Lyu2020CollaborativeFairness} relies on a \emph{public validation set} to evaluate each client’s data diversity and local-model performance, then allocates rewards accordingly. 
{FedAVE}~\cite{wang2024fedave} computes reputation by examining each client’s local model performance and data distribution, offering better adaptability to various distributional scenarios.
{FedSAC}~\cite{wang2024fedsac} avoids the need for a global validation set by distributing varying submodels to high-contribution clients.
However, it depends on \emph{pre-known standalone training results}, which may be unrealistic in real-world deployments.
Moreover, submodel-based pruning alone cannot fully address feature-distribution mismatch, as high-quality but distribution-mismatched data may still be underrepresented.

%% file: Conclusion.tex
\section{Conclusion}
This paper investigates a practical yet challenging form of heterogeneity that impacts collaborative fairness: \textit{imbalanced covariate shift}. To address this issue, we propose a novel approach, \ours (Federated Asynchronous Knowledge Distillation), which mitigates the effects of imbalanced covariate shift by excluding incorrectly predicted samples from the global model update—an insight derived from our preliminary findings. Experimental results on three datasets compared against ten baselines demonstrate the effectiveness and fairness of \ours across various heterogeneity settings in federated learning.

%% file: Appendix.tex


\appendix  
\section{Imbalanced Covariate Shift Proof}
\label{sec:appedixa}
\subsection{Proof of Theorem~\ref{theorem1}}
\noindent
We first aim to expand
\begin{align*}
D_{\mathrm{KL}}\!\bigl(p_{\omega+\delta}\,\big\|\,p_{\omega}\bigr)
~:=~
\mathbb{E}_{X\sim p_{\omega+\delta}}\!\Bigl[
  \log p_{\omega+\delta}(X)
  ~-~
  \log p_{\omega}(X)
\Bigr].
\end{align*}
Observe that
\begin{align*}
\log p_{\omega}(x)
~=~
\log\bigl(p_{(\omega+\delta)-\delta}(x)\bigr)
~=~
\ell\bigl((\omega+\delta)-\delta,\,x\bigr).
\end{align*}
We make a second-order Taylor expansion of $\ell(\theta-\delta,x)$ 
around $\theta=\omega+\delta$ at $\delta=0$. 
Concretely, with $\ell(\theta,x) = \log p_{\theta}(x)$, we have
\begin{align*}
\ell\bigl((\omega+\delta)-\delta,\,x\bigr)
~\approx~&\;
\ell(\omega+\delta,x)
~-\;
\delta^\top\,
   \nabla_{\theta}\ell(\theta,x)\big\rvert_{\theta=\omega+\delta}\\
&~+\;
\frac{1}{2}\,\delta^\top\,
   \nabla_{\theta}^2\ell(\theta,x)\big\rvert_{\theta=\omega+\delta}
\,\delta,\notag
\end{align*}
\begin{align*}
\log p_{\omega+\delta}(x) \;-\;\log p_{\omega}(x)
~\approx~&
\delta^\top\,
\nabla_{\theta}\ell(\theta,x)\big\rvert_{\theta=\omega+\delta}\\
&-\frac12\,\delta^\top\,
\nabla_{\theta}^2\ell(\theta,x)\big\rvert_{\theta=\omega+\delta}
\,\delta.\notag
\end{align*}
We set 
\begin{align*}   
I(\omega+\delta) 
~:=~-\,\mathbb{E}_{p_{\omega+\delta}}\!\Bigl[\nabla_{\theta}^2 \log p_{\omega+\delta}(X)\Bigr].
\end{align*}
Taking expectation under $X\sim p_{\omega+\delta}$ gives
\begin{align}
D_{\mathrm{KL}}\!\bigl(p_{\omega+\delta}\,\big\|\,p_{\omega}\bigr)
=&\;
\mathbb{E}_{p_{\omega+\delta}}\!\Bigl[
  \log p_{\omega+\delta}(X) - \log p_{\omega}(X)
\Bigr]
\\
\approx&
\mathbb{E}_{p_{\omega+\delta}}\!\bigl[
  -
  \tfrac12\,\delta^\top
           \nabla_{\theta}^2\ell(\theta,X)\big\rvert_{\theta=\omega+\delta}\,\delta
\bigr]\\
=& \tfrac12\,\delta^\top I(\omega+\delta)\,\delta\\
\approx& \tfrac12\,\delta^\top I(\omega)\,\delta+\tfrac12\,\delta^\top \nabla_\omega I(\omega)\,\delta \delta\label{eq:w+delta||w}
\end{align}
We want to compute or approximate the following KL divergence:
\[
D_{\mathrm{KL}}\bigl(\widehat{p}_{\omega+\delta}\,\big\|\,p_{\omega+\delta}\bigr),
\]
where $\widehat{p}_{\omega+\delta}$ is the empirical distribution (or fitted model) from $A$ i.i.d.\ samples $X_1,\dots,X_A$ drawn from the ``true'' distribution $p_{\omega+\delta}$. 
To simplify notation, let us define:
\[
\theta^* 
~:=~
\omega+\delta,
\quad\text{and}\quad
\widehat{\theta} 
~\text{be the MLE fitted using $A$ samples from }p_{\theta^*}.
\]
Thus the distribution $\widehat{p}_{\omega+\delta}$ can be written as $p_{\widehat{\theta}}$ (i.e.\ parameterized by $\widehat{\theta}$). We often denote
\[
\Delta 
~:=~ 
\widehat{\theta} - \theta^* 
~=~
\widehat{\theta} - (\omega+\delta).
\]
Hence the divergence of interest is
\[
D_{\mathrm{KL}}\!\bigl(p_{\widehat{\theta}} \,\big\|\,p_{\theta^*}\bigr).
\]
Let $\ell(\theta,x) := \log p_{\theta}(x)$. Taylor expand $\log p_{\widehat{\theta}}(x)$ around $\theta^*$.
\begin{align*}
\ell(\widehat{\theta}, x) 
~=&~
\ell(\theta^*, x)
~+\;
(\widehat{\theta}-\theta^*)^\top
\nabla_{\theta}\ell(\theta^*, x)\\
&+\;
\tfrac12\,
(\widehat{\theta}-\theta^*)^\top
\nabla_{\theta}^2 \ell(\widetilde{\theta}, x)\,
(\widehat{\theta}-\theta^*)
\notag
\\
=&
\ell(\theta^*, x)
~+\;
\Delta^\top\nabla_{\theta}\ell(\theta^*, x)
~+\;
\tfrac12\,\Delta^\top
\nabla_{\theta}^2 \ell(\widetilde{\theta}, x)
\,\Delta,
\end{align*}
where $\widetilde{\theta}$ is between $\theta^*$ and $\widehat{\theta}$.
\noindent
\begin{align*}
\log p_{\widehat{\theta}}(x) 
~-~ 
\log p_{\theta^*}(x)
=&
(\widehat{\theta}-\theta^*)^\top
\nabla_{\theta}\ell(\theta^*, x)\\
&+\;
\tfrac12\,
(\widehat{\theta}-\theta^*)^\top
\nabla_{\theta}^2 \ell(\widetilde{\theta}, x)\,
(\widehat{\theta}-\theta^*).\notag
\end{align*}
\begin{align}
\mathbb{E}_{p_{\widehat{\theta}}} 
\bigl[
  \log p_{\widehat{\theta}}(X) 
  - \log p_{\theta^*}(X)
\bigr]
&=
\mathbb{E}_{p_{\widehat{\theta}}}\Bigl[
  \Delta^\top \nabla_{\theta}\ell(\theta^*,X)
  + 
  \tfrac12\,\Delta^\top
  \nabla_{\theta}^2 \ell(\widetilde{\theta}, X)
  \,\Delta
\Bigr].
\label{eq:KL-split}
\end{align}
We deal with the first-order term $\Delta^\top \nabla_{\theta}\ell(\theta^*,X)$.
Since $\Delta$ does not depend on $X$, we have
\begin{align*}
\Delta^\top
\,\mathbb{E}_{p_{\widehat{\theta}}}
\bigl[\nabla_{\theta}\ell(\theta^*,X)\bigr]
&\approx
\Delta^\top 
\,\mathbb{E}_{p_{\theta^*}}\bigl[\nabla_{\theta}\ell(\theta^*,X)\bigr]
=
\Delta^\top 
\,\mathbf{0}
~=~
0,
\end{align*}
where we used 
$\mathbb{E}_{p_{\theta^*}}[\nabla_{\theta}\ell(\theta^*,X)] = 0$ 
and $p_{\widehat{\theta}} \approx p_{\theta^*}$ for large $a$.\\
Consider the remaining piece in Eq.~\eqref{eq:KL-split}:
\begin{align*}
\tfrac12 
\,\mathbb{E}_{p_{\widehat{\theta}}}
\bigl[
 \Delta^\top
 \nabla_{\theta}^2 \ell(\widetilde{\theta}, X)\,
 \Delta
\bigr].
\end{align*}
When $\widehat{\theta}\approx \theta^*$, we have 
$\nabla_{\theta}^2 \ell(\widetilde{\theta},X)\approx \nabla_{\theta}^2 \ell(\theta^*,X)$,
and
\begin{align*}
\mathbb{E}_{X\sim p_{\theta^*}}\bigl[\nabla_{\theta}^2 \ell(\theta^*, X)\bigr]
= 
-\;I(\theta^*).
\end{align*}
Hence 
\begin{align*}
\mathbb{E}_{p_{\widehat{\theta}}}\bigl[
  \Delta^\top
  \nabla_{\theta}^2 \ell(\widetilde{\theta}, X)\,
  \Delta
\bigr]
~\approx~
-\;
\Delta^\top\,I(\theta^*)\,\Delta.
\end{align*}
Recall the asymptotic distribution:
\[
\Delta 
~=~ 
\widehat{\theta}-\theta^*
~\approx~
\frac{1}{\sqrt{A}}
\,\mathcal{N}\bigl(0,\;I(\theta^*)^{-1}\bigr).
\]
Thus
\begin{align*}
\mathbb{E}\bigl[\Delta^\top\,I(\theta^*)\,\Delta\bigr]
&=
\mathrm{trace}\Bigl[
  I(\theta^*)\,
  \mathbb{E}(\Delta\Delta^\top)
\Bigr]\\
&=
\mathrm{trace}\Bigl[
  I(\theta^*)\,
  \tfrac{1}{a}\,I(\theta^*)^{-1}
\Bigr]\\
&=
\tfrac{1}{A}\,\mathrm{trace}(I_R)\\
&=
\tfrac{R}{A}.
\end{align*}
Therefore, 
\begin{align*}
\tfrac12
\bigl[
  -\;\Delta^\top\,I(\theta^*)
  \,\Delta
\bigr]
&\approx
-\;\tfrac12\,\frac{R}{A}.
\end{align*}
So we conclude the derivation that
\begin{align}
    D_{\mathrm{KL}}\bigl(\widehat{p}_{\omega+\delta}\,\big\|\,p_{\omega+\delta}\bigr)
~\approx~
\frac{R}{2\,A}
\quad
\text{for large sample size $A$.}\label{eq:R/2A}
\end{align}
Due to Eq.~\eqref{eq:R/2A} and Eq.~\eqref{eq:w+delta||w}, we can get
\begin{align}
D_{\mathrm{KL}}\bigl(\widehat{p}_{\omega+\delta}\,\big\|\,p_{\omega}\bigr)
&~\approx~
\tfrac12\,\delta^\top I(\omega)\,\delta+\tfrac12\,\delta^\top \nabla_\omega I(\omega)\,\delta \delta+\frac{R}{2\,A}+e\\
&~\approx~
\tfrac12\,\delta^\top I(\omega)\,\delta+\tfrac12\,\delta^\top \nabla_\omega I(\omega)\,\delta \delta+\frac{R}{2\,A},
\end{align}
where $
e=
\int
\bigl[\widehat{p}_{\omega+\delta}(x) \;-\; p_{\omega+\delta}(x)\bigr]\,
\log\!\Bigl(\tfrac{p_{\omega}(x)}{p_{\omega+\delta}(x)}\Bigr)
\,\mathrm{d}x \approx 0.
$
\subsection{Proof of Theorem~\ref{theorem2}}
\noindent
For the family \(p_{\mu,\Sigma}(x) = \mathcal{N}(x;\,\mu,\Sigma)\), the Fisher information matrix
\(I(\mu,\Sigma)\) can be written in a block-diagonal form:
\begin{align*}
  I(\mu,\Sigma)
  &=
  \begin{pmatrix}
    \Sigma^{-1} & 0 \\
    0 & \tfrac12\,\bigl(\Sigma^{-1}\otimes\Sigma^{-1}\bigr)
  \end{pmatrix},
\end{align*}
where \(\otimes\) is the Kronecker product. Thus, if we set
\[
  \delta \;=\; \bigl(\delta_\mu,\;\vecDeltaSigma\bigr),
\]
\[
  I(\omega)
  \;=\;
  \begin{pmatrix}
    I_{\mu,\mu} & 0 \\
    0 & I_{\Sigma,\Sigma}
  \end{pmatrix}
  \;=\;
  \begin{pmatrix}
    \Sigma^{-1} & 0 \\
    0 & \tfrac12\,(\Sigma^{-1}\otimes\Sigma^{-1})
  \end{pmatrix},
\]

\medskip
\noindent
We decompose the following expression into three parts:

\begin{align*}
  \underbrace{\frac12\,\delta^\top I(\omega)\,\delta}_{\text{(\textcolor{red}{\textbf{A}})}}
  \;+\;
  \underbrace{\frac12\,\delta^\top \nabla_{\omega} I(\omega)\,\delta\,\delta}_{\text{(\textcolor{red}{\textbf{B}})}}
  \;+\;
  \underbrace{\frac{R}{2\,A}}_{\text{(\textcolor{red}{\textbf{C}})}} \label{eq:total}
\end{align*}
For the term \text{(\textcolor{red}{\textbf{A}})} \(= \tfrac12\,\delta^\top I(\omega)\,\delta\), since \(I(\omega)\) is block-diagonal in \((\mu,\Sigma)\), then we have
\begin{align}
  \tfrac12\,\delta^\top I(\omega)\,\delta
  &= \tfrac12\,(\delta_\mu)^\top \Sigma^{-1}\,\delta_\mu
  \;+\;
  \tfrac12\,\bigl(\vecDeltaSigma\bigr)^\top 
     \!\Bigl[\tfrac12\bigl(\Sigma^{-1}\otimes\Sigma^{-1}\bigr)\Bigr]\!
     \vecDeltaSigma
  \\[6pt]
  &= \tfrac12\,(\delta_\mu)^\top \Sigma^{-1}\,\delta_\mu
  \;+\;
  \tfrac14\,\bigl(\vecDeltaSigma\bigr)^\top 
     \bigl(\Sigma^{-1}\otimes\Sigma^{-1}\bigr)\!
     \vecDeltaSigma.
  \label{eq:deltaIomega-split}
\end{align}

\noindent Recall that \(\matDeltaSigma \in \mathbb{R}^{d\times d}\) is the matrix-form perturbation to \(\Sigma\), 
while \(\vecDeltaSigma = \mathrm{vec}(\matDeltaSigma)\in\mathbb{R}^{d^2}\) is its vectorization.
A standard identity for the Frobenius norm is
\[
  \bigl(\vecDeltaSigma\bigr)^\top \bigl(\Sigma^{-1}\otimes \Sigma^{-1}\bigr)\vecDeltaSigma
  \;=\;
  \|\Sigma^{-1}\,\matDeltaSigma\,\Sigma^{-1}\|_F^2.
\]
Therefore,
\begin{align}
  \tfrac12\,\delta^\top I(\omega)\,\delta
  &= \tfrac12\,(\delta_\mu)^\top \Sigma^{-1}\,\delta_\mu
  \;+\;
  \tfrac14\,\|\Sigma^{-1}\,\matDeltaSigma\,\Sigma^{-1}\|_F^2.
  \label{eq:a}
\end{align}

\medskip
\noindent For the term \text{(\textcolor{red}{\textbf{B}})}   $=\tfrac12\,\delta^\top (\nabla_{\omega} I(\omega)\,\delta\,)\delta$,
because $I(\mu,\Sigma)$ is block-diagonal, its derivative with respect to $(\mu,\Sigma)$ also splits 
into two blocks. Consequently, we can separate:
\[
  \nabla_{\omega} I(\omega) 
  \;=\;
  \begin{pmatrix}
    \nabla_{\omega}\bigl[\Sigma^{-1}\bigr] & 0 \\[3pt]
    0 & \nabla_{\omega}\bigl[\tfrac12(\Sigma^{-1}\otimes \Sigma^{-1})\bigr]
  \end{pmatrix}.
\]
\noindent \textbf{B1} denotes contribution from $(\mu,\mu)$ block: 
\begin{align*}
    B_1= \tfrac12\,\delta_\mu^\top 
    \Bigl[\nabla_\Sigma \bigl(\Sigma^{-1}\bigr)\!(\matDeltaSigma)\Bigr]\,
    \delta_\mu.
\end{align*}
  We have $I_{\mu,\mu}(\Sigma) = \Sigma^{-1}$.  Its derivative with respect 
  to $\Sigma$ (a matrix perturbation $\matDeltaSigma$) is
  \[
    \nabla_\Sigma \bigl[\Sigma^{-1}\bigr]\!(\matDeltaSigma)
    \;=\;
    -\,\Sigma^{-1}\,\matDeltaSigma\,\Sigma^{-1}.
  \]
  Inserting $\delta_\mu^\top$ and $\delta_\mu$ then yields
  \begin{align}
    B_1
    \;=\;
    -\,\tfrac12\,\delta_\mu^\top
      \Bigl(\Sigma^{-1}\,\matDeltaSigma\,\Sigma^{-1}\Bigr)\,
    \delta_\mu.\label{eq:b1}
  \end{align}
  This term vanishes if $\delta_\Sigma=0$ or if one chooses to ignore $\delta_\mu$, 
  but is otherwise a valid second-order effect.

\noindent \textbf{B2} denotes contribution from $(\Sigma,\Sigma)$ block:
\begin{align}
    B_2=\tfrac12\,\bigl(\vecDeltaSigma\bigr)^\top
  \Bigl[\nabla_{\Sigma} I_{\Sigma,\Sigma}(\Sigma)\bigl(\matDeltaSigma\bigr)\Bigr]
  \vecDeltaSigma
\end{align}
  As in the main text, we set 
  \[
    I_{\Sigma,\Sigma}(\Sigma) 
    \;=\;
    \tfrac12\,\bigl(\Sigma^{-1}\otimes \Sigma^{-1}\bigr).
  \]
Hence we define
\[
  \nabla_{\Sigma} I_{\Sigma,\Sigma}(\Sigma)
  \;=\;
  \tfrac12\,\nabla_{\Sigma}\bigl(\Sigma^{-1}\otimes\Sigma^{-1}\bigr).
\]
Suppose \(\matDeltaSigma \in \mathbb{R}^{d\times d}\) is a small matrix perturbation of \(\Sigma\), 
and let \(\vecDeltaSigma = \mathrm{vec}(\matDeltaSigma)\) be its vectorized form. 
Then the second-order expansion term of interest is
\begin{align}
  \tfrac12\,\delta^\top \nabla_{\omega}\bigl[\tfrac12(\Sigma^{-1}\otimes \Sigma^{-1})\bigr]\,\delta\,\delta
  \;=\;
  \tfrac12\,\bigl(\vecDeltaSigma\bigr)^\top
  \Bigl[\nabla_{\Sigma} I_{\Sigma,\Sigma}(\Sigma)\bigl(\matDeltaSigma\bigr)\Bigr]\,
  \vecDeltaSigma,
  \label{eq:HalfDeltaNablaI_vectorized}
\end{align}
where we note that \(\delta_\Sigma\) in the original notation is replaced by \(\matDeltaSigma\) in matrix form 
(or \(\vecDeltaSigma\) in vectorized form).

By a standard matrix-calculus result,
\[
  \delta\bigl(\Sigma^{-1}\otimes\Sigma^{-1}\bigr)
  \;=\;
  -\,\Sigma^{-1}\,\matDeltaSigma\,\Sigma^{-1}\,\otimes\,\Sigma^{-1}
  \;-\;
  \Sigma^{-1}\,\otimes\,\Sigma^{-1}\,\matDeltaSigma\,\Sigma^{-1},
\]
where \(\matDeltaSigma\) appears inside each product as the perturbation in matrix form. Hence,
\[
  \nabla_{\Sigma}\bigl(\Sigma^{-1}\otimes\Sigma^{-1}\bigr)\bigl(\matDeltaSigma\bigr)
  \;=\;
  -\,\Sigma^{-1}\,\matDeltaSigma\,\Sigma^{-1}\,\otimes\,\Sigma^{-1}
  \;-\;
  \Sigma^{-1}\,\otimes\,\Sigma^{-1}\,\matDeltaSigma\,\Sigma^{-1}.
\]
Since \(I_{\Sigma,\Sigma}(\Sigma)\) has the extra factor \(\tfrac12\), we obtain
\begin{align}
      \nabla_{\Sigma} I_{\Sigma,\Sigma}(\Sigma)\bigl(\matDeltaSigma\bigr)
  \;=\;
  \tfrac12
  \Bigl[
    -\,\Sigma^{-1}\,\matDeltaSigma\,\Sigma^{-1}\,\otimes\,\Sigma^{-1}
    \;-\;
    \Sigma^{-1}\,\otimes\,\Sigma^{-1}\,\matDeltaSigma\,\Sigma^{-1}
  \Bigr].\label{eq:Sigma}
\end{align}
\noindent
We take Eq.~\eqref{eq:Sigma} into Eq.~\eqref{eq:HalfDeltaNablaI_vectorized},
\begin{align}
  B_2
  =& \tfrac12\,\bigl(\vecDeltaSigma\bigr)^\top
  \Bigl[\nabla_{\Sigma} I_{\Sigma,\Sigma}(\Sigma)\bigl(\matDeltaSigma\bigr)\Bigr]
  \vecDeltaSigma
  \nonumber\\[6pt]
  =&\tfrac12\,\bigl(\vecDeltaSigma\bigr)^\top
  \Bigl[
     \tfrac12
     \Bigl(
        -\,\Sigma^{-1}\,\matDeltaSigma\,\Sigma^{-1}\,\otimes\,\Sigma^{-1}
        \;\\
        &-\;
        \Sigma^{-1}\,\otimes\,\Sigma^{-1}\,\matDeltaSigma\,\Sigma^{-1}
     \Bigr)
  \Bigr]
  \vecDeltaSigma
  \nonumber\\[6pt]
  =& \tfrac14\,\bigl(\vecDeltaSigma\bigr)^\top
  \Bigl[
    -\,\Sigma^{-1}\,\matDeltaSigma\,\Sigma^{-1}\,\otimes\,\Sigma^{-1}
    \;\\
    &-\;
    \Sigma^{-1}\,\otimes\,\Sigma^{-1}\,\matDeltaSigma\,\Sigma^{-1}
  \Bigr]
  \vecDeltaSigma
  \nonumber\\[6pt]
  =& -\,\tfrac14
  \bigl(\vecDeltaSigma\bigr)^\top
  \Bigl[
    \Sigma^{-1}\,\matDeltaSigma\,\Sigma^{-1}\,\otimes\,\Sigma^{-1}
    \;\\
    &+\;
    \Sigma^{-1}\,\otimes\,\Sigma^{-1}\,\matDeltaSigma\,\Sigma^{-1}
  \Bigr]
  \vecDeltaSigma.\\
    \stackrel{(a)}=&\;
  -\,\tfrac12\,
  \mathrm{trace}\!\Bigl(\matDeltaSigma\,\Sigma^{-1}\,\matDeltaSigma\,\Sigma^{-1}\,\matDeltaSigma\,\Sigma^{-1}\Bigr)\\
  =&-\,\tfrac12\,
  \mathrm{trace}\!\Bigl((\matDeltaSigma\,\Sigma^{-1})^3\Bigr),
  \label{eq:b2}
\end{align}
where $(a)$ is due to $(\mathrm{vec}(X))^\top (Y \otimes Z)\,(\mathrm{vec}(X)) 
= \mathrm{trace}[ZXYX^{\top}]$.

\noindent For (\textcolor{red}{\textbf{C}}),
since the 
M-dimensional Gaussian distribution, we can obtain
\begin{align*}
    R =&\underbrace{M}_{\text{Mean }}
+
\underbrace{\frac{M(M+1)}{2}}_{\text{Covariance}}\\
\;=&\;
M + \frac{M(M+1)}{2}\\
\;=&\;
\frac{M^2 + 3M}{2}.
\end{align*}
So we take this term into C,
\begin{align}
    C=\frac{R}{2\,A}= \frac{M(M+4)}{4A} \label{eq:c}
\end{align}
We take Eq.~\eqref{eq:c}, Eq.~\eqref{eq:b2}, Eq.~\eqref{eq:b1} and Eq.~\eqref{eq:a} into Eq.~\eqref{eq:total}: 
\begin{align*}
D_{\mathrm{KL}}\bigl(\widehat{p}_{\omega+\delta}\,\big\|\,p_{\omega}\bigr)
~\approx~&
    \tfrac12\,(\delta_\mu)^\top \Sigma^{-1}\,\delta_\mu
  \;+\;
  \tfrac14\,\|\Sigma^{-1}\,\matDeltaSigma\,\Sigma^{-1}\|_F^2+\frac{M(M+4)}{4A}\\
  &-\,\tfrac12\,\delta_\mu^\top
      \Bigl(\Sigma^{-1}\,\matDeltaSigma\,\Sigma^{-1}\Bigr)\,
    \delta_\mu
    -\,\tfrac12\,
  \mathrm{trace}\!\Bigl((\matDeltaSigma\,\Sigma^{-1})^3\Bigr)\\
  =&\tfrac12\,(\delta_\mu)^\top( \Sigma^{-1}(I-\matDeltaSigma\Sigma^{-1}))\,\delta_\mu+\;
  \tfrac14\,\|\Sigma^{-1}\,\matDeltaSigma\,\Sigma^{-1}\|_F^2\\
  &-\,\tfrac12\,
  \mathrm{trace}\!\Bigl((\matDeltaSigma\,\Sigma^{-1})^3\Bigr)+\frac{M(M+4)}{4A} 
\end{align*}


\section{Theoretical Approximation Validation}\label{sec:the-2}
\begin{figure}[!t]
    \centering
    \includegraphics[width=0.75\linewidth]{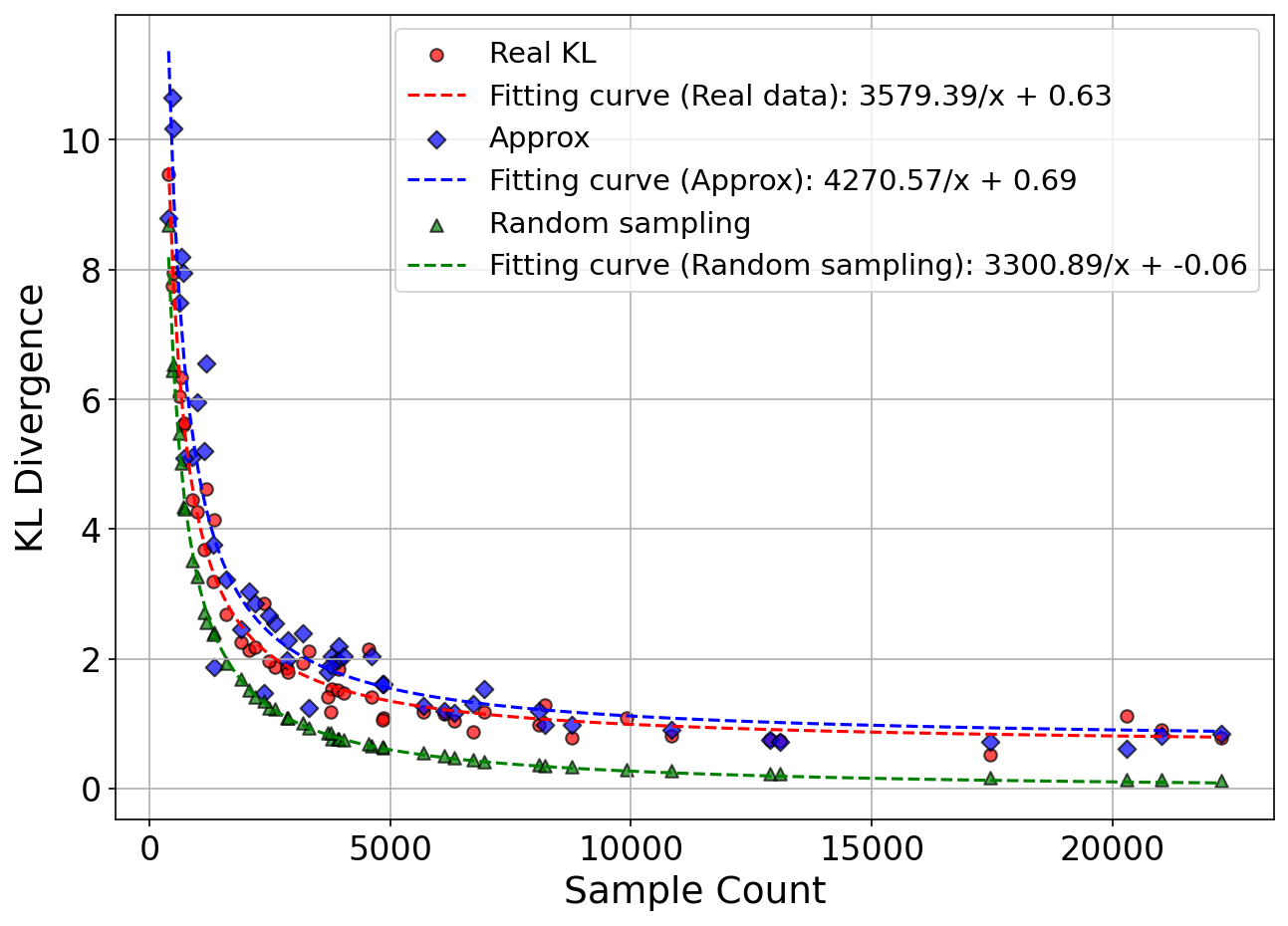}
    \caption{Comparison of KL divergences on real data (red), the Theorem~\ref{theorem2} approximation (blue), and a random-subsampling baseline (green). Our approximation aligns well with the actual KL divergence, while random subsampling underestimates the real shift.}
    \label{fig:vae_approx}
\end{figure}

To validate our theoretical approximations, we conduct experiments on the real EHR dataset by first training a Variational Autoencoder (VAE) to embed each client’s data into an \(M\)-dimensional latent space, where $M=100$. We then use these client representations to approximate a Gaussian distribution for each client. Similarly, we aggregate all client representations to fit a global Gaussian distribution. The results are shown in Figure~\ref{fig:vae_approx}.
In Figure~\ref{fig:vae_approx}, ``\textit{Real KL}'' means the empirical KL divergence between each client’s approximated distribution and the global fitted distribution. ``\textit{Approx}'' refers to our theoretical approximation, where small Gaussian perturbations are added to the mean vector and covariance matrix of each client’s approximated distribution, allowing the KL divergence to be computed directly via Eq.~\eqref{eq:gaussian_KL}. ``\textit{Random sampling}'' serves as a control experiment, where each client’s sample size is matched by randomly subsampling from the global distribution (i.e., without introducing any intentional distributional shift).
The results indicate that our theoretical approximation (\textcolor{blue}{blue curve}) closely follows the empirical KL values (\textcolor{red}{red curve}) across clients. Additionally, the random sampling baseline (\textcolor{OliveGreenRGB}{green curve}) remains significantly lower, confirming that the real dataset exhibits substantial covariate shift effects beyond simple sample-size imbalance. These empirical findings validate the correctness of our theorems and provide a principled basis for constructing simulation datasets to evaluate model performance under the new setting.

\section{FedAKD Convergence Proof}
\label{appendix:c}
\textbf{Global $\to$ Local KD.}
We compute the gradient of the distillation loss where the global model serves as the teacher 
and the local model serves as the student. The parameters are given by:
\[
\mathrm{KD} \leftarrow
\mathrm{KD}\bigl(\mathbf{w},\, \mathbf{w}_g^{t+1};\, \mathcal{D}_k\bigr).
\]
\begin{align}
\nabla \mathrm{KD} =&  \nabla\mathbb{E}_{\mathbf{x}_{k,i} \in \mathcal{D}_k} [ -\sigma(\mathbf{x}_{k,i}^\top\,\mathbf{w}_{g}^{t+1}) \log(\sigma(\mathbf{x}_{k,i}^\top\,\mathbf{w}))\\
&-(1-\sigma(\mathbf{x}_i^\top\mathbf{w}_{g}^{t+1}))\log(1-\sigma(\mathbf{x}_i^\top\mathbf{w}))]\notag\\
\stackrel{(a)}=& \mathbb{E}_{x_{k,i} \in \mathcal{D}_k}\Bigl[
  \bigl(\sigma\bigl(\mathbf{x}_{k,i}^{\top}\,\mathbf{w}\bigr)
       -\sigma\bigl(\mathbf{x}_{k,i}^{\top}\mathbf{w}_{g}^{t+1}\bigr)
  \bigr)\mathbf{x}_{k,i}
\Bigr]\\
\stackrel{(b)}=&\mathbb{E}_{\mathbf{x}_{k,i} \in \mathcal{D}_k}\Bigl[
  \sigma(\xi_{k,i})\,\bigl(1 - \sigma(\xi_{k,i})\bigr)\,
  \bigl(\mathbf{x}_{k,i}^{\top}\,\bigl(\mathbf{w} - \mathbf{w}^{t+1}_g\bigr)\bigr)\,
  \mathbf{x}_{k,i}
\Bigr]\\
=&\mathbb{E}_{\mathbf{x}_{k,i} \in \mathcal{D}_k}\Bigl[
  \sigma(\xi_{k,i})\,\bigl(1 - \sigma(\xi_{k,i})\bigr)\,
  \mathbf{x}_{k,i} \mathbf{x}_{k,i}^{\top}\Bigr]\,\bigl(\mathbf{w} - \mathbf{w}^{t+1}_g\bigr)\label{eq:kd loss}
\end{align}
where 
\begin{itemize}[leftmargin=*]
    \item \((a)\):  
     This step employs the standard derivative of the binary cross-entropy term 
\(
-\,p\,\log(q)\;-\;\bigl(1-p\bigr)\,\log\bigl(1-q\bigr).
\)
Differentiation with respect to \(q\), and then accounting for the chain rule via the feature vector, 
yields 
\(\bigl(q - p\bigr)\,\mathbf{x}\).
\item \((b)\):  
    This step applies the Mean Value Theorem (MVT) to the sigmoid difference: 
    \(
      \sigma\bigl(\mathbf{x}_{k,i}^\top\,\mathbf{w}\bigr) \;-\;
      \sigma\bigl(\mathbf{x}_{k,i}^\top\,\mathbf{w}_{g}^{t+1}\bigr)
      =
      \sigma'(\xi_1)\;
      \Bigl[\mathbf{x}_{k,i}^\top\bigl(\mathbf{w}-\mathbf{w}_{g}^{t+1}\bigr)\Bigr],
    \)
    for some \(\xi_1\) between \(\mathbf{x}_{k,i}^\top\,\mathbf{w}\) and \(\mathbf{x}_{k,i}^\top\,\mathbf{w}_{g}^{t+1}\). Since \(\sigma'(z)=\sigma(z)\,[1-\sigma(z)]\), we use \(\sigma(\xi_1)\,[1-\sigma(\xi_1)]\) as a factor. 
\end{itemize}

\smallskip
\noindent\textbf{Local $\to$ Global KD.}
Then we compute the distillation loss definition where the local model serves as the teacher 
and the global model serves as the student. The parameters are given by:
\[
\mathrm{KD} \leftarrow
\mathrm{KD}\bigl(\mathbf{w},\, \mathbf{w}_k^{t};\, \mathcal{I}_k^t\bigr).
\]
Based on Eq.~{\eqref{eq:kd loss}}, we can similarly derive the following equation:
\begin{align*}
\nabla \mathrm{KD} = \mathbb{E}_{\mathbf{x}_{k,i} \in \mathcal{I}_k^t}\Bigl[
  \sigma(\zeta_{k,i})\,\bigl(1 - \sigma(\zeta_{k,i})\bigr)\,
  \mathbf{x}_{k,i} \mathbf{x}_{k,i}^{\top}\Bigr]\,\bigl(\mathbf{w} - \mathbf{w}^{t}_k\bigr)
\end{align*}
We define two errors:
\begin{align}
e_{k}^{t} =&
\nabla \overrightarrow{\mathcal{L}}(\mathbf{w}_{k}^{t},\mathbf{w}^{t}_{g};\, \mathcal{D}_k) \\
=& \nabla \mathcal{L}(\mathbf{w}_{k}^{t};\mathcal{D}_k) +\alpha\mathcal{L}_{KD}\bigl(\mathbf{w}_{k}^{t},\, \mathbf{w}_g^{t};\, \mathcal{D}_k\bigr)\\
=&\nabla \mathcal{L}(\mathbf{w}_{k}^{t};\mathcal{D}_k) +\alpha\mathbb{E}_{\mathbf{x}_{k,i} \in \mathcal{D}_k}\Bigl[
  \sigma(\xi_{k,i})\,\bigl(1 - \sigma(\xi_{k,i})\bigr)\,\label{eq:ekt}\\  
  &\cdot \mathbf{x}_{k,i} \mathbf{x}_{k,i}^{\top}\Bigr]\,\bigl(\mathbf{w}_{k}^{t} - \mathbf{w}^{t}_g\bigr)\notag,
\end{align}
and
\begin{align}
e_{g,k}^{t+1} =&
\nabla \overleftarrow{\mathcal{L}}(\mathbf{w}_{g,k}^{t+1},\mathbf{w}^{t}_{k};\, \mathcal{I}_k^t) \\
=& \nabla \mathcal{L}(\mathbf{w}_{g,k}^{t+1};\mathcal{I}_k^t) +\beta \mathcal{L}_{KD}\bigl(\mathbf{w}_{g,k}^{t+1},\, \mathbf{w}_k^{t};\, \mathcal{I}_k^t\bigr)\\
=&\nabla \mathcal{L}(\mathbf{w}_{g,k}^{t+1};\mathcal{I}_k^t) +\beta \mathbb{E}_{\mathbf{x}_{k,i} \in \mathcal{I}_k^t}\Bigl[
  \sigma(\zeta_{k,i})\,\bigl(1 - \sigma(\zeta_{k,i})\bigr)\,\label{eq: e_gk^t+1}\\  
  &\cdot \mathbf{x}_{k,i} \mathbf{x}_{k,i}^{\top}\Bigr]\,\bigl(\mathbf{w}_{g,k}^{t+1} - \mathbf{w}^{t}_k\bigr)\notag. 
\end{align}
From these Eq.~\eqref{eq: e_gk^t+1} and Eq.~\eqref{eq:ekt}, we isolate expressions of \(\mathbf{w}_{k}^{t} - \mathbf{w}^{t}_g\) and \(\mathbf{w}_{g,k}^{t+1} - \mathbf{w}^{t}_k\):
\begin{align}
\mathbf{w}_{k}^{t} - \mathbf{w}^{t}_g = &\frac{1}{\alpha \mathbb{E}_{\mathbf{x}_{k,i} \in \mathcal{D}_k}[
  \sigma(\xi_{k,i})\,\bigl(1 - \sigma(\xi_{k,i})\bigr)]} \mathbb{E}_{\mathbf{x}_{k,i} \in \mathcal{D}_k}[\mathbf{x}_{k,i} \mathbf{x}_{k,i}^{\top}]^{-1}\label{eq: kt-gt}\\
  &\times (e_{k}^{t}-\nabla \mathcal{L}(\mathbf{w}_{k}^{t};\mathcal{D}_k))\notag, 
\end{align}
and
\begin{align}
\mathbf{w}_{g,k}^{t+1} - \mathbf{w}^{t}_k = &\frac{1}{\beta \mathbb{E}_{\mathbf{x}_{k,i} \in \mathcal{I}_k^t}[
  \sigma(\zeta_{k,i})\,\bigl(1 - \sigma(\zeta_{k,i})\bigr)]} \mathbb{E}_{\mathbf{x}_{k,i} \in \mathcal{I}_k^t}[\mathbf{x}_{k,i} \mathbf{x}_{k,i}^{\top}]^{-1}\label{eq:gkt+1-kt}\\
  &\times (e_{g,k}^{t+1}-\nabla \mathcal{L}(\mathbf{w}_{g,k}^{t+1};\mathcal{I}_k^t))\notag. 
\end{align}
Using the inexactness measure, we set $\gamma := \max\{\gamma_1,\gamma_2\}$. Then we can bound 
Eqs.~\eqref{eq: kt-gt} and \eqref{eq:ekt} into the following form:
\begin{align}
||e^t_k|| =& || \nabla \overrightarrow{\mathcal{L}}(\mathbf{w}_{k}^{t},\mathbf{w}^{t}_{g};\, \mathcal{D}_k)||\\
\stackrel{(a)}\leq&\gamma_1 || \nabla \overrightarrow{\mathcal{L}}(\mathbf{w}_{g}^{t},\mathbf{w}^{t}_{g};\, \mathcal{D}_k)||\\
=&\gamma_1 || \nabla \mathcal{L}(\mathbf{w}_g^t;\mathcal{D}_k)||\\
\leq&\gamma || \nabla \mathcal{L}(\mathbf{w}_g^t;\mathcal{D}_k)||,
\label{ineq:ekt}
\end{align}
where $(a)$ is based on \textbf{Definition~\ref{def:gamma-inexact}}.\\
Next, we consider$||\mathbf{w}_{k}^{t}-\mathbf{w}_{g}^{t}||$, we bound Eq.~\eqref{eq: kt-gt} in another form:
\begin{align}
\label{eq:gk-argmin}
\hat{\mathbf{w}}_{k}^{t}
\;=\;
\arg\min_{\mathbf{w}}
\;
\overrightarrow{\mathcal{L}}\bigl(\mathbf{w},\,\mathbf{w}_g^{t};\mathcal{D}_k\bigr).
\end{align}
\begin{align}
||\mathbf{w}_{k}^{t}-\mathbf{w}_{g}^{t}|| \leq& ||\hat{\mathbf{w}}_{k}^{t}-\mathbf{w}_{g}^{t}||+||\hat{\mathbf{w}}_{k}^{t}-\mathbf{w}_{k}^{t}||\\
\stackrel{(b)}\leq&\frac{1}{\mu}(|| \nabla \overrightarrow{\mathcal{L}}(\mathbf{w}_{g}^{t},\mathbf{w}^{t}_{g};\, \mathcal{D}_k)||+|| \nabla \overrightarrow{\mathcal{L}}(\mathbf{w}_{k}^{t},\mathbf{w}^{t}_{g};\, \mathcal{D}_k)||)\\
\leq&\frac{1+\gamma}{\mu}|| \nabla \overrightarrow{\mathcal{L}}(\mathbf{w}_{g}^{t},\mathbf{w}^{t}_{g};\, \mathcal{D}_k)||\\
=&\frac{1+\gamma}{\mu}|| \nabla\mathcal{L}(\mathbf{w}^{t}_{g};\, \mathcal{D}_k)||,\label{eq: wkt-wgt}
\end{align}
where $(b)$ is based on \textbf{Assumption~\ref{assump:mu-strong-convex}}.\\
Thus, we can bound Eq.~\eqref{eq: e_gk^t+1} into the following form:
\begin{align}
||e^{t+1}_{g,k}|| =& || \nabla \overleftarrow{\mathcal{L}}(\mathbf{w}_{g,k}^{t+1},\mathbf{w}^{t}_{k};\, \mathcal{I}_k^t)||\\
\stackrel{(c)}\leq&\gamma_2 || \nabla \overleftarrow{\mathcal{L}}(\mathbf{w}_{k}^{t},\mathbf{w}^{t}_{k};\, \mathcal{I}_k^t)||\\
=&\gamma_2 || \nabla \mathcal{L}(\mathbf{w}_k^t;I_k^t)||\\
\leq&\gamma || \nabla \mathcal{L}(\mathbf{w}_k^t;I_k^t)||\\
\leq&\gamma( || \nabla \mathcal{L}(\mathbf{w}_k^t;D_k)||+||\nabla \mathcal{L}(\mathbf{w}_k^t;D_k)-\nabla \mathcal{L}(\mathbf{w}_k^t;I_k^t) ||)\\
\stackrel{(d)}\leq&\gamma(1+\theta) || \nabla \mathcal{L}(\mathbf{w}_k^t;D_k)||\\
\leq&\gamma(1+\theta) (|| \nabla \mathcal{L}(\mathbf{w}_g^t;D_k)||+|| \nabla \mathcal{L}(\mathbf{w}_g^t;D_k)\\
&-\nabla \mathcal{L}(\mathbf{w}_k^t;D_k)||)\\
\stackrel{(e)}\leq&\gamma(1+\theta) (|| \nabla \mathcal{L}(\mathbf{w}_g^t;D_k)||+L|| \mathbf{w}_g^t-\mathbf{w}_k^t||)\\
\stackrel{(f)}\leq&(L(1+\gamma)+\mu)\frac{\gamma(1+\theta)}{\mu}||\nabla \mathcal{L}(\mathbf{w}_g^t;D_k)||,
\end{align}
where $(c)$ is based on \textbf{Definition~\ref{def:gamma-inexact}}, $(d)$ is based on \textbf{Assumption~\ref{assump:bounded-subset-dissimilarity}}, $(e)$ is based on \textbf{Assumption~\ref{assump:L-smooth}}, and $(f)$ is based on Eq.~\eqref{eq: wkt-wgt}.

We denote
$\Lambda_1 =\mathbb{E}_{\mathbf{x}_{k,i} \in \mathcal{D}_k}[\mathbf{x}_{k,i}\,\mathbf{x}_{k,i}]$, $c_1 =\mathbb{E}_{\mathbf{x}_{k,i} \in \mathcal{D}_k}\!\bigl[\sigma(\xi_{k,i})\,(1 - \sigma(\xi_{k,i}))\bigr]$, $\Lambda_2 =\mathbb{E}_{\mathbf{x}_{k,i} \in \mathcal{I}_k^t}[\mathbf{x}_{k,i}\,\mathbf{x}_{k,i}]$, $c_2 =\mathbb{E}_{\mathbf{x}_{k,i} \in \mathcal{I}_k^t}\!\bigl[\sigma(\zeta_{k,i})\,(1 - \sigma(\zeta_{k,i}))\bigr]$, $\mathbb{E}_k[\Lambda_1]=\Omega_1$, $\mathbb{E}_k[\Lambda_2]=\Omega_2$, $\mathbb{E}_k[c_1]=d_1$, $\mathbb{E}_k[c_2]=d_2$,

\begin{align}
\mathbb{E}_k[\mathbf{w}_{k}^{t}] - \mathbb{E}_k[\mathbf{w}^{t}_g] = &\frac{\mathbb{E}_k\left[\mathbb{E}_{\mathbf{x}_{k,i} \in \mathcal{D}_k}[\mathbf{x}_{k,i} \mathbf{x}_{k,i}^{\top}]^{-1}\right]}{\alpha \mathbb{E}_k\mathbb{E}_{\mathbf{x}_{k,i} \in \mathcal{D}_k}[
  \sigma(\xi_{k,i})\,\bigl(1 - \sigma(\xi_{k,i})\bigr)] } \\
  &\times \mathbb{E}_k(e_{k}^{t}-\mathbb{E}_k\nabla \mathcal{L}(\mathbf{w}_{k}^{t};\mathcal{D}_k))\notag \\
  =& \frac{\Omega_1^{-1}}{\alpha d_1} \times \mathbb{E}_k(e_{k}^{t}-\mathbb{E}_k\nabla \mathcal{L}(\mathbf{w}_{k}^{t};\mathcal{D}_k)),\label{eq:Ewk}
\end{align}
and
\begin{align}
\mathbb{E}_k[\mathbf{w}_{g,k}^{t+1}] - \mathbb{E}_k[\mathbf{w}^{t}_k] = &\frac{\mathbb{E}_k \left[\mathbb{E}_{\mathbf{x}_{k,i} \in \mathcal{I}_k^t}[\mathbf{x}_{k,i} \mathbf{x}_{k,i}^{\top}]^{-1}\right]}{\beta \mathbb{E}_k \left[\mathbb{E}_{\mathbf{x}_{k,i} \in \mathcal{I}_k^t}[
  \sigma(\zeta_{k,i})\,\bigl(1 - \sigma(\zeta_{k,i})\bigr)]\right]}\\
  & \times(\mathbb{E}_k[e_{g,k}^{t+1}]-\mathbb{E}_k[\nabla \mathcal{L}(\mathbf{w}_{g,k}^{t+1};\mathcal{I}_k^t)])\notag\\
  =&\frac{\Omega_2^{-1}}{\beta d_2} \times (\mathbb{E}_k[e_{g,k}^{t+1}]-\mathbb{E}_k[\nabla \mathcal{L}(\mathbf{w}_{g,k}^{t+1};\mathcal{I}_k^t)]).\label{eq:E(wgk)}
\end{align}
According to Eq.~\eqref{eq:Ewk} and Eq.~\eqref{eq:E(wgk)}, we have
\begin{align}
\mathbf{w}_{g}^{t+1} -\mathbf{w}_{g}^{t} 
         =& \mathbb{E}_k[\mathbf{w}_{g,k}^{t+1}]-\mathbb{E}_k[\mathbf{w}^{t}_k]+\mathbb{E}_k[\mathbf{w}^{t}_k]-\mathbb{E}_k[\mathbf{w}_{g}^{t}]\\
         =& \frac{\Omega_2^{-1}}{\beta d_2} \times (\mathbb{E}_k[e_{g,k}^{t+1}]-\mathbb{E}_k[\nabla \mathcal{L}(\mathbf{w}_{g,k}^{t+1};\mathcal{I}_k^t)])\\
         &+\frac{\Omega_1^{-1}}{\alpha d_1}\times( \mathbb{E}_k [e_{k}^{t}]-\mathbb{E}_k[\nabla \mathcal{L}(\mathbf{w}_{k}^{t};\mathcal{D}_k)])\\
         =& -\frac{\Omega_2^{-1}}{\beta d_2} \times (\mathbb{E}_k[\nabla \mathcal{L}(\mathbf{w}_{g}^{t};\mathcal{D}_k)+M])\\
         &-\frac{\Omega_1^{-1}}{\alpha d_1} \times( \mathbb{E}_k[\nabla \mathcal{L}(\mathbf{w}_{g}^{t};\mathcal{D}_k)+N]),
\end{align}
where we set
\begin{align}
M &= \nabla\mathcal{L} (\mathbf{w}_{g,k}^{t+1};\mathcal{I}_k^t)-\nabla\mathcal{L} (\mathbf{w}_{g}^{t};\mathcal{D}_k)-e_{g,k}^{t+1}\label{eq:m}\\
N &= \nabla\mathcal{L} (\mathbf{w}_{k}^{t};\mathcal{D}_k)-\nabla\mathcal{L} (\mathbf{w}_{g}^{t};\mathcal{D}_k)-e_{k}^{t}\label{eq:n}
\end{align}

Firstly, we bound the following term:
\begin{align}
    || \nabla \overleftarrow{\mathcal{L}}(\mathbf{w}_{g}^{t},\mathbf{w}^{t}_{k};\, \mathcal{I}_k^t)|| =&||\nabla \mathcal{L}(\mathbf{w}_{g}^{t};\mathcal{I}_k^t) +\beta c_2\cdot\Lambda_2 \,\bigl(\mathbf{w}_{g}^{t} - \mathbf{w}^{t}_k\bigr)||\\
    \stackrel{(a)}\leq& ||\nabla \mathcal{L}(\mathbf{w}_{g}^{t};\mathcal{I}_k^t)||+\frac{\beta||\Lambda_2||}{4}||\mathbf{w}_{g}^{t} - \mathbf{w}^{t}_k||\\
    \stackrel{(b)}\leq& (1+\theta)||\nabla \mathcal{L}(\mathbf{w}_{g}^{t};\mathcal{D}_k)||\\&+\frac{(1+\gamma)\beta||\Lambda_2||}{4\mu} ||\nabla \mathcal{L}(\mathbf{w}_{g}^{t};\mathcal{D}_k)||\label{eq:l_left}\\ \notag
    \leq&((1+\theta)+\frac{(1+\gamma)\beta||\Lambda_2||}{4\mu})||\nabla \mathcal{L}(\mathbf{w}_{g}^{t};\mathcal{D}_k)||,
\end{align}
where $(a)$ is due to $0<c_2\leq\frac{1}{4}$, and $(b)$ is due to \textbf{Assumption~\ref{assump:bounded-subset-dissimilarity}} and Eq.~\eqref{eq: wkt-wgt}.

 We consider $||\mathbf{w}_{g,k}^{t+1}-\mathbf{w}_{g}^{t}||$:
\begin{align}
\label{eq:gk-argmin}
\hat{\mathbf{w}}_{g,k}^{t+1}
\;=\;
\arg\min_{\mathbf{w}}
\;
\overleftarrow{\mathcal{L}}\bigl(\mathbf{w},\,\mathbf{w}_k^{t};\mathcal{I}_k^t\bigr).
\end{align}
\begin{align}
||\mathbf{w}_{g,k}^{t+1}-\mathbf{w}_{g}^{t}|| \leq& ||\hat{\mathbf{w}}_{g,k}^{t+1}-\mathbf{w}_{g}^{t}||+||\hat{\mathbf{w}}_{g,k}^{t+1}-\mathbf{w}_{g,k}^{t+1}||\\
\stackrel{(a)}\leq&\frac{1}{\mu}(|| \nabla \overleftarrow{\mathcal{L}}(\mathbf{w}_{g}^{t},\mathbf{w}^{t}_{k};\, \mathcal{I}_k^t)||+|| \nabla \overleftarrow{\mathcal{L}}(\mathbf{w}_{g,k}^{t+1},\mathbf{w}^{t}_{k};\, \mathcal{I}_k^t)||)\\
=&\frac{1}{\mu}(|| e_{g,k}^{t+1}||+|| \nabla \overleftarrow{\mathcal{L}}(\mathbf{w}_{g}^{t},\mathbf{w}^{t}_{k};\, \mathcal{I}_k^t)||)\\
\stackrel{(b)}\leq& ((L(1+\gamma)+\mu)\frac{\gamma(1+\theta)}{\mu}+(1+\theta)+\frac{(1+\gamma)\beta||\Omega_2||}{4\mu})\label{eq:wgkt+1-wgt}\\
&\cdot ||\nabla \mathcal{L}(\mathbf{w}_{g}^{t};\mathcal{D}_k)||,
\end{align}
where $(a)$ is due to \textbf{Assumption~\ref{assump:mu-strong-convex}}, and $(b)$ is based on Eq.~\eqref{eq: e_gk^t+1} and Eq.~\eqref{eq:l_left}.

We can bound $||M||$ and $||N||$ in the following form:
\begin{align}
    ||M||\leq& ||\nabla\mathcal{L} (\mathbf{w}_{g,k}^{t+1};\mathcal{D}_k)-\nabla\mathcal{L} (\mathbf{w}_{g}^{t};\mathcal{D}_k)||\\
    &+||\nabla\mathcal{L} (\mathbf{w}_{g,k}^{t+1};\mathcal{I}_k^t)-\nabla\mathcal{L} (\mathbf{w}_{g,k}^{t+1};\mathcal{D}_k)||+||e_{g,k}^{t+1}||\\
    \leq& ||\nabla\mathcal{L} (\mathbf{w}_{g,k}^{t+1};\mathcal{D}_k)-\nabla\mathcal{L} (\mathbf{w}_{g}^{t};\mathcal{D}_k)||\\
    &+\theta||\nabla\mathcal{L} (\mathbf{w}_{g,k}^{t+1};\mathcal{D}_k)||+||e_{g,k}^{t+1}||\\
\leq& (1+\theta)||\nabla\mathcal{L} (\mathbf{w}_{g,k}^{t+1};\mathcal{D}_k)-\nabla\mathcal{L} (\mathbf{w}_{g}^{t};\mathcal{D}_k)||\\
    &+\theta||\nabla\mathcal{L} (\mathbf{w}_{g}^{t};\mathcal{D}_k)||+||e_{g,k}^{t+1}||\\
\leq& (1+\theta)L||\mathbf{w}_{g,k}^{t+1}-\mathbf{w}_{g}^{t}|| +\theta||\nabla\mathcal{L} (\mathbf{w}_{g}^{t};\mathcal{D}_k)||+||e_{g,k}^{t+1}||\label{eq:m}\\
\stackrel{(a)}\leq& r_1^\prime ||\nabla\mathcal{L} (\mathbf{w}_{g}^{t};\mathcal{D}_k)||
\end{align}
where $(a)$ is based on Eq.~\eqref{eq:wgkt+1-wgt} and Eq.~\eqref{eq: e_gk^t+1}, and $r_1^{\prime} =((L(1+\gamma)+\mu)\frac{\gamma(1+\theta)}{\mu}+(1+\theta)+\frac{(1+\gamma)\beta||\Lambda_2||}{4\mu})(1+\theta)L+\theta+(L(1+\gamma)+\mu)\frac{\gamma(1+\theta)}{\mu}) $ and $r_1 =((L(1+\gamma)+\mu)\frac{\gamma(1+\theta)}{\mu}+(1+\theta)+\frac{(1+\gamma)\beta||\Omega_2||}{4\mu})(1+\theta)L+\theta+(L(1+\gamma)+\mu)\frac{\gamma(1+\theta)}{\mu}) $.
\begin{align}
    ||N|| \leq& ||\nabla\mathcal{L} (\mathbf{w}_{k}^{t};\mathcal{D}_k)-\nabla\mathcal{L} (\mathbf{w}_{g}^{t};\mathcal{D}_k)||+||e_{k}^{t}||\\
    \leq& L||\mathbf{w}_{k}^{t}-\mathbf{w}_{g}^{t}||+||e_{k}^{t}||\label{eq:n}
        \stackrel{(a)}\leq r_2||\nabla\mathcal{L} (\mathbf{w}_{g}^{t};\mathcal{D}_k)||
\end{align}
where $(a)$ is based on Eq.~\eqref{eq: kt-gt} and Eq.~\eqref{eq:ekt}, and $r_2 = \frac{L(1+\gamma)}{\mu}+\gamma$.

We set $<\mathcal{L}(\mathbf{w}_{g}^{t};\mathcal{D}_g),\mathbf{w}_{g}^{t+1} -\mathbf{w}_{g}^{t}>= U$, and
according to Eq.~\eqref{eq:m}, Eq.~\eqref{eq:n} and \textbf{Assumption~\ref{assump:bounded-grad-dissimilarity}}, we have
\begin{align}
    U =& (-\frac{\Omega_2^{-1}}{\beta d_2}-\frac{\Omega_1^{-1}}{\alpha d_1})<\mathcal{L}(\mathbf{w}_{g}^{t};\mathcal{D}_g),\mathbb{E}_k[\nabla \mathcal{L}(\mathbf{w}_{g}^{t};\mathcal{D}_k)]>\\
    &+(-\frac{\Omega_2^{-1}}{\beta d_2}-\frac{\Omega_1^{-1}}{\alpha d_1})<\mathcal{L}(\mathbf{w}_{g}^{t};\mathcal{D}_g),\mathbb{E}_k[M]+\mathbb{E}_k[N]>\\
    \leq&||\frac{\Omega_2^{-1}}{\beta d_2}+\frac{\Omega_1^{-1}}{\alpha d_1}||( -B||\mathcal{L}(\mathbf{w}_{g}^{t};\mathcal{D}_g)||^2\\
    &+||\mathcal{L}(\mathbf{w}_{g}^{t};\mathcal{D}_g)||(\mathbb{E}_k[||M||]+\mathbb{E}_k[||N||]))\\
    \leq&(||\frac{\Omega_2^{-1}}{\beta d_2}||+||\frac{\Omega_1^{-1}}{\beta d_1}||)(r_1+r_2-1)B\cdot\mathcal{L}(\mathbf{w}_{g}^{t};\mathcal{D}_g)\\
    \leq&4(||\frac{\Omega_2^{-1}}{\beta }||+||\frac{\Omega_1^{-1}}{\alpha }||)(r_1+r_2-1)B\cdot\mathcal{L}(\mathbf{w}_{g}^{t};\mathcal{D}_g)\\
    \leq&(\frac{4}{\beta||\Omega_2|| }+\frac{4}{\alpha||\Omega_1|| })(r_1+r_2-1)B\cdot\mathcal{L}(\mathbf{w}_{g}^{t};\mathcal{D}_g)\label{eq:u}
\end{align}
From Eq.~\eqref{eq:wgkt+1-wgt} and \textbf{Assumption~\ref{assump:bounded-grad-dissimilarity}}, we can get 
\begin{align}
    ||\mathbf{w}_g^{t+1}-\mathbf{w}_g^t||=&\mathbb{E}_k[||\mathbf{w}_{g,k}^{t+1}-\mathbf{w}_{g}^{t}||]
    \\\leq& B((L(1+\gamma)+\mu)\frac{\gamma(1+\theta)}{\mu}\label{wgt+1-wgt}\\
&+(1+\theta)+\frac{(1+\gamma)\beta||\Omega_2||}{4\mu})\cdot ||\nabla \mathcal{L}(\mathbf{w}_{g}^{t};\mathcal{D}_g)||\notag
\end{align}
According to Eqs.~\eqref{wgt+1-wgt}, ~\eqref{eq:u}, \textbf{Assumption~\ref{assump:mu-strong-convex}}, and \textbf{Assumption~\ref{assump:L-smooth}}, we can get
\begin{align}
\mathcal{L}(\mathbf{w}_{g}^{t+1};\mathcal{D}_g) \leq& \mathcal{L}(\mathbf{w}_{g}^{t};\mathcal{D}_g)+<\mathcal{L}(\mathbf{w}_{g}^{t};\mathcal{D}_g),\mathbf{w}_{g}^{t+1}-\mathbf{w}_{g}^{t}>\\
&+\frac{L}{2}||\mathbf{w}_{g}^{t+1}-\mathbf{w}_{g}^{t}||^2\\
\leq&\mathcal{L}(\mathbf{w}_{g}^{t};\mathcal{D}_g) -r||\nabla\mathcal{L}(\mathbf{w}_{g}^{t};\mathcal{D}_g)||^2
\end{align}
\begin{align*}
\mathcal{L}(\mathbf{w}_g^{t+1};\mathcal{D}_g) - \mathcal{L}(\mathbf{w}^*;\mathcal{D}_g)\leq& (1-2\mu r)
[\mathcal{L}(\mathbf{w}^t_g;\mathcal{D}_g) - \mathcal{L}(\mathbf{w}^*;\mathcal{D}_g)].
\end{align*}
where we set
\(
    r=(\frac{4}{\beta||\Omega_2|| }+\frac{4}{\alpha||\Omega_1|| })B-\frac{LB^2}{2}((L(1+\gamma)+\mu)\frac{\gamma(1+\theta)}{\mu}+(1+\theta)+\frac{(1+\gamma)\beta||\Omega_2||}{4\mu})^2
    - (\frac{4}{\beta||\Omega_2|| }+\frac{4}{\alpha||\Omega_1|| })(r_1+r_2)B\), $r_1 =((L(1+\gamma)+\mu)\frac{\gamma(1+\theta)}{\mu}+(1+\theta)+\frac{(1+\gamma)\beta||\Omega_2||}{4\mu})(1+\theta)L+\theta+(L(1+\gamma)+\mu)\frac{\gamma(1+\theta)}{\mu}) $, $r_2 = \frac{L(1+\gamma)}{\mu}+\gamma$, $\Omega_1 =\mathbb{E}_k[\mathbb{E}_{\mathbf{x}_{k,i} \in \mathcal{D}_k}[\mathbf{x}_{k,i}\,\mathbf{x}_{k,i}]]$ and $\Omega_2 =\mathbb{E}_k[\mathbb{E}_{\mathbf{x}_{k,i} \in \mathcal{I}_k^t}[\mathbf{x}_{k,i}\,\mathbf{x}_{k,i}]]$.

\section{Traditional non-iid setting EXPERIMENTS}
\label{sec:appendix-exp1}
\subsection{Non-IID Settings}
To ensure fair comparisons and consistent experimental setups with existing collaborative fairness approaches, 
we adopt the same environment and configurations as {FedAVE}~\cite{wang2024fedave} and 
{FedSAC}~\cite{wang2024fedsac}. We adopt the following traditional non-IID setting. \textbf{CLA} (Imbalanced Class Distributions) used in~\cite{xu2021gradient,Lyu2020CollaborativeFairness} and \textbf{DIR} (Imbalanced Sizes + Class Distributions) used in~\cite{g,yurochkin2019bayesian} 
are commonly used non-IID settings in federated learning. 
Under these scenarios, 
the label types across clients become severely imbalanced, resulting in skewed 
partitions that pose significant challenges for model training and fairness.
\textbf{POW} (Imbalanced Dataset Sizes) was also used in this experiment.


\subsection{Federated Data Simulation}
In the simulation experiments, we use two image classification datasets: \textbf{Fashion MNIST}~\cite{xiao2017fashion} and 
\textbf{CIFAR10}~\cite{krizhevsky2009learning}. 
\textbf{Fashion MNIST} contains 70,000 grayscale images 
(\(28\times28\)) evenly split into 10 classes (e.g., T-shirt/top, trousers). \textbf{CIFAR10} consists of 60,000 color images (\(32\times32\)) across 10 classes (e.g., airplane, bird). 
We partition the datasets into training, validation, and testing in a ratio of 7:1:2. In addition, we set the number of clients as $K=10$. 
To simulate the \textbf{POW} partition, we follow a power law with an exponent of $1$ to divide the global data into 10 clients. For the $k$-th client, its data size is $|\mathcal{D}_k|=\frac{1}{kZ}|\mathcal{D}_g|$, where $Z = \sum_{k=1}^{10} \frac{1}{k}$. 
For the \textbf{CLA} partition, we fix the local data size and assign labels in an imbalanced manner: the first client receives data from 1 class, the second client from 2 classes, the third client from 3 classes, and so on until the tenth client, which receives data from 10 classes. For the \textbf{DIR} partition, we construct three different splits using a Dirichlet distribution 
with concentration parameters \(\alpha = 1\), \(\alpha = 2\), and \(\alpha = 3\). 
Concretely, suppose there are \(C\) classes in the global dataset \(\mathcal{D}_g\). 
For each class \(c\), let \(|\mathcal{D}_c|\) denote the total number of samples in class \(c\), 
and sample a probability vector
\(
\mathbf{p}_c = \bigl(p_{c,1}, \, p_{c,2}, \,\dots,\, p_{c,K}\bigr) 
    \;\sim\; \mathrm{Dirichlet}\bigl(\alpha,\alpha,\dots,\alpha\bigr),
\)
where \(K=10\) is the number of clients. We then allocate 
\(\lfloor p_{c,k}\,|\mathcal{D}_c|\rfloor\) samples of class \(c\) to client \(k\). Hence, 
the local dataset for client \(k\) is
\(
\mathcal{D}_k = \bigcup_{c=1}^C \mathcal{D}_{c,k}, 
\quad \text{where} \quad 
|\mathcal{D}_{c,k}| = \Bigl\lfloor p_{c,k} \cdot |\mathcal{D}_c| \Bigr\rfloor.
\)
\subsection{Baselines}
We compare our method against two categories of baselines: 
\textit{Collaborative Fairness} algorithms designed for non-IID data,
and two standard references without fairness considerations.
Specifically, we include \textbf{CGSV}~\cite{xu2021gradient}, 
\textbf{CFFL}~\cite{Lyu2020CollaborativeFairness}, 
\textbf{FedAVE}~\cite{wang2024fedave}, 
and \textbf{FedSAC}~\cite{wang2024fedsac}, 
which explicitly address fairness by measuring client contributions or customizing reward allocations. 
Additionally, we compare with 
\textbf{Standalone} (each client trains independently without aggregation) 
and the classic \textbf{FedAvg}~\cite{mcmahan2017communication} for federated averaging.
Unlike personalized federated learning approaches that tailor models to each client's local distribution,
our setting uses a global test set and aims to evaluate overall model performance under traditional non-IID data partitions.
Hence, we do not include personalized FL baselines here, since they focus on fitting each client's individual feature space. 
Instead, we follow the conventional setup in collaborative fairness (CF) research and compare only with CF-oriented baselines 
under these non-IID settings.

\subsection{Implementation}
Following the literature~\cite{wang2024fedave,wang2024fedsac}, we adopt a 2-layer Convolutional Neural Network (CNN) for the \emph{Fashion MNIST} dataset, with mini-batch size $B = 32$ and learning rate $\text{lr} = 0.15$. Then, for \emph{CIFAR10}, we employ a 3-layer CNN~\cite{wang2024fedave,wang2024fedsac} with mini-batch size $B = 128$ and learning rate $\text{lr} = 0.015$. We implement all baselines and our model in PyTorch and train them on an NVIDIA RTX A6000 GPU.

\subsection{Evaluation Metrics}
Since CLA and DIR settings only allow data with partial labels on each client, they cannot use the local evaluation as we did in the main experiments for the imbalanced covariate shift setting. Thus, following existing work~\cite{wang2024fedave,wang2024fedsac}, we conduct global evaluations, i.e., all clients are evaluated on a single global test set.
We still report the average values of \textit{Maximum Client Accuracy} and \textit{Collaborative Fairness (CF) Coefficient} for \textbf{three runs}. 

\subsection{Results of Experiments}
In the traditional federated learning setting (using a single global test set under non-IID data partitions),
Table~\ref{tab:exp1_res} compares both \emph{maximum accuracy} and \emph{collaborative fairness} 
of our method against various baselines. We observe that our approach consistently outperforms the baselines 
in terms of both metrics. Specifically, for the POW partition on FashionMNIST, our method achieves 
\(\mathbf{87.93}\pm0.25\%\) in max accuracy, surpassing FedSAC (\(87.83\pm0.07\%\)) and other baselines. A similar advantage is seen on CIFAR10, where \ours outperforms FedSAC and FedAVE under the same partition.
For the DIR partition, our method \ours attains higher maximum accuracy on both FashionMNIST and CIFAR10 with different Dirichlet parameters (\(\alpha=1,2,3\)). 
When \(\alpha=1.0\), we consistently see around \(+0.2\%\) to \(+1.2\%\) improvement over the best baseline in maximum accuracy.
For the collaborative fairness metric, our approach yields higher fairness scores, e.g., \(98.93\pm0.14\%\) (POW/FashionMNIST) compared with \(96.53\pm1.20\%\) by FedSAC. This indicates that the performance gap among clients is smaller under our method, demonstrating better collaborative fairness.


\begin{table*}[t]
  \centering
  \small
  \setlength{\tabcolsep}{3pt}
  \caption{Comparison of fairness and accuracy for traditional non-IID setting experiments.}
  \label{tab:exp1_res}

  \subcaptionbox{Maximum Client Accuracy (\%)}{\label{tab:max-acc-noniid}%
  \resizebox{0.95\textwidth}{!}{%
  \begin{tabular}{|l|c|c|c|c|c|c|c|c|c|c|}
    \hline
    \multirow{2}{*}{\textbf{Method}} &
    \multicolumn{5}{c|}{\textbf{FashionMNIST}} &
    \multicolumn{5}{c|}{\textbf{CIFAR10}} \\
    \cline{2-11}
      & \textbf{POW} & \textbf{CLA} & \textbf{DIR(1.0)} & \textbf{DIR(2.0)} & \textbf{DIR(3.0)}
      & \textbf{POW} & \textbf{CLA} & \textbf{DIR(1.0)} & \textbf{DIR(2.0)} & \textbf{DIR(3.0)} \\
    \hline
    \textbf{Standalone} & 84.10$\pm$0.22 & 79.00$\pm$0.87 & 81.92$\pm$0.19 & 83.22$\pm$0.36 & 84.64$\pm$0.29
                       & 59.06$\pm$0.23 & 32.61$\pm$0.22 & 44.19$\pm$0.84 & 45.59$\pm$0.29 & 47.40$\pm$0.89 \\
    \textbf{FedAvg}    & 87.57$\pm$0.23 & 81.04$\pm$1.22 & 85.68$\pm$0.77 & 86.16$\pm$1.14 & 86.80$\pm$0.30
                       & 59.87$\pm$0.95 & 50.84$\pm$0.14 & 49.82$\pm$1.63 & 49.00$\pm$0.76 & 48.16$\pm$0.97 \\
    \textbf{CFFL}      & 83.78$\pm$0.56 & 85.88$\pm$0.82 & 86.90$\pm$0.19 & 87.09$\pm$0.08 & 87.44$\pm$0.27
                       & 58.41$\pm$1.15 & 42.89$\pm$1.22 & 48.28$\pm$1.72 & 42.15$\pm$0.93 & 44.21$\pm$1.27 \\
    \textbf{CGSV}      & 82.68$\pm$0.97 & 82.72$\pm$0.99 & 81.39$\pm$1.24 & 84.92$\pm$0.83 & 86.76$\pm$1.46
                       & 53.24$\pm$0.86 & 43.98$\pm$1.81 & 47.92$\pm$1.33 & 48.92$\pm$1.24 & 46.21$\pm$0.41 \\
    \textbf{FedAVE}    & 85.99$\pm$0.82 & 80.26$\pm$0.64 & 85.29$\pm$0.73 & 85.60$\pm$0.44 & 85.27$\pm$0.22
                       & 59.11$\pm$0.54 & 45.16$\pm$0.91 & 50.11$\pm$1.93 & 51.24$\pm$1.14 & 42.14$\pm$0.33 \\
    \textbf{FedSAC}    & 87.83$\pm$0.07 & 85.28$\pm$0.66 & 87.33$\pm$0.49 & 87.15$\pm$0.45 & 87.81$\pm$0.14
                       & 58.24$\pm$0.11 & 47.92$\pm$0.65 & 52.14$\pm$0.13 & 50.31$\pm$1.33 & 49.92$\pm$0.14 \\
    \rowcolor{greyL}
    \textbf{\ours}     & \textbf{87.93}$\pm$0.25 & \textbf{86.03}$\pm$0.48 & \textbf{87.57}$\pm$0.13 & \textbf{87.27}$\pm$0.38 & \textbf{87.96}$\pm$0.11
                       & \textbf{60.03}$\pm$0.38 & \textbf{50.62}$\pm$0.72 & \textbf{53.32}$\pm$0.24 & \textbf{53.12}$\pm$0.77 & \textbf{50.09}$\pm$0.15 \\
    \hline
  \end{tabular}}}

  \bigskip   

  \subcaptionbox{Collaborative Fairness (CF) Coefficient}{\label{tab:cf-noniid}%
  \resizebox{0.95\textwidth}{!}{%
  \begin{tabular}{|l|c|c|c|c|c|c|c|c|c|c|}
    \hline
    \multirow{2}{*}{\textbf{Method}} &
    \multicolumn{5}{c|}{\textbf{FashionMNIST}} &
    \multicolumn{5}{c|}{\textbf{CIFAR10}} \\
    \cline{2-11}
      & \textbf{POW} & \textbf{CLA} & \textbf{DIR(1.0)} & \textbf{DIR(2.0)} & \textbf{DIR(3.0)}
      & \textbf{POW} & \textbf{CLA} & \textbf{DIR(1.0)} & \textbf{DIR(2.0)} & \textbf{DIR(3.0)} \\
    \hline
    \textbf{FedAvg}  & 16.78$\pm$21.51 & 83.23$\pm$6.53 & 20.18$\pm$16.52 & 23.24$\pm$27.33 & 29.00$\pm$31.87
                     & 18.84$\pm$6.83  & 86.83$\pm$4.82 & 32.77$\pm$6.43 & 27.92$\pm$8.52 & 50.73$\pm$10.39 \\
    \textbf{CFFL}    & 89.43$\pm$1.52  & 91.85$\pm$2.39 & 88.09$\pm$3.43 & 87.29$\pm$4.21 & 76.23$\pm$10.32
                     & 83.52$\pm$4.28  & 70.62$\pm$3.62 & 66.28$\pm$4.15 & 70.51$\pm$8.31 & 71.35$\pm$2.51  \\
    \textbf{CGSV}    & 90.87$\pm$0.98  & 86.63$\pm$2.89 & 92.43$\pm$0.64 & 91.29$\pm$1.93 & 87.53$\pm$2.34
                     & 90.15$\pm$3.37  & 94.26$\pm$0.82 & 89.27$\pm$2.05 & 91.74$\pm$0.45 & 93.63$\pm$1.37  \\
    \textbf{FedAVE}  & 94.21$\pm$1.41  & 89.42$\pm$0.44 & 93.22$\pm$1.28 & 90.34$\pm$0.33 & 94.34$\pm$0.98
                     & 93.64$\pm$0.35  & 96.25$\pm$1.87 & 95.85$\pm$1.86 & 98.03$\pm$0.13 & 90.62$\pm$1.69  \\
    \textbf{FedSAC}  & 96.53$\pm$1.20  & 93.45$\pm$1.75 & 97.31$\pm$0.83 & 95.78$\pm$0.89 & 94.34$\pm$2.93
                     & 99.05$\pm$0.24  & 95.63$\pm$0.51 & 93.54$\pm$0.99 & 95.73$\pm$1.37 & 94.73$\pm$0.97  \\
    \rowcolor{greyL}
    \textbf{\ours}   & \textbf{98.93}$\pm$0.14 & \textbf{95.13}$\pm$1.83 & \textbf{99.27}$\pm$0.23 & \textbf{97.87}$\pm$0.45 & \textbf{98.82}$\pm$0.42
                     & \textbf{99.78}$\pm$0.08 & \textbf{97.35}$\pm$1.20 & \textbf{97.05}$\pm$0.21 & \textbf{98.86}$\pm$0.72 & \textbf{97.76}$\pm$1.59 \\
    \hline
  \end{tabular}}}
\end{table*}

\section{Covariate Shift Data Generation}
\label{sec:appendix-single-gauss}

In this section, we detail how to construct a covariate-shifted dataset from a baseline distribution with Algorithm~\ref{alg:gaussian-covariate-shift}. We assume the underlying global dataset 
\(\mathcal{D}_g\) approximately follows a single Gaussian distribution 
\(\mathcal{N}(\mu,\Sigma)\), from which we draw subsets for different clients. 
For each client, we shift the \emph{features} by a fixed Mahalanobis distance.

\begin{algorithm}[h]
\caption{Covariate Shift Data Generation}
\label{alg:gaussian-covariate-shift}
\begin{algorithmic}[1]
\Require 
  Global labeled dataset $\mathcal{D}_g = \{(x_i,y_i)\}_{i=1}^{|\mathcal{D}_g|}$;
  constant $C>0$;
  number of clients $K$;
  desired sample counts $\{n_k\}_{k=1}^{K}$ (such that $\sum_{k=1}^K n_k = \lfloor |\mathcal{D}_g|/2 \rfloor$);
  feature dimension $d$.
\Ensure 
  Covariate-shifted client datasets $\{\mathcal{D}_k\}_{k=1}^{K}$, 
  where each $\mathcal{D}_k$ retains the original labels but has shifted features 
  (via importance sampling with respect to a Gaussian distribution with mean $\mu_k$ and covariance $\Sigma$).

\vspace{3pt}
\State \textbf{Vectorize Features:} 
Convert each feature $x_i$ in $\mathcal{D}_g$ into a vector in $\mathbb{R}^d$. 
Let $\mathbf{X}$ be the resulting $|\mathcal{D}_g|\times d$ matrix. 

\State \textbf{Estimate Baseline Gaussian:}
\begin{align*}
\mu &\;=\; \frac{1}{|\mathcal{D}_g|}\,\sum_{i=1}^{|\mathcal{D}_g|} x_i, 
\quad
\Sigma \;=\; \mathrm{Cov}(\mathbf{X}).
\end{align*}

\State \textbf{Half-sampling setup:} 
\State Let $M = \lfloor |\mathcal{D}_g| / 2 \rfloor$. 
We only use half of the global dataset for experiments. 
Hence, set $\sum_{k=1}^K n_k = M$. Adjust $\{n_k\}$ if needed.

\For{$k=1,\dots,K$}
    \State \textbf{Compute mean shift $\delta_k$:}
    \Statex \quad Sample a vector $\delta_{k} \in \mathbb{R}^d$ 
                 such that $\delta_{k}^\top \Sigma^{-1}\delta_{k} = C$.
    \State $\mu_{k} \gets \mu + \delta_{k}$ 
    \Comment{Shifted mean for client $k$.}
    \Statex

    \State \textbf{Construct Shifted Dataset for client $k$:}
    \Statex \quad  \emph{(1) Compute importance weights}  
    \Statex \quad For each $(x_i, y_i)$ in $\mathcal{D}_g$, compute:
    \[
       w_i \;=\; \exp\!\Bigl(-\tfrac{1}{2}\,(x_i - \mu_k)^\top \Sigma^{-1}(x_i - \mu_k)\Bigr).
    \]
    \Statex \quad Then normalize:
    \[
       \tilde{w}_i \;=\; \frac{w_i}{\sum_{j=1}^{|\mathcal{D}_g|} w_j}.
    \]

    \Statex \quad \emph{(2) Sample from $\mathcal{D}_g$ by $\tilde{w}_i$}  
    \Statex \quad Sample $n_k$ pairs $(x_i,y_i)$ from $\mathcal{D}_g$ \emph{with replacement} 
      according to probabilities $\tilde{w}_i$.
    \Statex \quad Store these sampled pairs in $\mathcal{D}_k$.
\EndFor

\State \Return $\{\mathcal{D}_k\}_{k=1}^{K}$.

\end{algorithmic}
\end{algorithm}

 As discussed in Theorem~\ref{theorem2}, under a large-sample limit, the Kullback-Leibler (KL) divergence 
can be used to measure how much the client-specific distribution \(p_{\theta'}\) deviates from the baseline model 
\(p_{\omega}\). By fixing \(\|\delta_k\|_{\Sigma^{-1}}^2 = \delta_{k}^\top \Sigma^{-1}\delta_{k} = C\) for each client, 
we ensure a controlled covariate shift in the mean. The covariance \(\Sigma\) remains the same, and the degree of shift is comparable across all clients. This construction thus provides a 
systematic way to generate imbalanced covariate shifts, enabling direct measurement 
and comparison of fairness or performance in federated learning experiments.

\section{Baselines}
\label{sec:appendix-baselines}

In this appendix, we present the details of the baseline methods used in our experiments. We classify them into two main categories and also include two additional traditional baselines: \textbf{Standalone} training and the classic \textbf{FedAvg}.

\smallskip
\noindent\textit{\underline{Standalone and FedAvg.}}
The first traditional baseline is \textbf{Standalone}, in which each client trains independently without any model aggregation. This approach ignores the potential benefits of federated collaboration, providing a lower-bound performance reference.
Next, we employ \textbf{FedAvg}~\cite{mcmahan2017communication}, the standard federated averaging method. FedAvg updates the global model by performing weighted averaging of the locally trained models from all clients, thereby enabling knowledge sharing while minimizing data exchange.

\smallskip
\noindent\textit{\underline{Collaborative Fairness Baselines.}}
We next consider a set of baselines specifically aimed at improving collaborative fairness. 
\textbf{CFFL}~\cite{Lyu2020CollaborativeFairness} focuses on distributing rewards proportionally by measuring each client’s contribution differences. In practice, it evaluates client updates on a held-out validation set (or an equivalent performance benchmark) before deciding how to compensate high- versus low-contribution clients.
On the other hand, \textbf{CGSV}~\cite{xu2021gradient} elies on gradient-based importance metrics—specifically using cosine similarities of gradients—so it does not require a separate validation set. Each client’s contribution is gauged by comparing its gradient direction against the aggregated global gradient.
Additionally, \textbf{FedAVE}~\cite{wang2024fedave} incorporates explicit fairness by maintaining a global validation set at the server. It periodically tests each client’s model update on this validation set to compute a “reputation” score, which then guides how the final global model is aggregated.
\textbf{FedSAC}~\cite{wang2024fedsac} similarly uses a global validation set for evaluating each client’s local update. Clients receive a proportionate “reward gradient” based on their evaluated performance, ensuring that clients with higher impact on the validation set obtain a larger share of the global update.

\noindent\textit{\underline{Personalized FL (Covariate Shift) Baselines.}}
Since our approach also falls under the umbrella of Personalized Federated Learning (FL), we include baselines originally designed to tackle feature-level (covariate) shifts, even though they do not explicitly target collaborative fairness. 
\textbf{FedDC}~\cite{g} tackles non-IID data by introducing a drift variable that aligns local models more closely with the global model.  
\textbf{FedAS}~\cite{yang2024fedas} alleviates intra- and inter-client inconsistencies through federated parameter alignment and client synchronization. 
\textbf{pFedCK}~\cite{zhang2024pFedCK} clusters clients by update similarity and performs mutual knowledge distillation between interactive and personalized models to enhance robustness under data heterogeneity. 
Lastly, \textbf{FedMPR}~\cite{goksu2024robust} combines iterative magnitude pruning with regularization techniques to improve robustness under highly heterogeneous client data distributions.

By comparing these diverse baselines, we can comprehensively evaluate our proposed method from both collaborative fairness and feature-level drift perspectives.

\section{Implementation Details}
\label{sec:appendix-exp2}
\subsection{Common Hyperparameters of All Algorithms Used in Simulation}

\textbf{Number of clients (NUM\_CLIENTS)} is set to 10. This indicates how many total clients are simulated or trained independently in the scenario. 
 \textbf{Global rounds (NUM\_GLOBAL\_ROUNDS)} is commonly set to 20. This is the total number of federated communication rounds.
\textbf{Local epochs (LOCAL\_EPOCHS)} is often set to 1. It denotes how many epochs each client trains on its local data per global round.
\textbf{Batch size (BATCH\_SIZE)} is often set to 32.   
\textbf{Learning rates (learning\_rates)} are set to 0.001 for \textit{FashionMNIST} and  0.005 for \textit{CIFAR10}. These control the local step size for optimizing \texttt{SGD}.

\subsection{Algorithm-Specific Hyperparameters}

Table~\ref{tab:alg-specific-hp-condensed} details each algorithm's unique or particularly important hyperparameters, along with their default values (or ranges) and a brief explanation.

\begin{table}[t]
\centering
\caption{Algorithm-Specific Hyperparameters (Condensed)}
\label{tab:alg-specific-hp-condensed}
 \resizebox{0.95\columnwidth}{!}{
\begin{tabular}{|p{1.4cm}|p{4.2cm}|p{2.2cm}|}
\hline
\textbf{Algorithm} & \textbf{Special Hyperparameters} & \textbf{Value} \\
\hline
\textbf{CFFL} &
\begin{tabular}[c]{@{}l@{}}
1) THETA\_U (grad upload) \\
2) CLIP\_NORM (clip thr.) \\
3) C\_TH (rep. thr.) \\
4) ALPHA (rep. update)
\end{tabular}
&
\begin{tabular}[c]{@{}l@{}}
(1) 0.5\\
(2) 5.0\\
(3) 0.05\\
(4) 1.0
\end{tabular}
\\ \hline

\textbf{CGSV} &
\begin{tabular}[c]{@{}l@{}}
1) ALPHA\_R (mov. avg) \\
2) BETA (sim. scaling) \\
3) SPARSITY \\
4) ALTRUISM
\end{tabular}
&
\begin{tabular}[c]{@{}l@{}}
(1) 0.9\\
(2) 2.0\\
(3) True\\
(4) 1.0
\end{tabular}
\\ \hline

\textbf{FedAVE} &
\begin{tabular}[c]{@{}l@{}}
1) UPLOAD\_FRAC \\
2) DOWNLOAD\_FRAC\_BASE\\
3) ALPHA / BETA
\end{tabular}
&
\begin{tabular}[c]{@{}l@{}}
(1) 0.5\\
(2) 0.3\\
(3) 0.9 / 1.0
\end{tabular}
\\ \hline

\textbf{FedAvg} &
--- & 
--- 
\\ \hline

\textbf{FedDC} &
\begin{tabular}[c]{@{}l@{}}
1) ALPHA (penalty) \\
2) drift\_vars
\end{tabular}
&
\begin{tabular}[c]{@{}l@{}}
(1) 1.0\\
(2) init=0
\end{tabular}
\\ \hline

\textbf{FedMPR} &
PRUNE\_PERCENT & 
0.1
\\ \hline

\textbf{FedProx} &
MU (prox. coeff.) &
0.01
\\ \hline

\textbf{FedSAC} &
\begin{tabular}[c]{@{}l@{}}
1) BETA (c\_i mapping)\\
2) mid\_round
\end{tabular}
&
\begin{tabular}[c]{@{}l@{}}
(1) 2.0\\
(2) 15
\end{tabular}
\\ \hline

\textbf{SCAFFOLD} &
\begin{tabular}[c]{@{}l@{}}
1) $\eta_g$ (global LR)\\
2) $\eta_l$ (local LR)\\
3) K (local steps)\\
4) c\_global, c\_local
\end{tabular}
&
\begin{tabular}[c]{@{}l@{}}
(1) 0.005\\
(2) 0.1\\
(3) 1\\
(4) init=0
\end{tabular}
\\ \hline
\textbf{FedAKD} &
\begin{tabular}[c]{@{}l@{}}
1) Distill $\alpha$ \\ 
2) Distill $\beta$ \\
3) Temp $T$
\end{tabular}
&
\begin{tabular}[c]{@{}l@{}}
(1) 1.0\\
(2) 1.0\\
(3) 1.0
\end{tabular}
\\ \hline
\end{tabular}
}
\end{table}

\section{The EHR Dataset}
\label{sec:appendix-ehr}
\begin{table}[htbp]
\centering
\caption{State Data Statistics}
\label{tab:state_data_stats}
 \resizebox{0.95\columnwidth}{!}{
\begin{tabular}{|lrrr||lrrr|}
\hline
\textbf{State} & \textbf{Total} & \textbf{Pos.} & \textbf{Neg.} & \textbf{State} & \textbf{Total} & \textbf{Pos.} & \textbf{Neg.}\\
\hline
AK & 558 & 196 & 362 & MT & 636 & 221 & 415 \\\hline
AL & 3,410 &  1,292 & 2,118 & NC & 7,263 & 2,222 & 5,041 \\\hline
AR & 2,341 & 842 & 1,499 & ND & 605  & 179 & 426 \\\hline
AZ & 5,521 & 2,347 & 3,174 & NE & 1,429 & 424 & 1,005 \\\hline
CA & 20,040 & 7,116 & 12,924 & NH & 900  & 344 & 556 \\\hline
CO & 4,164 & 1,362 & 2,802 & NJ & 7,347  & 3,762 & 3,585 \\\hline
CT & 2,595 & 1,180 & 1,415 & NM & 1,203  & 416 & 787 \\\hline
DE & 805 & 342 & 463 & NV & 1,994 & 792 & 1,202 \\\hline
FL & 18,898 &  8,158 & 10,740 & NY & 18,268 & 8,134 & 10,134 \\\hline
GA & 7,901 & 2,461 & 5,440 & OH & 11,634 & 4,339 & 7,295 \\\hline
HI & 1,060 & 433 & 627 & OK & 2,575  & 865 & 1,710 \\\hline
IA & 2,951 & 1,093 & 1,858 & OR & 3,637 & 1,322 & 2,315 \\\hline
ID & 1,021 & 361 & 660 & PA & 11,817  & 4,658 & 7,159 \\\hline
IL & 8,932 & 3,497 & 5,435 & RI & 423 & 158 & 265 \\\hline
IN & 4,388 &1,523 & 2,865 & SC & 3,386 & 1,194 & 2,192 \\\hline
KS & 1,857  & 524 & 1,333 & SD & 651 & 266 & 385 \\\hline
KY & 4,110 & 1,396 & 2,714 & TN & 5,671 & 2,038 & 3,633 \\\hline
LA & 3,316 & 1,150 & 2,166 & TX & 15,785 & 5,152 & 1,0633 \\\hline
MA & 3,492 &1,453 & 2,039 & UT & 2,117 & 495 & 1,622 \\\hline
MD & 5,145  & 1,959 & 3,186 & VA & 6,057 & 1,924 & 4,133 \\\hline
ME & 1,183 &  473 & 710 & VT & 445 & 150 & 295 \\\hline
MI & 9,744 & 3,629 & 6,115 & WA & 6,247 & 2,009 & 4,238 \\\hline
MN & 3,504 & 1,371 & 2,133 & WI & 2,893 & 1,281 & 1,612 \\\hline
MO & 4,341 & 1,572 & 2,769 & WV & 1,712 & 617 & 1,095 \\\hline
MS & 2,246 &  710 & 1,536 & WY & 358 & 113 & 245 \\\hline
\end{tabular}
}
\end{table}

We begin with a real-world healthcare dataset composed of 17 tables. To simplify preprocessing and retain the most relevant information, we keep only:
\textit{Patient Demographic Table},
\textit{Diagnosis Table},
\textit{Procedure Table},
\textit{Medication Drug Table},
\textit{Lab Result Table},
and
\textit{Vital Sign Table}.

\smallskip
\noindent \textbf{Data Processing.}
We merge these tables by \textit{patient\_ID}, ensuring each patient record contains both \textit{static features} (e.g.,\ zipcode, sex) and a series of \textit{events} (medical codes plus numerical attributes, and the patient’s age at each event).
Because multiple medical coding systems are used across the U.S.\ healthcare spectrum, we unify these codes into a single standardized terminology.

Next, a board-certified medical oncologist provides a set of \textit{pancreatic cancer} diagnostic codes, which are used to extract data labels. In ICD-10, the codes for malignant neoplasm of the pancreas include: 
\begin{itemize}[leftmargin=*]
    \item C25 (Malignant neoplasm of pancreas, general),
    \item C25.0 (Malignant neoplasm of head of pancreas),
    \item C25.1 (Malignant neoplasm of body of pancreas),
    \item C25.2 (Malignant neoplasm of tail of pancreas),
    \item C25.3 (Malignant neoplasm of pancreatic duct),
    \item C25.4 (Malignant neoplasm of endocrine pancreas),
    \item C25.7 (Malignant neoplasm of other specified parts of pancreas),
    \item C25.8 (Malignant neoplasm of overlapping lesions of pancreas),
    \item C25.9 (Malignant neoplasm of pancreas, unspecified).
\end{itemize}
In ICD-9, the corresponding codes are: 
\begin{itemize}[leftmargin=*]
    \item 157 (Malignant neoplasm of pancreas, general),
    \item 157.0 (Malignant neoplasm of head of pancreas),
    \item 157.1 (Malignant neoplasm of body of pancreas), 
    \item 157.2 (Malignant neoplasm of tail of pancreas), 
    \item 157.3 (Malignant neoplasm of pancreatic duct),
    \item 157.4 (Malignant neoplasm of islets of Langerhans),
    \item 157.8 (Malignant neoplasm of other specified sites of pancreas),
    \item 157.9 (Malignant neoplasm of pancreas, unspecified).
\end{itemize}

For each patient, if any of these codes appear in the longitudinal record, we set the label $y=1$; furthermore, we remove any events that occur at or after the time of pancreatic cancer diagnosis to prevent data leakage.
If none of the pancreatic cancer codes appear for a patient, we set that patient’s label to $y=0$.

After these preprocessing steps, we obtain $265{,}085$ de-identified patient samples spanning $50$ U.S.\ states.\footnote{All sensitive identifiers are removed or hashed; events after a pancreatic cancer diagnosis are excluded.}
We randomly sample $10\%$ ($26{,}509$) of this dataset as a \textit{global validation set}.
The remaining $90\%$ of the data is returned to each state, where each state splits its local data into a \textit{local training set} and a \textit{local test set} (e.g.\ $7{:}2$).
The binary label $y$ indicates whether the patient eventually develops pancreatic cancer.
Table~\ref{tab:state_data_stats} summarizes, for each of the 50 states, the number of local samples, how many are used for training/testing, and the distribution of positive/negative labels.
Globally, we observe $99{,}604$ positive and $165{,}481$ negative samples from local total data(train and test).

\smallskip
\noindent \textbf{Model and Training Setup.}
To handle this longitudinal EHR data, we embed the event codes, concatenate them with numerical attributes (e.g.\ \textit{age, value}), and feed the sequence into a two-layer bidirectional GRU, which captures both forward and backward dependencies.
We then apply an attention mechanism over the GRU outputs, computing importance weights for each time step and aggregating them into a context vector.
This vector is concatenated with the embedded static features (\textit{sex, postal\_code}), and a linear classifier predicts whether the patient will develop pancreatic cancer.

We adopt a focal loss~\cite{lin2017focal} to mitigate label imbalance and train with an Adam optimizer (default PyTorch settings).
The mini-batch size is set to $B=64$, GRU hidden dimension to $256$, dropout probability to $0.3$, and learning rate to $\texttt{lr}=0.00001$.
We embed \textit{sex} and \textit{postal\_code} into vectors of dimensions 8 and 16, respectively, zero-padding time-series inputs (with sequence lengths tracked via \texttt{pack\_padded\_sequence}).
Specifically, the two-layer GRU produces a $2\times256$-dim vector per time step, mapped by the attention layer to a $256$-dim energy vector and finally to scalar scores for softmax weighting.
Ultimately, we train and validate our model in this federated setting, treating each state as one client ($K=50$).

%% file: main.bbl

\begin{thebibliography}{31}


\ifx \showCODEN    \undefined \def \showCODEN     #1{\unskip}     \fi
\ifx \showDOI      \undefined \def \showDOI       #1{#1}\fi
\ifx \showISBNx    \undefined \def \showISBNx     #1{\unskip}     \fi
\ifx \showISBNxiii \undefined \def \showISBNxiii  #1{\unskip}     \fi
\ifx \showISSN     \undefined \def \showISSN      #1{\unskip}     \fi
\ifx \showLCCN     \undefined \def \showLCCN      #1{\unskip}     \fi
\ifx \shownote     \undefined \def \shownote      #1{#1}          \fi
\ifx \showarticletitle \undefined \def \showarticletitle #1{#1}   \fi
\ifx \showURL      \undefined \def \showURL       {\relax}        \fi
\providecommand\bibfield[2]{#2}
\providecommand\bibinfo[2]{#2}
\providecommand\natexlab[1]{#1}
\providecommand\showeprint[2][]{arXiv:#2}

\bibitem[Gao et~al\mbox{.}(2022)]%
        {g}
\bibfield{author}{\bibinfo{person}{Liang Gao}, \bibinfo{person}{Hongchao Fu}, \bibinfo{person}{Lili Li}, \bibinfo{person}{Yanyan Chen}, \bibinfo{person}{Min Xu}, {and} \bibinfo{person}{Cheng-Zhong Xu}.} \bibinfo{year}{2022}\natexlab{}.
\newblock \showarticletitle{FedDC: Federated Learning with Non-iid Data via Local Drift Decoupling and Correction}. In \bibinfo{booktitle}{\emph{Proceedings of the IEEE/CVF Conference on Computer Vision and Pattern Recognition}}. \bibinfo{pages}{10112--10121}.
\newblock


\bibitem[Goksu and Pugeault(2024)]%
        {goksu2024robust}
\bibfield{author}{\bibinfo{person}{Ozgu Goksu} {and} \bibinfo{person}{Nicolas Pugeault}.} \bibinfo{year}{2024}\natexlab{}.
\newblock \showarticletitle{Robust Federated Learning in the Face of Covariate Shift: A Magnitude Pruning with Hybrid Regularization Framework for Enhanced Model Aggregation}.
\newblock \bibinfo{journal}{\emph{arXiv preprint arXiv:2412.15010}} (\bibinfo{year}{2024}).
\newblock
\urldef\tempurl%
\url{https://arxiv.org/abs/2412.15010}
\showURL{%
\tempurl}


\bibitem[Guo et~al\mbox{.}(2019)]%
        {guo2019crossover}
\bibfield{author}{\bibinfo{person}{Wei Guo}, \bibinfo{person}{Wei Ge}, \bibinfo{person}{Longbo Cui}, \bibinfo{person}{Hua Li}, {and} \bibinfo{person}{Li Kong}.} \bibinfo{year}{2019}\natexlab{}.
\newblock \showarticletitle{An interpretable disease onset predictive model using crossover attention mechanism from electronic health records}.
\newblock \bibinfo{journal}{\emph{IEEE Access}}  \bibinfo{volume}{7} (\bibinfo{year}{2019}), \bibinfo{pages}{134236--134244}.
\newblock


\bibitem[Hinton(2015)]%
        {hinton2015distilling}
\bibfield{author}{\bibinfo{person}{Geoffrey Hinton}.} \bibinfo{year}{2015}\natexlab{}.
\newblock \showarticletitle{Distilling the Knowledge in a Neural Network}.
\newblock \bibinfo{journal}{\emph{arXiv preprint arXiv:1503.02531}} (\bibinfo{year}{2015}).
\newblock


\bibitem[Karimireddy et~al\mbox{.}(2020)]%
        {karimireddy2020scaffold}
\bibfield{author}{\bibinfo{person}{Sai~Praneeth Karimireddy}, \bibinfo{person}{Satyen Kale}, \bibinfo{person}{Mehryar Mohri}, \bibinfo{person}{Sanjiv Reddi}, \bibinfo{person}{Sebastian~U. Stich}, {and} \bibinfo{person}{Ananda~Theertha Suresh}.} \bibinfo{year}{2020}\natexlab{}.
\newblock \showarticletitle{{Scaffold: Stochastic Controlled Averaging for Federated Learning}}. In \bibinfo{booktitle}{\emph{Proceedings of the 37th International Conference on Machine Learning}} \emph{(\bibinfo{series}{ICML})}. \bibinfo{publisher}{PMLR}, \bibinfo{pages}{5132--5143}.
\newblock


\bibitem[Konecn{\'y}(2016)]%
        {konecny2016federated}
\bibfield{author}{\bibinfo{person}{Jakub Konecn{\'y}}.} \bibinfo{year}{2016}\natexlab{}.
\newblock \showarticletitle{Federated Learning: Strategies for Improving Communication Efficiency}.
\newblock \bibinfo{journal}{\emph{arXiv preprint arXiv:1610.05492}} (\bibinfo{year}{2016}).
\newblock
\showeprint[arxiv]{1610.05492}~[cs.LG]


\bibitem[Krizhevsky and Hinton(2009)]%
        {krizhevsky2009learning}
\bibfield{author}{\bibinfo{person}{Alex Krizhevsky} {and} \bibinfo{person}{Geoffrey Hinton}.} \bibinfo{year}{2009}\natexlab{}.
\newblock \bibinfo{booktitle}{\emph{Learning Multiple Layers of Features from Tiny Images}}.
\newblock \bibinfo{type}{{T}echnical {R}eport}. \bibinfo{institution}{University of Toronto}.
\newblock
\newblock
\shownote{Technical Report}.


\bibitem[Kweon et~al\mbox{.}(2021)]%
        {kweon2021bidirectional}
\bibfield{author}{\bibinfo{person}{Wonbin Kweon}, \bibinfo{person}{SeongKu Kang}, {and} \bibinfo{person}{Hwanjo Yu}.} \bibinfo{year}{2021}\natexlab{}.
\newblock \showarticletitle{Bidirectional distillation for top-K recommender system}. In \bibinfo{booktitle}{\emph{Proceedings of the Web Conference 2021}}. \bibinfo{pages}{3861--3871}.
\newblock


\bibitem[Li et~al\mbox{.}(2020a)]%
        {li2020federated}
\bibfield{author}{\bibinfo{person}{Tian Li}, \bibinfo{person}{Anit~Kumar Sahu}, \bibinfo{person}{Ameet Talwalkar}, {and} \bibinfo{person}{Virginia Smith}.} \bibinfo{year}{2020}\natexlab{a}.
\newblock \showarticletitle{Federated learning: Challenges, methods, and future directions}.
\newblock \bibinfo{journal}{\emph{IEEE Signal Processing Magazine}} \bibinfo{volume}{37}, \bibinfo{number}{3} (\bibinfo{year}{2020}), \bibinfo{pages}{50--60}.
\newblock


\bibitem[Li et~al\mbox{.}(2020b)]%
        {Li2020FedOpt}
\bibfield{author}{\bibinfo{person}{Tian Li}, \bibinfo{person}{Anit~K. Sahu}, \bibinfo{person}{Manzil Zaheer}, \bibinfo{person}{Maziar Sanjabi}, \bibinfo{person}{Ameet Talwalkar}, {and} \bibinfo{person}{Virginia Smith}.} \bibinfo{year}{2020}\natexlab{b}.
\newblock \showarticletitle{Federated optimization in heterogeneous networks}. In \bibinfo{booktitle}{\emph{Proceedings of Machine Learning and Systems}}, Vol.~\bibinfo{volume}{2}. \bibinfo{pages}{429--450}.
\newblock


\bibitem[Li et~al\mbox{.}(2019)]%
        {li2019convergence}
\bibfield{author}{\bibinfo{person}{Xiang Li}, \bibinfo{person}{Kaixuan Huang}, \bibinfo{person}{Wenhao Yang}, \bibinfo{person}{Shusen Wang}, {and} \bibinfo{person}{Zhihua Zhang}.} \bibinfo{year}{2019}\natexlab{}.
\newblock \bibinfo{title}{On the Convergence of {FedAvg} on Non-IID Data}.
\newblock
\newblock
\showeprint[arxiv]{1907.02189}~[cs.LG]
\newblock
\shownote{arXiv preprint arXiv:1907.02189}.


\bibitem[Lin et~al\mbox{.}(2017)]%
        {lin2017focal}
\bibfield{author}{\bibinfo{person}{Tsung-Yi Lin}, \bibinfo{person}{Priyal Goyal}, \bibinfo{person}{Ross Girshick}, \bibinfo{person}{Kaiming He}, \bibinfo{person}{Piotr Doll{\'a}r}, {and} \bibinfo{person}{Serge Belongie}.} \bibinfo{year}{2017}\natexlab{}.
\newblock \showarticletitle{Focal Loss for Dense Object Detection}.
\newblock \bibinfo{journal}{\emph{arXiv preprint arXiv:1708.02002}} (\bibinfo{year}{2017}).
\newblock


\bibitem[Lyu et~al\mbox{.}(2020)]%
        {Lyu2020CollaborativeFairness}
\bibfield{author}{\bibinfo{person}{Lingjuan Lyu}, \bibinfo{person}{Xinyang Xu}, \bibinfo{person}{Qiang Wang}, \bibinfo{person}{Han Yu}, {et~al\mbox{.}}} \bibinfo{year}{2020}\natexlab{}.
\newblock \showarticletitle{Collaborative Fairness in Federated Learning}. In \bibinfo{booktitle}{\emph{Federated Learning: Privacy and Incentive}}. \bibinfo{pages}{189--204}.
\newblock


\bibitem[Ma et~al\mbox{.}(2017)]%
        {ma2017dipole}
\bibfield{author}{\bibinfo{person}{Fenglong Ma}, \bibinfo{person}{Radha Chitta}, \bibinfo{person}{Jing Zhou}, \bibinfo{person}{Quanzeng You}, \bibinfo{person}{Tong Sun}, {and} \bibinfo{person}{Jing Gao}.} \bibinfo{year}{2017}\natexlab{}.
\newblock \showarticletitle{Dipole: Diagnosis prediction in healthcare via attention-based bidirectional recurrent neural networks}. In \bibinfo{booktitle}{\emph{Proceedings of the 23rd ACM SIGKDD International Conference on Knowledge Discovery and Data Mining}}. \bibinfo{pages}{1903--1911}.
\newblock


\bibitem[McMahan et~al\mbox{.}(2017)]%
        {mcmahan2017communication}
\bibfield{author}{\bibinfo{person}{Brendan McMahan}, \bibinfo{person}{Eider Moore}, \bibinfo{person}{Daniel Ramage}, \bibinfo{person}{Seth Hampson}, {and} \bibinfo{person}{Blaise Ag{\"u}era~y Arcas}.} \bibinfo{year}{2017}\natexlab{}.
\newblock \showarticletitle{Communication-Efficient Learning of Deep Networks from Decentralized Data}. In \bibinfo{booktitle}{\emph{Proceedings of the 20th International Conference on Artificial Intelligence and Statistics (AISTATS)}} \emph{(\bibinfo{series}{Proceedings of Machine Learning Research})}. \bibinfo{publisher}{PMLR}, \bibinfo{pages}{1273--1282}.
\newblock


\bibitem[Ni et~al\mbox{.}(2022)]%
        {Ni2022}
\bibfield{author}{\bibinfo{person}{Xuanming Ni}, \bibinfo{person}{Xinyuan Shen}, {and} \bibinfo{person}{Huimin Zhao}.} \bibinfo{year}{2022}\natexlab{}.
\newblock \showarticletitle{Federated optimization via knowledge codistillation}.
\newblock \bibinfo{journal}{\emph{Expert Systems with Applications}}  \bibinfo{volume}{191} (\bibinfo{year}{2022}), \bibinfo{pages}{116310}.
\newblock
\urldef\tempurl%
\url{https://doi.org/10.1016/j.eswa.2021.116310}
\showURL{%
\tempurl}


\bibitem[Shang et~al\mbox{.}(2023)]%
        {shang2023fedbikd}
\bibfield{author}{\bibinfo{person}{Ertong Shang}, \bibinfo{person}{Hui Liu}, \bibinfo{person}{Zhuo Yang}, \bibinfo{person}{Junzhao Du}, {and} \bibinfo{person}{Yiming Ge}.} \bibinfo{year}{2023}\natexlab{}.
\newblock \showarticletitle{FedBiKD: Federated Bidirectional Knowledge Distillation for Distracted Driving Detection}.
\newblock \bibinfo{journal}{\emph{IEEE Internet of Things Journal}} (\bibinfo{year}{2023}).
\newblock


\bibitem[Tan et~al\mbox{.}(2020)]%
        {tan2020data}
\bibfield{author}{\bibinfo{person}{Qingxiong Tan}, \bibinfo{person}{Min Ye}, \bibinfo{person}{Bin Yang}, \bibinfo{person}{S. Liu}, \bibinfo{person}{A.~J. Ma}, \bibinfo{person}{T.~C.~F. Yip}, \bibinfo{person}{Y. Zhao}, \bibinfo{person}{S.~C. Hui}, \bibinfo{person}{T.~M.~F. Chan}, \bibinfo{person}{F.~K. Chan}, \bibinfo{person}{J.~J.~Y. Sung}, \bibinfo{person}{E.~C. Cheung}, {and} \bibinfo{person}{P. Yuen}.} \bibinfo{year}{2020}\natexlab{}.
\newblock \showarticletitle{{Data-GRU: Dual-Attention Time-Aware Gated Recurrent Unit for Irregular Multivariate Time Series}}. In \bibinfo{booktitle}{\emph{Proceedings of the AAAI Conference on Artificial Intelligence}}, Vol.~\bibinfo{volume}{34}. \bibinfo{pages}{930--937}.
\newblock


\bibitem[Wang et~al\mbox{.}(2024a)]%
        {wang2024fedave}
\bibfield{author}{\bibinfo{person}{Zihui Wang}, \bibinfo{person}{Zhe Peng}, \bibinfo{person}{Xinyu Fan}, \bibinfo{person}{Zheng Wang}, \bibinfo{person}{Siyang Wu}, \bibinfo{person}{Rui Yu}, \bibinfo{person}{...}, {and} \bibinfo{person}{Chunyan Wang}.} \bibinfo{year}{2024}\natexlab{a}.
\newblock \showarticletitle{FedAVE: Adaptive data value evaluation framework for collaborative fairness in federated learning}.
\newblock \bibinfo{journal}{\emph{Neurocomputing}}  \bibinfo{volume}{574} (\bibinfo{year}{2024}), \bibinfo{pages}{127227}.
\newblock


\bibitem[Wang et~al\mbox{.}(2024b)]%
        {wang2024fedsac}
\bibfield{author}{\bibinfo{person}{Zihui Wang}, \bibinfo{person}{Zheng Wang}, \bibinfo{person}{Lingjuan Lyu}, \bibinfo{person}{Zhigang Peng}, \bibinfo{person}{Zhiquan Yang}, \bibinfo{person}{Chuan Wen}, {and} \bibinfo{person}{Xiaohui Fan}.} \bibinfo{year}{2024}\natexlab{b}.
\newblock \showarticletitle{{FedSAC: Dynamic Submodel Allocation for Collaborative Fairness in Federated Learning}}. In \bibinfo{booktitle}{\emph{{Proceedings of the 30th ACM SIGKDD Conference on Knowledge Discovery and Data Mining}}}. \bibinfo{pages}{3299--3310}.
\newblock


\bibitem[Wickramaratne and Mahmud(2020)]%
        {wickramaratne2020sepsis}
\bibfield{author}{\bibinfo{person}{Sajila~D. Wickramaratne} {and} \bibinfo{person}{Md~Shaad Mahmud}.} \bibinfo{year}{2020}\natexlab{}.
\newblock \showarticletitle{Bi-directional gated recurrent unit based ensemble model for the early detection of sepsis}. In \bibinfo{booktitle}{\emph{2020 42nd Annual International Conference of the IEEE Engineering in Medicine \& Biology Society (EMBC)}}. \bibinfo{publisher}{IEEE}, \bibinfo{pages}{70--73}.
\newblock


\bibitem[Xiao et~al\mbox{.}(2017)]%
        {xiao2017fashion}
\bibfield{author}{\bibinfo{person}{Han Xiao}, \bibinfo{person}{Kashif Rasul}, {and} \bibinfo{person}{Roland Vollgraf}.} \bibinfo{year}{2017}\natexlab{}.
\newblock \showarticletitle{Fashion-MNIST: A Novel Image Dataset for Benchmarking Machine Learning Algorithms}.
\newblock \bibinfo{journal}{\emph{arXiv preprint arXiv:1708.07747}} (\bibinfo{year}{2017}).
\newblock


\bibitem[Xu et~al\mbox{.}(2021)]%
        {xu2021gradient}
\bibfield{author}{\bibinfo{person}{Xinyi Xu}, \bibinfo{person}{Lingjuan Lyu}, \bibinfo{person}{Xiaofeng Ma}, \bibinfo{person}{Chunyan Miao}, \bibinfo{person}{Chee~Seng Foo}, {and} \bibinfo{person}{Bo~An Kiat~Huat Low}.} \bibinfo{year}{2021}\natexlab{}.
\newblock \showarticletitle{Gradient driven rewards to guarantee fairness in collaborative machine learning}. In \bibinfo{booktitle}{\emph{Advances in Neural Information Processing Systems}}, Vol.~\bibinfo{volume}{34}. \bibinfo{pages}{16104--16117}.
\newblock


\bibitem[Yan et~al\mbox{.}(2023)]%
        {yan2023criticalfl}
\bibfield{author}{\bibinfo{person}{Gang Yan}, \bibinfo{person}{Haiyan Wang}, \bibinfo{person}{Xue Yuan}, {and} \bibinfo{person}{Jia Li}.} \bibinfo{year}{2023}\natexlab{}.
\newblock \showarticletitle{Criticalfl: A critical learning periods augmented client selection framework for efficient federated learning}. In \bibinfo{booktitle}{\emph{Proceedings of the 29th ACM SIGKDD Conference on Knowledge Discovery and Data Mining}}. ACM, \bibinfo{pages}{2898--2907}.
\newblock


\bibitem[Yang et~al\mbox{.}(2024)]%
        {yang2024fedas}
\bibfield{author}{\bibinfo{person}{Xiyuan Yang}, \bibinfo{person}{Wenke Huang}, {and} \bibinfo{person}{Mang Ye}.} \bibinfo{year}{2024}\natexlab{}.
\newblock \showarticletitle{{FedAS}: Bridging Inconsistency in Personalized Federated Learning}. In \bibinfo{booktitle}{\emph{Proceedings of the IEEE/CVF Conference on Computer Vision and Pattern Recognition (CVPR)}}. \bibinfo{publisher}{IEEE}, \bibinfo{pages}{11986--11995}.
\newblock
\urldef\tempurl%
\url{https://doi.org/10.1109/CVPR52733.2024.01139}
\showDOI{\tempurl}


\bibitem[Yang et~al\mbox{.}(2019)]%
        {yang2019disease}
\bibfield{author}{\bibinfo{person}{Yang Yang}, \bibinfo{person}{Xiangwei Zheng}, {and} \bibinfo{person}{Cun Ji}.} \bibinfo{year}{2019}\natexlab{}.
\newblock \showarticletitle{Disease prediction model based on bilstm and attention mechanism}. In \bibinfo{booktitle}{\emph{2019 IEEE International Conference on Bioinformatics and Biomedicine (BIBM)}}. \bibinfo{publisher}{IEEE}, \bibinfo{pages}{1141--1148}.
\newblock


\bibitem[Ye et~al\mbox{.}(2020)]%
        {ye2020predicting}
\bibfield{author}{\bibinfo{person}{Xiangyang Ye}, \bibinfo{person}{Q.~T. Zeng}, \bibinfo{person}{Julio~C. Facelli}, \bibinfo{person}{Diana~I. Brixner}, \bibinfo{person}{Mike Conway}, {and} \bibinfo{person}{Bradley~E. Bray}.} \bibinfo{year}{2020}\natexlab{}.
\newblock \showarticletitle{Predicting optimal hypertension treatment pathways using recurrent neural networks}.
\newblock \bibinfo{journal}{\emph{International Journal of Medical Informatics}}  \bibinfo{volume}{139} (\bibinfo{year}{2020}), \bibinfo{pages}{104122}.
\newblock


\bibitem[Yurochkin et~al\mbox{.}(2019)]%
        {yurochkin2019bayesian}
\bibfield{author}{\bibinfo{person}{Mikhail Yurochkin}, \bibinfo{person}{Mayank Agarwal}, \bibinfo{person}{Soumya Ghosh}, \bibinfo{person}{Kristjan Greenewald}, \bibinfo{person}{Natesh Hoang}, {and} \bibinfo{person}{Yasaman Khazaeni}.} \bibinfo{year}{2019}\natexlab{}.
\newblock \showarticletitle{Bayesian Nonparametric Federated Learning of Neural Networks}. In \bibinfo{booktitle}{\emph{International Conference on Machine Learning}} \emph{(\bibinfo{series}{Proceedings of Machine Learning Research})}. \bibinfo{publisher}{PMLR}, \bibinfo{pages}{7252--7261}.
\newblock


\bibitem[Zhang et~al\mbox{.}(2018)]%
        {zhang2018patient2vec}
\bibfield{author}{\bibinfo{person}{Jinghe Zhang}, \bibinfo{person}{Kamran Kowsari}, \bibinfo{person}{James~H Harrison}, \bibinfo{person}{Jason~M Lobo}, {and} \bibinfo{person}{Laura~E Barnes}.} \bibinfo{year}{2018}\natexlab{}.
\newblock \showarticletitle{{Patient2vec: A personalized interpretable deep representation of the longitudinal electronic health record}}.
\newblock \bibinfo{journal}{\emph{IEEE Access}}  \bibinfo{volume}{6} (\bibinfo{year}{2018}), \bibinfo{pages}{65333--65346}.
\newblock


\bibitem[Zhang and Shi(2024)]%
        {zhang2024pFedCK}
\bibfield{author}{\bibinfo{person}{Jianfei Zhang} {and} \bibinfo{person}{Yongqiang Shi}.} \bibinfo{year}{2024}\natexlab{}.
\newblock \showarticletitle{A Personalized Federated Learning Method Based on Clustering and Knowledge Distillation}.
\newblock \bibinfo{journal}{\emph{Electronics}} \bibinfo{volume}{13}, \bibinfo{number}{5} (\bibinfo{year}{2024}), \bibinfo{pages}{857}.
\newblock
\showISSN{2079-9292}
\urldef\tempurl%
\url{https://doi.org/10.3390/electronics13050857}
\showDOI{\tempurl}


\bibitem[Zhao et~al\mbox{.}(2018)]%
        {zhao2018federated}
\bibfield{author}{\bibinfo{person}{Yue Zhao}, \bibinfo{person}{Meng Li}, \bibinfo{person}{Liangzhen Lai}, \bibinfo{person}{Naveen Suda}, \bibinfo{person}{Dave Civin}, {and} \bibinfo{person}{Vikas Chandra}.} \bibinfo{year}{2018}\natexlab{}.
\newblock \showarticletitle{Federated learning with non-iid data}.
\newblock \bibinfo{journal}{\emph{arXiv preprint arXiv:1806.00582}} (\bibinfo{year}{2018}).
\newblock


\end{thebibliography}
